\documentclass[journal]{IEEEtran}

\ifCLASSINFOpdf
\else
\fi
\usepackage{amsmath}
\usepackage{amsthm}
\usepackage{algorithm}
\usepackage{algorithmic}
\usepackage{url}
\usepackage{color}
\usepackage{listings}
\usepackage{multirow}
\usepackage{amssymb}
\usepackage{url}
\usepackage{subfigure}
\usepackage{graphicx}
\newtheorem{thm}{Theorem}[section]
\newtheorem{lem}{Lemma}[section]

\newtheorem{defn}{Definition}[section]

\newtheorem{cor}{Corollary}[section]

\usepackage{lineno}
\usepackage{tabularx}

\makeatletter
\renewcommand{\@thesubfigure}{\hskip\subfiglabelskip}
 \makeatother

\def\C{{\cal C}}

\def\M{{\cal M}}

\def\S{{\cal S}}

\def\N{{\mathbb N}}
\def\R{{\mathbb R}}

\def\bfc{{\bf c}}

\def\bfu{{\bf u}}
\def\bfv{{\bf v}}
\def\bfx{{\bf x}}

\def\bfu{{\bf u}}

\def\bfX{{\bf X}}
\def\bfY{{\bf Y}}

\def\bfI{{\bf I}}
\def\bfA{{\bf A}}
\def\bfB{{\bf B}}
\def\bfD{{\bf D}}

\def\bfG{{\bf G}}

\begin{document}

\title{Fuzzy Discriminant Clustering with Fuzzy Pairwise Constraints}
%
%

\author{Zhen Wang, Shan-Shan Wang, Lan Bai, Wen-Si Wang, Yuan-Hai Shao
\IEEEcompsocitemizethanks{
\IEEEcompsocthanksitem Zhen Wang is with
School of Mathematical Sciences, Inner Mongolia University, Hohhot,
010021, P.R.China, and Key Laboratory of Symbolic Computation and Knowledge Engineering of Ministry of Education, Jilin University, Changchun, 130012, P.R.China e-mail: wangzhen@imu.edu.cn.
\IEEEcompsocthanksitem Shan-Shan Wang is with
School of Mathematical Sciences, Inner Mongolia University, Hohhot,
010021, P.R.China e-mail: wangshanshan202103@163.com.
\IEEEcompsocthanksitem Lan Bai is with
School of Mathematical Sciences, Inner Mongolia University, Hohhot,
010021, P.R.China e-mail: imubailan@163.com.
\IEEEcompsocthanksitem Wen-Si Wang is
with Engineering Research Center of Intelligent Perception and Autonomous Control (Ministry of Education), and Faculty of Information Technology, Beijing University of Technology, Beijing 100124, P.R.China, e-mail: wensi.wang@bjut.edu.cn.
\IEEEcompsocthanksitem Yuan-Hai Shao (*Corresponding author) is with School of Management, Hainan University, Haikou,
570228, P.R.China e-mail: shaoyuanhai21@163.com.
}}

\markboth{IEEE TRANSACTIONS ON FUZZY SYSTEMS,~VOL.~X, NO.~X,XXXXXX}%
{Shell \MakeLowercase{\textit{et al.}}: Bare Demo of IEEEtran.cls
for Computer Society Journals}

%



\IEEEcompsoctitleabstractindextext{
\begin{abstract}
In semi-supervised fuzzy clustering, this paper extends the traditional pairwise constraint (i.e., must-link or cannot-link) to fuzzy pairwise constraint. The fuzzy pairwise constraint allows a supervisor to provide the grade of similarity or dissimilarity between the implicit fuzzy vectors of a pair of samples. This constraint can present more complicated relationship between the pair of samples and avoid eliminating the fuzzy characteristics. We propose a fuzzy discriminant clustering model (FDC) to fuse the fuzzy pairwise constraints. The nonconvex optimization problem in our FDC is solved by a modified expectation-maximization algorithm, involving to solve several indefinite quadratic programming problems (IQPPs). Further, a diagonal block coordinate decent (DBCD) algorithm is proposed for these IQPPs, whose stationary points are guaranteed, and the global solutions can be obtained under certain conditions. To suit for different applications, the FDC is extended into various metric spaces, e.g., the Reproducing Kernel Hilbert Space. Experimental results on several benchmark datasets and facial expression database demonstrate the outperformance of our FDC compared with some state-of-the-art clustering models.
\end{abstract}

}

%

\maketitle

\IEEEdisplaynotcompsoctitleabstractindextext

\IEEEpeerreviewmaketitle


\section{Introduction}
Clustering \cite{ClusterBook1}, assigning the given samples into several clusters, has been employed in many real world applications \cite{PartitonSideInformation,PartitonSideInformation2,ClusterTC2,Shen2}. Different from traditional clustering that a sample can belong in only one cluster, fuzzy clustering \cite{FCM,KFCM2} allows it to belong in more clusters with fuzzy memberships. In some applications, e.g., facial expression recognition \cite{FacialA1,FacialA2}, fuzzy clustering is more suitable to present the ground truth than traditional clustering \cite{FacialFuzzy1,FacialFuzzy2,FacialFuzzy3,KFCM,FCM5}. To guide the assignment in clustering, external information was imported given by supervisors. An simple external information is the pairwise constraints \cite{PCCA,DSC}, which assign several pairs of samples in either a cluster or two different clusters, called must-link or cannot-link respectively. In the literature, there have been many clustering and fuzzy clustering models guided by pairwise constraints \cite{PCCA,DSC,SSDC,FCM3,FCMreview2015,FCMreview2019,FCM4,FCM2}.

\begin{figure}[htbp]
\centering
\includegraphics[width=0.35\textheight]{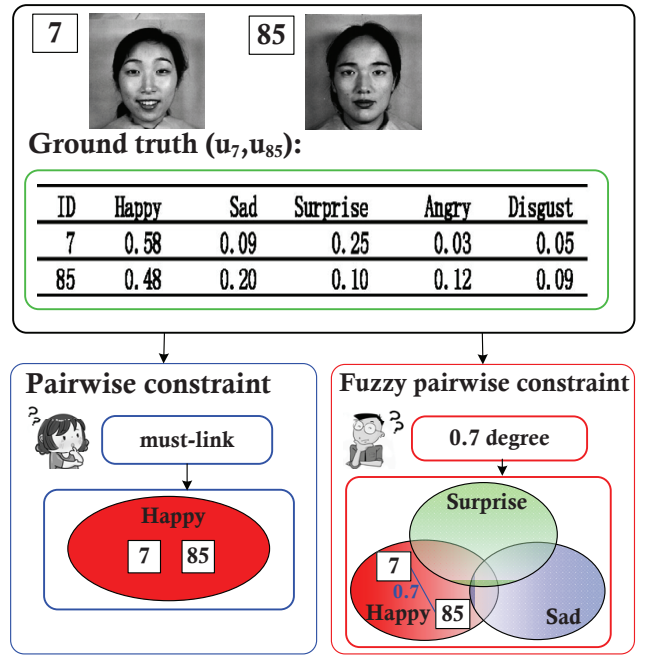}
\caption{Illustration of the traditional pairwise constraint and the proposed fuzzy pairwise constraint in facial expression fuzzy clustering. The images are derived from the JAFFE database \cite{JAFFE}, and this database also provides a labeled fuzzy expressions (i.e., the ground truth). Its details can be found in Section IV.}\label{FigFicalExp}
\end{figure}

Note that the fuzzy vectors are formed by the fuzzy memberships. However, the above alternative choice in pairwise constraint conceals the complicated relationship between the pair of samples in fuzzy clustering, where the must-link constraints require the largest memberships of the fuzzy vectors to be in the same cluster, and the cannot-link constraints require the largest memberships of the fuzzy vectors to be in different clusters. For a supervisor, it is usually not easy to make sure whether they are must-link or cannot-link when the fuzzy vectors have two or more dominant memberships. Furthermore, for the sample under such a pairwise constraint, its fuzzy memberships assemble in a definite cluster, losing its fuzzy characteristics. Fig. \ref{FigFicalExp} illustrates this restriction of pairwise constraint encountered in facial expression fuzzy clustering. Given a pair of images, a supervisor needs to decide whether they are must-link or cannot-link, while a personal image often contains more facial expressions that an adult can understand. As shown in Fig. \ref{FigFicalExp}, the 7th image expresses happy with a bit surprise, while the 85th image also shows happy but with a bit sad. Thus, the supervisor is hard to decide the constraint between the 7th and 85th images. Even though the supervisor gives a decision (e.g., must-link) by her expertise, the pairwise constraint would obliterate the fuzzy characteristics, resulting in a pure cluster. Similar phenomenon may appear on the cannot-link decision. In fact, if a supervisor decides a pair of samples to be must-link or cannot-link, she/he offers two viewpoints: the first one implies that each of the two samples belongs to a cluster without any hesitation, and the second one manifests that the two samples are in the same cluster or in different clusters. Therefore, the pairwise constraints does not quite match with fuzzy clustering.

The pairwise constraints have been softened to allow supervisors give the confidence level for their decisions, e.g., probabilistic constraints in model-based clustering \cite{ProbPairCon}, soft constraints in clustering applications \cite{SoftPairCon,SoftPairCon2,SoftConSurvey}, and fuzzy constraints in hierarchical clustering \cite{FuzzyCon,FPC}. However, these pairwise constraints still fuse the hypothesis of must-link and cannot-link, and thus they cannot reveal the complicated relationship between the pair of samples in fuzzy clustering.

In this paper, completely ruled must-link and cannot-link out, we propose a new instance-level pairwise constraint in fuzzy clustering, called fuzzy pairwise constraint. For a pair of samples, it is concerned with the similarity or dissimilarity between their implicit fuzzy vectors. As shown in Fig. \ref{FigFicalExp}, the supervisor gives his decision breezily, and the fuzzy characteristics of fuzzy vectors are retained. To fuse our fuzzy pairwise constraints in fuzzy clustering, a fuzzy discriminant clustering model (FDC) is proposed subsequently. Mathematically, the nonconvex optimization problem in our model is solved by a modified expectation-maximization (MEM) algorithm, involving to solve indefinite quadratic programming problems (IQPPs). It is well known that the general IQPP are still intrinsically hard problem \cite{IQPPreview1,IQPPreview2}. Noticing the specific properties of our IQPP, a diagonal block coordinate decent (DBCD) algorithm is constructed, where a stationary point is guaranteed, and its global solution can be obtained under certain conditions. To promote the performance of our FDC, it is extended to various metric spaces to suit for different data types and distributions \cite{Jaccard,Cosine,Manhattan,Minkowski}. As an example, the kernel FDC is proposed via kernel tricks \cite{KFCM,KFCM2,Kernel1}.

The main contributions of this paper includes:

\noindent(i) To match fuzzy clustering, a new fuzzy pairwise constraint is proposed, resulting in a fuzzy discriminant clustering model (FDC).

\noindent(ii) The optimization problem in our FDC is solved by a modified expectation-maximization (MEM) algorithm. To solve nonconvex subproblems involved, an efficient diagonal block coordinate decent (DBCD) algorithm is designed.

\noindent(iii) Our model is extended to various metric spaces.

\noindent(iv) Experimental results on the benchmark datasets and facial expression database confirm its competitive performance.

This paper is organized as follows. Section II briefly reviews the pairwise constraints and the related fuzzy clustering. In Section III, the fuzzy pairwise constraint is defined first, and then, we elaborate the FDC, including its formation, the MEM algorithm to FDC, the DBCD algorithm to IQPP, and the extension of FDC in sequence. Experiments are arranged in Section IV and conclusions are given in Section V.

\section{Background}
Remind the clustering problem with $m$ samples in the $n$-dimensional real space denoted by $\bfX=(\bfx_1,\bfx_2,\ldots,\bfx_m)\in \R^{n\times m}$. In non-fuzzy clustering, it aims at assigning the samples into $k$ clusters with their corresponding labels $\bfY=(y_1,y_2,\ldots,y_m)^\top\in\{1,2,\ldots,k\}^m$. In fuzzy clustering, it aims at assigning each sample into all the $k$ clusters with $k$ fuzzy memberships. These fuzzy memberships can be organized by $m$ fuzzy vectors $\{\bfu_i\in\R^{k}|i=1,\ldots,m\}$. The $i$-th $(i=1,\ldots,m)$ fuzzy vector $\bfu_i=(\bfu_i(1),\ldots,\bfu_i(k))^\top$ indicates the membership degrees of sample $\bfx_i$ to the $k$ clusters. Apparently, the fuzzy memberships can be converted to labels easily by
\begin{eqnarray}\label{UtoY}
\begin{array}{l}
y_i=\arg\underset{j=1,\ldots,k}{\max}~\bfu_i(j).
\end{array}
\end{eqnarray}

\subsection{Hard/Soft Pairwise Constraints}
The traditional pairwise constraints (called hard pairwise constraints) are defined by the must-link set $\M$ and cannot-link set $\C$, where a pair of samples in $\M$ or $\C$ indicates they are in the same cluster or different clusters, respectively. In the literature, hard pairwise constraints require that clustering models must comply with these constrains \cite{SoftConSurvey}. However, the constraints supplied by supervisors are not exactly correct and often contradict from different supervisors \cite{PartitonSideInformation,PartitonSideInformation2}. Therefore, hard pairwise constraints were always slacked in clustering models \cite{PCCA,FCM3,FCM7,FCMreview2019}.

In model-based clustering, the probabilistic pairwise constraints \cite{ProbPairCon} were proposed by setting the confidence of constraints in $\M$ with probability $p$ and ignoring the cannot-link constraints. The probability $p$ of a must-link pair is given by the supervisor according to her/his expertise. It indicates that the two samples are with the same label with probability $p$, and with different labels with probability $1-p$. Then, the probabilistic constraints were extended to soft/fuzzy constraints both on $\M$ and $\C$ with probabilities in hierarchical clustering \cite{FPC} and other clustering problems \cite{SoftPairCon,SoftPairCon2}. The confidence level of a pair of samples is in $[0,1]$ corresponding to $\M$ or $\C$ by the supervisor, where $1$ indicates the definite decision and $0$ denotes unknown. For consistency, the above probabilistic, soft or fuzzy constraints based on hard pairwise constrains are called soft pairwise constraints uniformly.

Neither hard nor soft pairwise constraints can reflect the fuzzy characteristics in fuzzy clustering, because these constraints indicate that the decisions are made from must-link or cannot-link alternatively.


\subsection{Fuzzy Clustering with Pairwise Constraints}
In fuzzy clustering, the fuzzy vectors are normalized as $\{\bfu_i\in \mathbf{[0,1]}^k|\sum\limits_{j=1}^k \bfu_i(j)=1,i=1,\ldots,m\}$ generally. The classical fuzzy c-means (FCM) \cite{FCM} is a fuzzy clustering model without external information. It requires that the samples are close to the cluster prototypes $\{\bfc_j\in\R^n|j=1,\ldots,k\}$ by the extent of their fuzzy memberships, resulting in a nonconvex problem as
\begin{eqnarray}\label{FCM}
\begin{array}{l}
\underset{\{\bfu_i\},\{\bfc_j\}}{\min}\sum\limits_{i=1}^m\sum\limits_{j=1}^k\bfu_i(j)^\gamma||\bfx_i-\bfc_j||^2,
\end{array}
\end{eqnarray}
where $\gamma>0$ is a fuzzy parameter to control the level of fuzzy memberships, and $||\cdot||$ denotes the $L_2$ norm. The above problem can be solved by the expectation-maximization (EM) algorithm \cite{EMalgorithm}.

Some fuzzy clustering models extends FCM with pairwise constraints \cite{FCMreview2015,FCMreview2019,SoftFCM}. For pairwise constraints, the typical way to utilize them in fuzzy clustering is to utilize the outer product. For a pair of samples (e.g., $\bfx_p$ and $\bfx_q$), the outer product of their fuzzy vectors is defined as
\begin{eqnarray}\label{OuterProduct}
\begin{array}{l}
\bfu_p\times \bfu_q=\left( \begin{array}{lll}
\bfu_p(1)\bfu_q(1)& \cdots&\bfu_p(1)\bfu_q(k)\\
\vdots&\ddots&\vdots\\
\bfu_p(k)\bfu_q(1)&\cdots&\bfu_p(k)\bfu_q(k)
\end{array}
\right).
\end{array}
\end{eqnarray}
If the above two samples are from pairwise constraints, they should be sheer samples, i.e., their fuzzy vectors $\bfu_p$ and $\bfu_q$ should tend to the vectors consist of $1$ and $0$. Furthermore, if they are from $\M$, the only $1$ of these vectors should be at the same index. Note the restrictions on $\bfu_p$ and $\bfu_q$. The two targets can be realized simultaneously by maximizing the trace of \eqref{OuterProduct} (i.e., inner product) or minimizing the sum of elements in \eqref{OuterProduct} without its diagonal (called non-diagonal product) \cite{PCCA}. On the contrary, if two samples are from $\mathcal{C}$, one can minimize the corresponding inner product or maximize the non-diagonal product. For instance, a semi-supervised fuzzy clustering model (PCCA) \cite{PCCA,FCM3} minimizes the non-diagonal product on $\mathcal{M}$ and the inner product on $\mathcal{C}$ as
\begin{eqnarray}\label{FCMinner}
\begin{array}{l}
\underset{\{\bfu_i\},\{\bfc_j\}}{\min}~~\sum\limits_{i=1}^m\sum\limits_{j=1}^k\bfu_{i}(j)^2
||\bfx_i-\bfc_j||^2+\beta(\sum\limits_{(p,q)\in \mathcal{M}}\sum\limits_{j=1}^k\\\sum\limits_{l=1,l\neq j}^k\bfu_p(j) \bfu_q(l)+\sum\limits_{(p,q)\in \mathcal{C}}\sum\limits_{j=1}^k\bfu_p(j) \bfu_q(j))\\
s.t.~~~~\sum\limits_{j=1}^k\bfu_i(j)=1,~~\forall i=1,\ldots,m,\\
~~~~~~~~\bfu_i(j)\geq0,~~~~\forall i=1,\ldots,m,~\forall j=1,\ldots,k,
\end{array}
\end{eqnarray}
where $\beta>0$ is a tradeoff parameter.
Correspondingly, the outer product is optimized on both $\mathcal{M}$ and $\mathcal{C}$ in ref. \cite{FCM5}. In ref. \cite{FCM4}, a weighted fuzzy clustering model with pairwise constraints was proposed to furnish dissimilarity among the samples. However, the nonconvex optimization problems in the above models were solved by some greedy methods instead of EM type algorithm, because the convergence of EM for these problems cannot be guaranteed due to the non-convex subproblems in the maximization step. Thus, to avoid the nonconvex subproblems, several researchers hired the pairwise constraints in a pre-step beyond the optimization problem, e.g., introducing the entropy regularization \cite{FCM6} or dissimilarity measurement \cite{FCM7} to keep it consistent with the pairwise constraints. Additionally, the cluster number $k$ should be given before implementing these FCM-based clustering models. A clustering regularization was introduced in semi-supervised clustering to select the cluster number \cite{PCCA,FCM2}. Other FCM-based semi-supervised clustering models refer to the review articles \cite{FCMreview2015,FCMreview2019}.

\section{Fuzzy Discriminant Clustering (FDC)}

\subsection{Fuzzy pairwise constraint}
\begin{defn}
Given a pair of samples $\{\bfx_p,\bfx_q\in\bfX\}$, its fuzzy pairwise constraint is defined as $s_{pq}\in[-1,1]$ to measure the similarity or dissimilarity degree between their implicit fuzzy vectors $\bfu_p$ and $\bfu_q$, where $(0,1]$ is used for similarity degree, $[-1,0)$ is used for dissimilarity degree, and $0$ denotes unknown.
\end{defn}

Given a pair of samples, supervisors should decide whether they are similar or dissimilar according to their implicit cluster vectors firstly, and then consider the similarity or dissimilarity degree. As shown in Fig. \ref{FigFicalExp}, the supervisor decides the similarity degree to be $0.7$ between the unknown facial expressions of the 7th and 85th images, on account of the two persons smile similar with a little different expressions. Mathematically, our fuzzy pairwise constraint refers to consider two implicit fuzzy vectors. Generally, they are apt to similar if one of the two vectors has large values on some components and meanwhile the other has as many large values as possible on the corresponding components. Conversely, they are apt to dissimilar if one has large values on some components and the other has as many small values as possible on the corresponding components. The similarity or dissimilarity degree measures the difference between the pairs of components in the fuzzy vectors. Therefore, considering the discrepancy between the pair of the largest components results in the hard pairwise constraint, which is a degeneration of our fuzzy pairwise constraint.

\subsection{Formation of the model}
Given a fuzzy pairwise constraint set $\S=\{s_{pq}\}$ with its index set $\N_\S=\{(p,q)\}$, the FDC is formulated as
\begin{eqnarray}\label{1}
\begin{array}{l}
\underset{\{\bfu_i\},\{\bfc_j\}}{\min}~~\sum\limits_{i=1}^m\sum\limits_{j=1}^k(\bfu_i(j)^\gamma-\alpha)||\bfx_i-\bfc_j||^2
\\~~~~~~~~~~~~+\beta\sum\limits_{(p,q)\in \N_\S}C(\bfu_p,\bfu_q)\\
s.t.~~~~\sum\limits_{j=1}^k\bfu_i(j)=1,~~\forall i=1,\ldots,m,\\
~~~~~~~~\bfu_i(j)\geq0,~~~~\forall i=1,\ldots,m,~\forall j=1,\ldots,k,
\end{array}
\end{eqnarray}
where $0\leq\alpha<1$ and $\beta>0$ are parameters, $\gamma>0$ is the fuzzy parameter, and the cost of each fuzzy pairwise constraint is defined as
\begin{eqnarray}\label{Cost3}
\begin{array}{l}
C(\bfu_p,\bfu_q)=\left\{\begin{array}{ll}\frac{1}{2}s_{pq}||\bfu_p-\bfu_q||^2&\text{if }s_{pq}\geq0,\\
-s_{pq}\bfu_p^\top \bfu_q &\text{otherwise}.
\end{array}\right.
\end{array}
\end{eqnarray}

Our FDC consists of the prototype aggregation and constraint guidance. In the prototype aggregation (i.e., the first term in the objective of problem \eqref{1}), a discriminative structure that each sample is close to its cluster prototype and far away from the other cluster prototypes is proposed. Due to each sample contributes on each cluster by its fuzzy memberships, we consider that a sample can affect each cluster prototype from positive (i.e., close to it) or negative (i.e, far away from it) aspect by the corresponding fuzzy membership. Thus, parameter $\alpha$ can be regarded as a threshold to control the effect of samples on the cluster prototypes. In addition, the metering of $\alpha$ should be consistent with $\gamma$. For example, for $\gamma=2$, set $\alpha=0.25$ if the user deems that the samples affects positively with $\bfu_i(j)\geq 0.5$.

In the constraint guidance (i.e., the last term in the objective of problem \eqref{1}), the $L_2$ norm or inner product are fused in \eqref{Cost3} for different fuzzy pairwise constraints. It is interesting that the measurements of similarity and dissimilarity are asymmetric to preserve the fuzzy characteristics. Suppose there are two fuzzy vectors $\bfu_p, \bfu_q\in\R^3$. Their similarity may be estimated in some manner, e.g., inner product or non-diagonal product stated in Section II.B. However, it is infeasible to use their inner product and/or non-diagonal product for similarity. Note that maximizing the inner product or minimizing the non-diagonal product leads to sheer samples. For example, $s_{pq}=1$ implies $\bfu_p=\bfu_q$, while maximizing inner product $\bfu_p^\top\bfu_q$ or minimizing non-diagonal product $\sum(\bfu_p\times\bfu_q)-\bfu_p^\top\bfu_q$ leads to one of the three equations: $\bfu_p=\bfu_q=(1,0,0)^\top$, $\bfu_p=\bfu_q=(0,1,0)^\top$, and $\bfu_p=\bfu_q=(0,0,1)^\top$. Thus, we should minimize $||\bfu_p-\bfu_q||$ for similar $\bfu_p$ and $\bfu_q$ to preserve their fuzzy characteristics. Correspondingly, the $L_2$ norm is not a good manner for dissimilarity. For example, $s_{pq}=-1$ implies that $\bfu_p$ and $\bfu_q$ are totally different, i.e., the nonzero elements in $\bfu_p$ correspond to zeros in $\bfu_q$, and vice versa. However, maximizing $||\bfu_p-\bfu_q||$ leads to sheer samples, e.g., $\bfu_p=(1,0,0)^\top$ and $\bfu_q=(0,0,1)^\top$. To preserve the fuzzy characteristics of dissimilarity, the inner product is our choice. In summary, to preserve the fuzzy characteristics, we should minimize the $L_2$ norm of two fuzzy vectors for similarity and maximize their inner product for dissimilarity, as the formation of \eqref{Cost3}.

%

\subsection{Solving the main problem}
For the fuzzy parameter $r$ in fuzzy clustering,
researchers suggested $1.5\leq \gamma\leq2.5$ for its better performance with the suitable fuzzy level \cite{FCM,FCM2,FCM3,FCM4,FCM5,FCM6,FCM7,FCMreview2015,FCMreview2019}. Thus, we set $\gamma=2$ in the main problem \eqref{1} for its computational simplicity.

\setlength{\parskip}{1\baselineskip}
\noindent\emph{C.a. Framework of MEM algorithm}
\setlength{\parskip}{0.5\baselineskip}

Problem \eqref{1} is a nonconvex optimization problem, and we propose a modified expectation-maximization (MEM) algorithm to solve it. Starting from an initial $\{\bfu_i^{(0)}\}$, the cluster prototypes $\{\bfc_j^{(t)}\}$ in the expectation step and the fuzzy vectors $\{\bfu_i^{(t)}\}$ in the maximization step are updated alternately with $t=1,2,\ldots$, by solving \eqref{1} with the fixed counterparts, until meet some terminate conditions. The framework of MEM algorithm is summarized in Algorithm 1.

\begin{algorithm}[H]\label{alg:EMS}
\caption{Framework of MEM algorithm to solve problem \eqref{1}}
\textbf{Input:} Dataset $X$, fuzzy pairwise constraints $\mathcal{S}$, cluster number upper bound $k$, parameter $\alpha\in[0,1)$ and $\beta>0$.\\
\textbf{Output:} Membership vectors $\{\bfu_i\}$.
\par\setlength\parindent{0em}Initialize $\{\bfu_i^{(0)}\}$ and set $t=0$.
\par\setlength\parindent{0em}\textbf{while} true
\par\setlength\parindent{1em}\textbf{(a) Expectation step:}
\par\setlength\parindent{1em}Fix $\{\bfu_i^{(t)}\}$ and update $\{\bfc_j^{(t)}\}$ by solving
\begin{eqnarray}\label{3}
\begin{array}{l}
\underset{\{\bfc_j\}}{\min}~~\sum\limits_{j=1}^k\sum\limits_{i=1}^m(\bfu_i^{(t)}(j)^2-\alpha)||\bfx_i-\bfc_j||^2,
\end{array}
\end{eqnarray}
which can be decomposed into $k$ subproblems with $j=1,\ldots,k$ as
\begin{eqnarray}\label{4}
\begin{array}{l}
\underset{\bfc_j}{\min}~~\sum\limits_{i=1}^m(\bfu_i^{(t)}(j)^2-\alpha )||\bfx_i-\bfc_j||^2.
\end{array}
\end{eqnarray}
\par\setlength\parindent{1em}\textbf{(b) Maximization step:}
\par\setlength\parindent{1em}Fix $\{\bfc_j^{(t)}\}$ and update $\{\bfu_i^{(t+1)}\}$ by solving
\begin{eqnarray}\label{6}
\begin{array}{l}
\underset{\{\bfu_i\}}{\min}~~\sum\limits_{i=1}^m\sum\limits_{j=1}^k\bfu_i(j)^2||\bfx_i-\bfc_j^{(t)}||^2
+\frac{\beta}{2}\sum\limits_{0<s_{pq}\in\S}\\~~~~~~~~s_{pq}||\bfu_p-\bfu_q||^2
-\beta\sum\limits_{0>s_{pq}\in\S}s_{pq}\bfu_p\top \bfu_q\\
s.t.~~~~\sum\limits_{j=1}^k\bfu_i(j)=1,~~\forall i=1,\ldots,m,\\
~~~~~~~~\bfu_i(j)\geq0,~~~~\forall i=1,\ldots,m,~\forall j=1,\ldots,k.
\end{array}
\end{eqnarray}
According to the index set $\N_\S$ of $\S$, the above problem can be decomposed into two subproblems:
\begin{eqnarray}\label{7}
\begin{array}{l}
\underset{\{\bfu_i|i\notin \N_\S\}}{\min}~~\sum\limits_{j=1}^k \bfu_i(j)^2||\bfx_i-\bfc_j^{(t)}||^2\\
s.t.~~~~\sum\limits_{j=1}^k\bfu_i(j)=1,\\
~~~~~~~~\bfu_i(j)\geq0,~~\forall j=1,\ldots,k,
\end{array}
\end{eqnarray}
and
\begin{eqnarray}\label{Type2}
\begin{array}{l}
\underset{\{\bfu_i|i\in \N_\S\}}{\min}~~\sum\limits_{i\in\N_{\S}}^m\sum\limits_{j=1}^k\bfu_i(j)^2||\bfx_i-\bfc_j^{(t)}||^2
+\frac{\beta}{2}\sum\limits_{0<s_{pq}\in\S}\\~~~~~~~~s_{pq}||\bfu_p-\bfu_q||^2
-\beta\sum\limits_{0>s_{pq}\in\S}s_{pq}\bfu_p\top \bfu_q\\
s.t.~~~~\sum\limits_{j=1}^k\bfu_i(j)=1,~~\forall i\in\N_{\S},\\
~~~~~~~~\bfu_i(j)\geq0,~~~~\forall i\in\N_{\S},~\forall j=1,\ldots,k.
\end{array}
\end{eqnarray}
\par\setlength\parindent{1em}\textbf{(c) Termination check:}
\par\setlength\parindent{1em}If $\{\bfu_i^{(t)}\}$ is unchanged, break the loop; Otherwise, set $t=t+1$.
\end{algorithm}
\noindent$\mathbf{Remark.}$ Compared with the EM algorithm, several modifications are added in the MEM algorithm. Firstly, the input includes only an upper bound $k$ as the cluster number instead of the cluster number itself. This is realized in the expectation step, where the solution to subproblem \eqref{4} may not exist, resulting in reducing cluster number. Then,
in the maximization step, if an iterative algorithm is employed to solve problem \eqref{Type2}, its initial point is set to be the previous one, resulting in the convergence of our algorithm.
\begin{thm}\label{ThmFrame}
The series of the objectives of problem \eqref{1} obtained by MEM (Algorithm 1) converges, whatever the global, local or saddle points to subproblems \eqref{Type2} are obtained in iteration.
\end{thm}
Its proof is given in Appendix A.

In Algorithm 1, subproblems \eqref{4}, \eqref{7} and \eqref{Type2} need to be solved in the loop, and we will elaborate their solutions in the following. For simplicity, the superscripts that denote the iterative step are ignored in these subproblems.


\setlength{\parskip}{1\baselineskip}
\noindent\emph{C.b. Solutions to subproblems \eqref{4} and \eqref{7}}
\setlength{\parskip}{0.5\baselineskip}


In the expectation step, for the $j$-th ($j=1,\ldots,k$) subproblem \eqref{4}, if
\begin{eqnarray}\label{check}
\begin{array}{l}
\sum\limits_{i=1}^m(\bfu_i(j)^2-\alpha)\leq0,
\end{array}
\end{eqnarray}
its solution does not exist, which implies that this cluster prototype $\bfc_j$ is infinite. Thus, any sample does not belong in this cluster, and we delete this cluster (i.e., delete prototype $\bfc_j$ and the $j$-th dimension in fuzzy vectors $\{\bfu_j\}$). Otherwise, the closed-form solution to subproblem \eqref{4} is
\begin{eqnarray}\label{5}
\begin{array}{l}
\bfc_j=\frac{\sum\limits_{i=1}^m(\bfu_i(j)^2-\alpha )\bfx_i}{\sum\limits_{i=1}^m(\bfu_i(j)^2-\alpha )}.
\end{array}
\end{eqnarray}

In the maximization step, the closed-form solution to subproblem \eqref{7} with $i\notin\N_\S$ is
\begin{eqnarray}\label{Solution7}
\begin{array}{l}
\bfu_i=\frac{1}{\sum\limits_{j=1}^k\frac{1}{||\bfx_i-\bfc_j||^2}}(\frac{1}{||\bfx_i-\bfc_1||^2},\ldots,\frac{1}{||\bfx_i-\bfc_k||^2})^\top.
\end{array}
\end{eqnarray}

\setlength{\parskip}{1\baselineskip}
\noindent\emph{C.c. Solution to subproblem \eqref{Type2}}
\setlength{\parskip}{0.5\baselineskip}

Note that subproblem \eqref{Type2} is separable. We partition $\mathcal{S}$ into several mutually disjoint subsets $S_1,\ldots,S_h$ w.r.t. the sample index. Let $\N_{S_1},\ldots,\N_{S_h}$ be the index sets of $S_1,\ldots,S_h$ respectively, where the samples in each index set are associated with each other directly or indirectly. Then, subproblem \eqref{Type2} is decomposed by $\{S_l,\N_{S_l}|l=1,\ldots,h\}$ into $h$ subproblems as
\begin{eqnarray}\label{8}
\begin{array}{l}
\underset{\{\bfu_i|i\in \N_{S_l}\}}{\min}\sum\limits_{i\in \N_{S_l}}\sum\limits_{j=1}^k \bfu_{i}(j)^2||\bfx_{i}-\bfc_j||^2
+\frac{\beta}{2}\sum\limits_{0<s_{pq}\in S_l}\\s_{pq}(\sum\limits_{j=1}^k \bfu_{p}(j)^2+\sum\limits_{j=1}^k \bfu_{q}(j)^2)
-\beta\sum\limits_{s_{pq}\in S_l}s_{pq}\bfu_p^{(t)\top} \bfu_q\\
s.t.~~~~\sum\limits_{j=1}^k\bfu_i(j)=1,~~\forall i\in \N_{S_l},\\
~~~~~~~~\bfu_i(j)\geq0,~~~~\forall i\in \N_{S_l},~\forall j=1,\ldots,k.
\end{array}
\end{eqnarray}

For simplicity, subproblem \eqref{8} is reformulated as
\begin{eqnarray}\label{OriginQPP}
\begin{array}{l}
\underset{\bfu}{\min}~~\frac{1}{2}\bfu^\top \bfD\bfu\\
s.t.~~~~\bfA\bfu=\mathbf{1}^{r},\\
~~~~~~~~\bfu\geq\mathbf{0}^{rk},
\end{array}
\end{eqnarray}
where $\bfu=(\bfu_1^\top,\ldots,\bfu_r^\top)^\top=(u_{11},\ldots,u_{1k},u_{21},\ldots,u_{2k},$ $\ldots,u_{rk})^\top\in \mathbb{R}^{rk}$, $\bfA=(\mathbf{1}^{k\top},\mathbf{0}^{(r-1)k\top};\mathbf{0}^{k\top},\mathbf{1}^{k\top},
\mathbf{0}^{(r-2)k\top};$ $\cdots;\mathbf{0}^{(r-1)k\top},\mathbf{1}^{k\top})\in \mathbb{R}^{r\times rk}$, $r=|\N_{S_l}|$, and the vectors $\mathbf{1}^{r}$ and $\mathbf{0}^{rk}$ consist of $r$ ones and $rk$ zeros, respectively. $\bfD\in\mathbb{R}^{rk\times rk}$ is a symmetric matrix and can be partitioned by $k$ rows and $k$ columns as
\begin{eqnarray}\label{D}
\bfD=\left(\begin{array}{ccc}
\bfD_{11}&\cdots&\bfD_{1r}\\
\vdots&\ddots&\vdots\\
\bfD_{r1}&\cdots&\bfD_{rr}
\end{array}\right),
\end{eqnarray}
where the diagonal blocks (i.e., $\bfD_{ii}$ with $i=1,\ldots,r$) are the diagonal matrices whose diagonal elements are larger than zero, and $\bfD_{ij}=\bfD_{ji}^\top$ with $i,j=1,\ldots,r$. Thereinto, for $i=1,\ldots,r$,
\begin{eqnarray}\label{utoD1}
\begin{array}{l}
\bfD_{ii}=\text{diag}(||\bfx_i-\bfc_1||^2,\ldots,||\bfx_i-\bfc_k||^2)+\frac{\beta}{2}\sum\limits_{s_{ij}>0} s_{ij}\bfI,
\end{array}
\end{eqnarray}
where $\bfI$ is the identity matrix, and for $i,j=1,\ldots,r$ ($i\neq j$),
\begin{eqnarray}\label{utoD2}
\begin{array}{l}
\bfD_{ij}=-\text{diag}(\frac{\beta s_{ij}}{2},\ldots,\frac{\beta s_{ij}}{2}).
\end{array}
\end{eqnarray}

Note $\text{tr}(\bfD)>0$. $\bfD$ has at least a positive eigenvalue, which supports the following lemma.
\begin{lem}\label{lem1}
$\bfD$ is positive semi-definite or indefinite alternatively.
\end{lem}
When $\bfD$ is positive semi-definite, problem \eqref{OriginQPP} is a convex quadratic programming problem (CQPP) and can be solved by some CQPP solvers \cite{CPP} to obtain its global solution. Otherwise, problem \eqref{OriginQPP} is an IQPP. Though there have been some algorithms to solve an IQPP \cite{IQPPsolver1,IQPPsolver2,IQPPsolver3,IQPPsolver4}, their specific formations or large amount of computation impedes the application to our problem. However, we still have an opportunity to obtain the global solution to problem \eqref{OriginQPP} with an indefinite $\bfD$.

Consider the following quadratic programming problem
\begin{eqnarray}\label{MainIQPP}
\begin{array}{l}
\underset{\bfv}{\min}~~\frac{1}{2}\bfv^\top \bfB^\top \bfD\bfB\bfv+\hat{\bfu}^\top \bfD\bfB\bfv\\
s.t.~~~~\bfG\bfv\leq\mathbf{1}^{r},\\
~~~~~~~~\bfv\geq\mathbf{0}^{r(k-1)},
\end{array}
\end{eqnarray}
where $\bfv=(\bfv_1^\top,\ldots,\bfv_r^\top)^\top=(v_{11},\ldots,v_{1(k-1)},v_{21},$ $\ldots,v_{2(k-1)},\ldots,v_{r(k-1)})\in \mathbb{R}^{r(k-1)},~\bfG=(\mathbf{1}^{k-1\top},$ $\mathbf{0}^{(r-1)(k-1)\top};\mathbf{0}^{k-1\top},\mathbf{1}^{k-1\top},
\mathbf{0}^{(r-2)(k-1)\top};\cdots;\mathbf{0}^{(r-1)(k-1)\top},$ $\mathbf{1}^{k-1\top})\in\mathbb{R}^{r\times r(k-1)}$,
\begin{eqnarray}\label{B}
\bfB=\left(\begin{array}{ccccccc}
-\mathbf{1}^{k-1}&\bfI&\\
& &-\mathbf{1}^{k-1}&\bfI\\
&&&&\ddots\\
&&&&&-\mathbf{1}^{k-1}&\bfI
\end{array}\right)^\top.
\end{eqnarray}
\begin{thm}\label{ThmConvert}
Suppose $\bfv^*$ is the solution to problem \eqref{MainIQPP}. Then
\begin{eqnarray}\label{uvSolution}
\bfu^*=\hat{\bfu}+\bfB\bfv^*
\end{eqnarray}
is the solution to problem \eqref{OriginQPP}, where $\hat{\bfu}=(1,\mathbf{0}^{k-1\top},\ldots,1,\mathbf{0}^{k-1\top})^\top$.
\end{thm}
Its proof is given in Appendix B.

\noindent$\mathbf{Remark.}$ Apparently, $\bfB^\top \bfD\bfB$ is a symmetric matrix and can be partitioned by $k-1$ rows and $k-1$ columns as
\begin{eqnarray}\label{BDB}
\bfB^\top \bfD\bfB=\left(\begin{array}{ccc}
\bfD_{11}'&\cdots&\bfD_{1r}'\\
\vdots&\ddots&\vdots\\
\bfD_{r1}'&\cdots&\bfD_{rr}'
\end{array}\right).
\end{eqnarray}
After some algebra, it is easy to deduce that the formation of $\{\bfD_{11}',\ldots,\bfD_{rr}'\}$ resembles the formation of $\{\bfD_{11},\ldots,\bfD_{rr}\}$, i.e., the diagonal blocks $\{\bfD_{11}',\ldots,\bfD_{rr}'\}$ are the diagonal matrices whose diagonal elements are larger than zero. From Lemma \ref{lem1}, we conclude that $\bfB^\top\bfD\bfB$ is positive semi-definite or indefinite alternatively. In fact, $\bfB^\top\bfD\bfB$ may be positive semi-definite even though $\bfD$ is indefinite. For example, $\bfD=(1,0,2,0;0,5,0,2;2,0,1,0;$ $0,2,0,5)$ is indefinite, because $-1$ is one of its eigenvalues. However, $\bfB^\top \bfD\bfB=(6,4;4,6)$ is positive definite obviously. Thus, if $\bfB^\top \bfD\bfB$ is positive semi-definite, the global solution to problem \eqref{OriginQPP} can be obtained by solving problem \eqref{MainIQPP} by some CQPP solvers \cite{CPP}.
\begin{lem}\label{Thm2}
If $\bfB^\top\bfD\bfB$ is indefinite, then $\bfD$ is indefinite.
\end{lem}
Its proof is given in Appendix C.

If $\bfB^\top\bfD\bfB$ is indefinite, we propose a Diagonal Block Coordinate Decent (DBCD) algorithm to solve IQPP \eqref{OriginQPP}. Starting from a feasible point, problem \eqref{OriginQPP} w.r.t. $\bfu_i$ ($i=1,\ldots,r$), i.e.,
\begin{eqnarray}\label{QPP1}
\begin{array}{l}
\underset{\bfu_i}{\min}~~\frac{1}{2}\bfu_i^\top \bfD_{ii}\bfu_i+(\bfu_1^\top,\ldots,\bfu_{i-1}^\top,\bfu_{i+1}^\top,\ldots,\bfu_r^\top)
(\bfD_{i1},\\~~~~~~~~\ldots,\bfD_{i(i-1)},\bfD_{i(i+1)},\ldots,\bfD_{ir})^\top \bfu_i\\
s.t.~~~~\mathbf{1}^{k\top}\bfu_i=1,\\
~~~~~~~~\bfu_i\geq\mathbf{0}^{k},
\end{array}
\end{eqnarray}
is solved in sequence to update $\bfu_i$. The above loop continues until some terminate conditions are satisfied. The final $\bfu$ is set to be the solution to problem \eqref{OriginQPP}.

In the DBCD algorithm, the Hessian matrix $\bfD_{ii}$ of problem \eqref{QPP1} is positive definite obviously from the formation of $\bfD$. Thus, the global solution to CQPP \eqref{QPP1} can be obtained by some CQPP solvers. The convergence of the DBCD algorithm is given as follows, and its proof can be found in Appendix D.
\begin{thm}\label{ThmConvergence}
DBCD algorithm converges to a stationary point to problem \eqref{OriginQPP}.
\end{thm}
Specifically, for a small size problem, we can get a global solution to IQPP \eqref{OriginQPP} by DBCD with exhaustive initial points.
\begin{cor}\label{Cor1}
There is a vertex $\bfu^{(0)}$ to problem \eqref{OriginQPP} with $k=r=2$ such that DBCD with this initial $\bfu^{(0)}$ converges to a global solution to problem \eqref{OriginQPP}, where the vertex is a such point that one of its elements is $1$ and the rest elements are $0$.
\end{cor}
Its proof is given in Appendix E.

The pseudocode to solve problem \eqref{8} is summarized in Algorithm 2.
\begin{algorithm}[H]\label{alg:cd1}
\caption{Solving problem \eqref{8}}
\textbf{Input:} Mutually disjoint fuzzy pairwise constraints $S_l$ with $\N_{S_l}$, $\{\bfx_i|i\in\N_{S_l}\}$, $\{\bfc_j|j=1,\ldots,k\}$, parameter $\beta>0$, CQPP solver, and a small tolerance (typically $tol=1e-3$).\\
\textbf{Output:} Solution $\{\bfu_i^*|i\in\N_{S_l}\}$.
\par\setlength\parindent{0em}1. Build problem \eqref{OriginQPP}.
\par\setlength\parindent{0em}2. If $\bfD$ is positive semi-definite, employ the CQPP solver to solve problem \eqref{OriginQPP}. Then, return the solution and terminate the algorithm.
\par\setlength\parindent{0em}3. If $\bfB^\top\bfD\bfB$ is positive semi-definite, employ the CQPP solver to solve problem \eqref{MainIQPP} and substitute its solution into \eqref{uvSolution} to obtain $\bfu^*$. Then, return $\bfu^*$ and terminate the algorithm.
\par\setlength\parindent{0em}4. If $k=r=2$, set $\bfu^{(0)}$ be each vertex and implement the following loop exhaustively to obtain the smallest objective of \eqref{OriginQPP}; Otherwise, initialize $\bfu^{(0)}$ and implement the following loop once. Set $t=1$.
\par\setlength\parindent{0em}5. \textbf{while} true
\par\setlength\parindent{2em}\textbf{for} $i=1,\ldots,r$
\par\setlength\parindent{3em}for all $j=1,\ldots,r$ and $j\neq i$, set $\bfu_j=\bfu_j^{(t-1)}$, then  \par\setlength\parindent{3em}set $\bfu_i^{(t)}$ be the solution to the CQPP \eqref{QPP1} by the CQPP
\par\setlength\parindent{3em}solver;
\par\setlength\parindent{2em}\textbf{end for}
\par\setlength\parindent{2em}\textbf{if} $||\bfu^{(t)}-\bfu^{(t-1)}||<tol$
\par\setlength\parindent{3em}set $\bfu^*=\bfu^{(t)}$;
\par\setlength\parindent{3em}\textbf{break};
\par\setlength\parindent{2em}\textbf{else} set $t=t+1$;
\par\setlength\parindent{2em}\textbf{end if}
\par\setlength\parindent{1em}\textbf{end while}
\end{algorithm}

\subsection{FDC for Various Metric Spaces}
To suit for different data types and distributions, our FDC is extended into various metric spaces in this subsection. Note that the Euclidean distances between samples and prototypes are fused in \eqref{1} and \eqref{Cost3}. The FDC in a metric space can be formulated by replacing the Euclidean distance with the new distance $d(\bfx_i,\bfc_j)$, and it can also be solved by Algorithm 1. It is worth to notice that the formula for updating the prototypes would be different from \eqref{5} in different metric spaces. However, we can skip updating the prototypes and update the fuzzy vectors immediately if necessary. As an example, we extend the FDC for nonlinear clustering via kernel tricks \cite{Kernel1}.

Suppose $\phi(\cdot):~\mathbb{R}^n\rightarrow\mathbb{R}^d$ is a nonlinear mapping. Our kernel FDC considers
\begin{eqnarray}\label{n1}
\begin{array}{l}
\underset{\{\bfu_i\},\{\bfc_j\}}{\min}~~\sum\limits_{i=1}^m\sum\limits_{j=1}^k(\bfu_i(j)^r-\alpha)||\phi(\bfx_i)-\bfc_j||^2\\
~~~~~~~~~~~~+\beta\sum\limits_{(p,q)\in \N_\mathcal{S}}C(\bfu_p,\bfu_q)\\
s.t.~~~~\sum\limits_{j=1}^k\bfu_i(j)=1,~~\forall i=1,\ldots,m,\\
~~~~~~~~\bfu_i(j)\geq0,~~~~\forall i=1,\ldots,m,~\forall j=1,\ldots,k,
\end{array}
\end{eqnarray}

Problem \eqref{n1} can be solved by MEM apparently. Correspondingly, for fixed $\{\bfu_i\}$ ($i=1,\ldots,m$), the $j$-th ($j=1,\ldots,k$) cluster is deleted if \eqref{check} holds. Otherwise, we have
\begin{eqnarray}\label{n5}
\begin{array}{l}
\bfc_j=\frac{\sum\limits_{i=1}^m(\bfu_i(j)^2-\alpha )\phi(\bfx_i)}{\sum\limits_{i=1}^m(\bfu_i(j)^2-\alpha)},~~~~j=1,\ldots,k.
\end{array}
\end{eqnarray}
If the nonlinear mapping $\phi(\cdot)$ is given, the rest part is similar to linear FDC by replacing $\bfx_i$ with $\phi(\bfx_i)$ for $i=1,\ldots,m$. Otherwise, we can obtain the fuzzy vectors by skipping over the computation of the prototypes and solving the subproblem in the maximization step via kernel tricks. For fixed implicit $\{\bfc_j\}$ ($j=1,\ldots,k$), the corresponding subproblem relates to $||\phi(\bfx_i)-\bfc_j||$ with $i=1,\ldots,m$ and $j=1,\ldots,k$. According to \eqref{n5}, we define
\begin{eqnarray}\label{GeneralForm}
\begin{array}{l}
d(\bfx,\bfc_j)=||\phi(\bfx)-\bfc_j||^2\\=\phi(\bfx)^\top\phi(\bfx)-\frac{2\sum\limits_{i=1}^m(\bfu_i(j)^2-
\alpha)\phi(\bfx_i)^\top\phi(\bfx)}{\sum\limits_{i=1}^m(\bfu_i(j)^2-\alpha)}\\
+\frac{(\sum\limits_{i=1}^m(\bfu_i(j)^2-\alpha )\phi(\bfx_i))^\top(\sum\limits_{i=1}^m(\bfu_i(j)^2-\alpha )\phi(\bfx_i))}{(\sum\limits_{i=1}^m(\bfu_i(j)^2-\alpha))^2}\\
=K(\bfx,\bfx)-\frac{2\sum\limits_{i=1}^m(\bfu_i(j)^2-\alpha)K(\bfx_i,\bfx)}{\sum\limits_{i=1}^m(\bfu_i(j)^2-\alpha)}\\
+\frac{\sum\limits_{i_1=1}^m\sum\limits_{i_2=1}^m(\bfu_{i_1}(j)^2-\alpha)(\bfu_{i_2}(j)^2-\alpha)K(\bfx_{i_1},\bfx_{i_2})}
{(\sum\limits_{i=1}^m(\bfu_i(j)^2-\alpha))^2},
\end{array}
\end{eqnarray}
where $K(\cdot,\cdot)$ is a predefined kernel function refers to the inner product in the Reproducing Kernel Hilbert Space. Thus, by substituting \eqref{GeneralForm} into \eqref{Solution7}, the closed-form solution for $i\notin\N_\S$ is
\begin{eqnarray}\label{nSolution7}
\begin{array}{l}
\bfu_i=\frac{1}{\sum\limits_{j=1}^kd(\bfx_i,\bfc_j)}(\frac{1}{d(\bfx_i,\bfc_1)},\ldots,\frac{1}{d(\bfx_i-\bfc_k)})^\top,
\end{array}
\end{eqnarray}
and the subproblem \eqref{Type2} for each subset $S_l$ with $\N_{S_l}$ becomes to
\begin{eqnarray}\label{nType2}
\begin{array}{l}
\underset{\{\bfu_i|i\in \N_\S\}}{\min}~~\sum\limits_{i\in\N_{\S}}^m\sum\limits_{j=1}^k\bfu_i(j)^2d(\bfx_i,\bfc_j)
+\frac{\beta}{2}\sum\limits_{0<s_{pq}\in\S}\\~~~~~~~~s_{pq}||\bfu_p-\bfu_q||^2
-\beta\sum\limits_{0>s_{pq}\in\S}s_{pq}\bfu_p^{(t)\top} \bfu_q\\
s.t.~~~~\sum\limits_{j=1}^k\bfu_i(j)=1,~~\forall i\in\N_{\S},\\
~~~~~~~~\bfu_i(j)\geq0,~~~~\forall i\in\N_{\S},~\forall j=1,\ldots,k,
\end{array}
\end{eqnarray}
which can be solved similar to \eqref{Type2} by Algorithm 2. So the details are omitted.

\section{Experiments}
In this section, we analyze the clustering performance of our FDC on some benchmark datasets \cite{UCI} and a facial expression database \cite{JAFFE} compared with several state-of-the-art semi-supervised clustering models, including semi-supervised denpeak clustering (SSDC\footnote{\url{https://github.com/Huxhh/SSDC}}) \cite{SSDC}, dominant set clustering (DSC\footnote{\url{https://github.com/erogol/DominantSetClustering}}) \cite{DSC}, fuzzy hierarchical semisupervised clustering (FHSS) \cite{FPC} and pairwise-constrained competitive agglomeration (PCCA) \cite{PCCA}. The classical FCM \cite{FCMreview2015} represented the baseline. All these models were implemented by MATLAB2017, on a PC with an Intel Core Duo Processor (4.2 GHz) with 16GB RAM.
In the experiments, the normalized adjusted rand index (ARI$\in[0\%,100\%]$) \cite{ARI} and normalized mutual information (NMI$\in[0\%,100\%]$) \cite{NMI} were used to measure the clustering performance. Their parameters were optimized to maximize the ARI and/or NMI by grid searching, and the cluster number was set to the real one. For practical convenience, the corresponding FDC Matlab codes have been uploaded upon the github\footnote{\url{https://github.com/gamer1882/FDC}}. The implementation details of these models are as follows.

\begin{table}[H]
\begin{tabularx}{\columnwidth}{lX}
FCM\cite{FCMreview2015}&Without any constraints, it was implemented 20 times by the fcm function with random initialization provided by MATLAB. It output fuzzy vectors by parameter $k$.\\
SSDC\cite{SSDC} &It accepted pairwise constraints and output cluster labels.\\
DSC\cite{DSC} &It accepted pairwise constraints and output cluster labels by parameter $k$.\\
FHSS\cite{FPC} &It accepted soft pairwise constraints and output cluster labels with a cutoff parameter $\alpha=0.05$.\\
PCCA\cite{PCCA} &It accepted pairwise constraints and was implemented 20 times with random initialization. It output fuzzy vectors by parameter $k$ and a tradeoff parameter selected from $\Omega:=\{2^i|i=-8,-7,\ldots,7\}$.\\
FDC &It accepted fuzzy pairwise constraints and was implemented 20 times with random initialization. It output fuzzy vectors by parameter $k$ and two tradeoff parameters, where parameter $\alpha$ was selected from $\Omega$ and parameter $\beta$ was selected from $\{0,0.02,0.04,\ldots,0.3\}$.
\end{tabularx}
\end{table}

\begin{table}
\centering
\caption{Details of benchmark datasets}
\begin{tabular}{llrrr}\hline
Data&Name&Samples ($m$)&Dimension ($n$)&Classes ($k$)\\\hline
(a)&Soybean&47&35&2\\\hline
(b)&Zoo&101&16&7\\\hline
(c)&Echocardiogram&131&10&2\\\hline
(d)&Hepatitis&155&19&2\\\hline
(e)&Wine&178&13&3\\\hline
(f)&Seeds&210&7&3\\\hline
(g)&Heartc&303&14&2\\\hline
(h)&Ecoli&336&7&8\\\hline
(i)&Dermatology&366&34&6\\\hline
(j)&Australia&690&14&2\\\hline
(k)&Creadit&690&15&2\\\hline
(l)&Phishing&1,353&9&3\\\hline
(m)&Car&1,728&6&4\\\hline
(n)&Segment&2,310&18&7\\\hline
(o)&Wave&5,000&21&3\\\hline
(p)&Satimage&6,435&36&6\\\hline
(q)&Two&7,400&20&2\\\hline
(r)&Letter&20,000&16&26\\\hline
(s)&Shuttle&58,000&10&7\\\hline
\end{tabular}\label{Datasets}
\end{table}

\begin{table*}[htbp]
\scriptsize
\caption{Performance of the state-of-the-art clustering models on the benchmark datasets} \centering
\begin{tabular}{llrrrrrr}
\hline
\multirow{2}{*}{Data}& \multirow{2}{*}{Group}&Baseline~~&SSDC~~&DSC~~& FHSS~~&PCCA~~&FDC\\
& &ARI(\%)/NMI(\%)~~&ARI(\%)/NMI(\%)~~&ARI(\%)/NMI(\%)~~&ARI(\%)/NMI(\%)~~&ARI(\%)/NMI(\%)~~&ARI(\%)/NMI(\%)\\\hline
\multirow{3}{*}{(a)}&(i) & \multirow{4}{*}{66.61$\pm$19.07/32.55$\pm$37.61~~}&55.05/32.80$\bullet$&50.00/0.00$\bullet$ &50.18/42.41$\bullet$&68.35$\pm$0.00/41.25$\pm$0.00~~&68.94$\pm$1.88/42.09$\pm$2.73 \\
&(ii) &$\bullet$ &55.05/32.80$\bullet$ &50.00/0.00$\bullet$  &60.71/55.76$\bullet$&68.35$\pm$0.00/41.25$\pm$0.00$\bullet$ &$\mathbf{82.65}$$\pm$4.90/$\mathbf{61.77}$$\pm$7.72\\
&(iii)& $\bullet$& 70.78/64.62$\bullet$&50.00/0.00$\bullet$ &90.75/83.55~~&68.35$\pm$0.00/41.25$\pm$0.00$\bullet$ &$\mathbf{94.26}$$\pm$9.69/$\mathbf{88.49}$$\pm$18.86 \\
&(iv) &$\bullet$ & 70.78/64.62$\bullet$&50.00/0.00$\bullet$ &90.75/83.55$\bullet$& 87.79$\pm$0.00/71.03$\pm$0.00$\bullet$ &$\mathbf{100.0}$$\pm$0.00/$\mathbf{100.0}$$\pm$0.00\\
\hline
\multirow{3}{*}{(b)}&(i)& \multirow{4}{*}{81.68$\pm$6.71/69.18$\pm$7.44~~}&75.89/74.99$\bullet$&50.80/8.78$\bullet$ &56.77/66.19$\bullet$ & 80.88$\pm$0.15/$\mathbf{78.98}$$\pm$0.76~~&80.91$\pm$0.83/77.46$\pm$2.38\\
&(ii)&   &78.37/71.93$\bullet$ &50.80/8.78$\bullet$ &57.24/66.61$\bullet$ & 80.89$\pm$0.17/79.01$\pm$0.91$\bullet$ &81.01$\pm$0.00/$\mathbf{79.89}$$\pm$0.00\\
&(iii) &$\bullet$ &75.96/74.01$\bullet$ &50.80/8.78$\bullet$&80.05/80.42~~& 81.64$\pm$0.15/$\mathbf{80.82}$$\pm$0.99$\bullet$& $\mathbf{86.51}$$\pm$6.53/74.70$\pm$8.75 \\
&(iv)&$\bullet$ & 78.50/74.58$\bullet$ &50.80/8.78$\bullet$ & 85.88/76.13$\bullet$&81.63$\pm$0.00/80.43$\pm$0.00$\bullet$ &$\mathbf{94.62}$$\pm$2.07/$\mathbf{87.56}$$\pm$3.66\\
\hline
\multirow{3}{*}{(c)}&(i) & \multirow{4}{*}{65.03$\pm$5.89/22.66$\pm$9.15}$\bullet$ &47.17/10.09$\bullet$ &60.27/12.65$\bullet$ &50.00/35.51$\bullet$ & 72.68$\pm$0.00/33.81$\pm$0.00$\bullet$&$\mathbf{76.93}$$\pm$0.00/$\mathbf{40.36}$$\pm$0.00 \\
&(ii)  &$\bullet$ &47.38/7.35$\bullet$ &60.27/12.65$\bullet$ &50.80/5.08$\bullet$ &72.68$\pm$0.00/33.81$\pm$0.00$\bullet$ &$\mathbf{78.75}$$\pm$5.58/$\mathbf{45.54}$$\pm$9.24\\
&(iii)&$\bullet$ & 48.17/3.41$\bullet$ &61.16/15.69$\bullet$  & 53.87/32.92$\bullet$&74.78$\pm$0.00/37.06$\pm$0.00$\bullet$ &$\mathbf{78.07}$$\pm$0.00/$\mathbf{42.50}$$\pm$0.00\\
&(iv) &$\bullet$ &  58.75/18.88$\bullet$ & 61.34/15.33$\bullet$&60.90/28.16$\bullet$ &75.96$\pm$0.00/40.33$\pm$0.00$\bullet$&$\mathbf{78.12}$$\pm$0.00/$\mathbf{42.75}$$\pm$0.00\\
\hline
\multirow{3}{*}{(d)}&(i) & \multirow{4}{*}{56.88$\pm$3.81/15.49$\pm$6.84}$\bullet$ & 45.90/5.51$\bullet$&60.12/9.81$\bullet$ &50.01/31.83$\bullet$ &57.29$\pm$2.44/19.60$\pm$5.23$\bullet$ &$\mathbf{65.11}$$\pm$2.40/$\mathbf{23.29}$$\pm$2.72 \\
&(ii) &$\bullet$ & 49.48/9.26$\bullet$&60.12/9.81$\bullet$  &58.41/24.51$\bullet$ &57.95$\pm$0.00/17.09$\pm$0.00$\bullet$ & $\mathbf{66.68}$$\pm$3.42/$\mathbf{22.31}$$\pm$3.80 \\
&(iii) &$\bullet$ & 52.38/9.86$\bullet$ &60.12/9.81$\bullet$ &61.21/21.22$\bullet$ &59.04$\pm$0.00/20.12$\pm$0.00$\bullet$&$\mathbf{67.46}$$\pm$5.54/$\mathbf{25.95}$$\pm$6.21 \\
&(iv)&$\bullet$ & 53.69/12.64$\bullet$&60.12/9.81$\bullet$ &68.07/22.26$\bullet$&59.48$\pm$0.00/18.72$\pm$0.00$\bullet$  &$\mathbf{70.20}$$\pm$0.42/$\mathbf{29.18}$$\pm$0.85\\
\hline
\multirow{3}{*}{(e)}&(i) & \multirow{4}{*}{87.05$\pm$11.50/75.87$\pm$17.01}$\bullet$ & 68.63/45.64$\bullet$&50.05/2.82$\bullet$ &49.46/6.57$\bullet$&94.87$\pm$0.00/87.59$\pm$0.00$\bullet$ &$\mathbf{96.21}$$\pm$0.74/$\mathbf{90.17}$$\pm$1.82\\
&(ii)  &$\bullet$ & 78.74/59.19$\bullet$&50.05/2.82$\bullet$ &50.45/46.95$\bullet$&94.87$\pm$0.00/87.59$\pm$0.00$\bullet$ &$\mathbf{96.54}$$\pm$0.00/$\mathbf{90.88}$$\pm$0.00\\
&(iii) & $\bullet$&78.96/60.16$\bullet$ &50.05/2.82$\bullet$  &55.07/53.33$\bullet$  & 94.87$\pm$0.00/87.59$\pm$0.00$\bullet$&$\mathbf{96.54}$$\pm$0.00/$\mathbf{90.88}$$\pm$0.00 \\
&(iv) & $\bullet$&80.88/63.95$\bullet$ &50.15/3.05$\bullet$ & 68.86/61.35$\bullet$& 94.87$\pm$0.00/87.59$\pm$0.00$\bullet$&$\mathbf{96.54}$$\pm$0.00/$\mathbf{90.88}$$\pm$0.00\\
\hline
\multirow{3}{*}{(f)}&(i) & \multirow{4}{*}{87.96$\pm$0.70/71.62$\pm$1.25}$\bullet$ & 54.55/15.05$\bullet$&59.65/37.95$\bullet$  &50.00/0.00$\bullet$&66.14$\pm$20.28/30.43$\pm$38.24$\bullet$ &88.79$\pm$0.28/73.09$\pm$0.53 \\
&(ii) &$\bullet$ & 54.58/15.59$\bullet$&59.65/37.95$\bullet$  &54.49/50.27$\bullet$& 88.62$\pm$0.00/72.75$\pm$0.00$\bullet$&$\mathbf{89.25}$$\pm$0.00/$\mathbf{73.84}$$\pm$0.00\\
&(iii)&$\bullet$ &73.71/58.55$\bullet$ &60.15/38.74$\bullet$ & 72.27/56.02$\bullet$&88.62$\pm$0.00/72.75$\pm$0.00$\bullet$ &$\mathbf{90.50}$$\pm$0.00/$\mathbf{76.16}$$\pm$0.00\\
&(iv) &$\bullet$ &83.64/67.27$\bullet$ &60.15/38.74$\bullet$ &74.67/57.52$\bullet$&89.25$\pm$0.00/73.84$\pm$0.00$\bullet$ &$\mathbf{91.22}$$\pm$0.23/$\mathbf{77.97}$$\pm$0.46\\
\hline
\multirow{3}{*}{(g)}&(i)& \multirow{4}{*}{70.73$\pm$6.49/32.86$\pm$10.83}$\bullet$ &52.95/15.93$\bullet$ &53.88/13.14$\bullet$ &50.00/34.77$\bullet$& $\mathbf{77.01}$$\pm$0.00/$\mathbf{43.96}$$\pm$0.00~~&74.89$\pm$5.84/40.14$\pm$9.45 \\
&(ii)&$\bullet$ &55.68/20.16$\bullet$ &53.88/13.14$\bullet$  &50.01/34.80$\bullet$ &  $\mathbf{77.49}$$\pm$0.00/$\mathbf{44.80}$$\pm$0.00$\circ$ &75.70$\pm$3.67/41.75$\pm$6.22\\
&(iii) &$\bullet$ & 61.84/27.32$\bullet$&53.95/14.60$\bullet$ & 50.22/34.23$\bullet$& 77.49$\pm$0.00/44.80$\pm$0.00$\bullet$ &$\mathbf{77.89}$$\pm$0.57/$\mathbf{45.53}$$\pm$1.17\\
&(iv) &$\bullet$ & 63.14/29.32$\bullet$&54.27/13.10$\bullet$ &56.63/29.19$\bullet$& 80.53$\pm$0.00/50.22$\pm$0.00$\bullet$ &$\mathbf{82.10}$$\pm$0.42/$\mathbf{53.30}$$\pm$0.81\\
\hline
\multirow{3}{*}{(h)}&(i)& \multirow{4}{*}{68.69$\pm$0.49/55.55$\pm$0.88}$\bullet$ &54.57/31.15$\bullet$ &65.91/34.44$\bullet$ &50.01/51.11$\bullet$ & 68.88$\pm$0.65/56.51$\pm$0.96$\bullet$&$\mathbf{75.85}$$\pm$3.53/52.34$\pm$5.61\\
&(ii) &$\bullet$ & 71.12/51.30$\bullet$ &65.91/34.44$\bullet$  & 50.01/51.11$\bullet$& 68.75$\pm$0.71/56.33$\pm$1.00$\bullet$& $\mathbf{78.37}$$\pm$4.48/53.72$\pm$6.08\\
&(iii)&$\bullet$ &  75.52/54.36$\bullet$ &65.91/34.44$\bullet$ &50.01/51.11$\bullet$ & 68.68$\pm$0.68/56.23$\pm$0.96$\bullet$&$\mathbf{81.32}$$\pm$6.08/$\mathbf{60.18}$$\pm$11.01\\
&(iv) &$\bullet$ & 76.17/58.75$\bullet$ &65.91/34.44$\bullet$ &72.54/57.56$\bullet$ & 68.77$\pm$0.65/56.33$\pm$0.91$\bullet$&$\mathbf{80.37}$$\pm$5.67/$\mathbf{62.85}$$\pm$4.35\\
\hline
\multirow{3}{*}{(i)}&(i)& \multirow{4}{*}{79.10$\pm$4.54/68.48$\pm$6.97}$\bullet$ & 54.30/36.43$\bullet$& 50.58/9.99$\bullet$ &50.04/53.57$\bullet$  &60.50$\pm$9.74/34.76$\pm$32.26$\bullet$  &81.52$\pm$3.81/77.29$\pm$1.78 \\
&(ii)  &$\bullet$ & 70.57/60.47$\bullet$ &50.65/10.81$\bullet$  &50.20/53.65$\bullet$ & 67.89$\pm$3.94/54.38$\pm$6.37$\bullet$&$\mathbf{85.46}$$\pm$4.23/78.87$\pm$5.42\\
&(iii) &$\bullet$ & 75.84/67.81$\bullet$ &50.65/10.81$\bullet$ & 50.42/53.89$\bullet$ &66.13$\pm$0.00/51.21$\pm$0.00$\bullet$ &$\mathbf{86.37}$$\pm$1.17/79.16$\pm$2.14 \\
&(iv) &$\bullet$ & 83.82/73.26$\bullet$ &50.65/10.81$\bullet$ &52.79/35.46$\bullet$ &66.13$\pm$0.00/51.21$\pm$0.00$\bullet$ &$\mathbf{89.73}$$\pm$1.21/80.74$\pm$1.81\\
\hline
\multirow{3}{*}{(j)}&(i)& \multirow{4}{*}{65.92$\pm$7.80/24.78$\pm$12.37}$\bullet$& 53.13/20.82$\bullet$&53.30/6.37$\bullet$ &50.00/32.34$\bullet$ &72.37$\pm$0.00/34.91$\pm$0.00$\bullet$ & $\mathbf{74.76}$$\pm$0.00/$\mathbf{39.62}$$\pm$0.00 \\
&(ii)&$\bullet$& 53.62/20.28$\bullet$&53.30/6.37$\bullet$  &50.08/2.53$\bullet$  &73.56$\pm$0.00/37.01$\pm$0.00$\bullet$& $\mathbf{74.76}$$\pm$0.00/$\mathbf{39.62}$$\pm$0.00\\
&(iii) &$\bullet$&53.78/21.18$\bullet$&53.30/6.37$\bullet$ &50.51/5.21$\bullet$ &73.95$\pm$0.00/37.71$\pm$0.00$\bullet$ & $\mathbf{76.00}$$\pm$0.00/$\mathbf{41.72}$$\pm$0.00\\
&(iv)&$\bullet$& 54.45/12.94$\bullet$ &53.30/6.37$\bullet$ & 52.20/23.06$\bullet$&74.56$\pm$0.00/38.81$\pm$0.00$\bullet$ &$\mathbf{76.66}$$\pm$0.20/$\mathbf{42.97}$$\pm$0.57\\
\hline
\multirow{3}{*}{(k)}&(i)& \multirow{4}{*}{62.91$\pm$6.48/20.05$\pm$9.73~~}& 52.31/14.94$\bullet$ & 56.42/13.03$\bullet$ &50.00/0.00$\bullet$ &$\mathbf{66.36}$$\pm$0.51/25.15$\pm$0.76~ &62.51$\pm$6.27/19.49$\pm$9.21\\
&(ii)&$\bullet$& 53.11/18.49$\bullet$ &56.71/13.50~~  &50.00/32.17$\bullet$ & 66.58$\pm$0.76/25.45$\pm$1.18~~& $\mathbf{67.87}$$\pm$5.11/$\mathbf{27.68}$$\pm$7.99 \\
&(iii)&$\bullet$ & 53.19/18.00$\bullet$&56.90/13.83$\bullet$ & 50.78/7.30$\bullet$ &69.75$\pm$2.50/30.76$\pm$3.71$\bullet$ &$\mathbf{71.13}$$\pm$0.10/$\mathbf{32.79}$$\pm$0.17\\
&(iv) &$\bullet$ & 53.46/18.49$\bullet$&58.24/15.45$\bullet$ &53.26/24.23$\bullet$ & 71.39$\pm$0.00/33.27$\pm$0.15$\bullet$ &$\mathbf{71.98}$$\pm$0.45/$\mathbf{34.62}$$\pm$0.77\\
\hline
\multirow{3}{*}{(l)}&(i)& \multirow{4}{*}{66.23$\pm$3.58/24.37$\pm$4.96~~} &51.17/10.49~~&51.06/5.44~~&50.14/5.96~~&59.81$\pm$0.00/$\mathbf{29.61}$$\pm$0.00~~&63.12$\pm$10.99/19.63$\pm$16.45\\
&(ii)  &$\bullet$ & 51.51/18.44$\bullet$&51.06/5.44$\bullet$ &51.60/$\mathbf{29.57}$$\bullet$ & 67.86$\pm$0.00/25.47$\pm$0.00~~&$\mathbf{68.02}$$\pm$1.44/26.32$\pm$1.55\\
&(iii) &$\bullet$  &52.57/19.28$\bullet$ &51.06/5.44$\bullet$  &52.03/29.68$\bullet$ & 68.07$\pm$0.00/26.34$\pm$0.00$\bullet$ &$\mathbf{70.45}$$\pm$0.30/$\mathbf{30.43}$$\pm$0.55\\
&(iv)&$\bullet$  & 53.14/19.88$\bullet$&51.06/5.44$\bullet$ &60.32/22.67$\bullet$& 69.09$\pm$0.00/27.86$\pm$0.00$\bullet$ &$\mathbf{73.34}$$\pm$5.53/$\mathbf{35.66}$$\pm$8.46\\
\hline
\multirow{3}{*}{(m)}&(i)& \multirow{4}{*}{52.18$\pm$2.00/9.22$\pm$4.63~~} & 51.17/$\mathbf{12.36}$~~&50.00/0.00$\bullet$ &50.00/0.00$\bullet$ &52.32$\pm$2.26/8.60$\pm$4.61~~&  51.73$\pm$2.12/4.91$\pm$5.11\\
&(ii)& & 51.94/18.97~~&50.00/0.00$\bullet$  &50.08/$\mathbf{30.26}$$\bullet$ &$\mathbf{52.56}$$\pm$1.61/9.50$\pm$4.37~~&52.10$\pm$1.67/7.85$\pm$2.87 \\
&(iii) & & 52.23/$\mathbf{19.09}$~~&50.00/0.00$\bullet$  &50.08/30.26$\bullet$  & 52.95 $\pm$2.21/9.99$\pm$4.84~~& $\mathbf{53.11}$$\pm$2.29/7.70$\pm$5.05\\
&(iv) &$\bullet$ &52.50/$\mathbf{19.02}$$\bullet$ &50.00/0.00$\bullet$ &53.12/2.08$\bullet$&55.20$\pm$3.94/14.87$\pm$4.74$\bullet$  &$\mathbf{59.33}$$\pm$4.77/18.22$\pm$8.68\\
\hline
\multirow{3}{*}{(n)}&(i) & \multirow{4}{*}{75.75$\pm$0.56/61.32$\pm$1.08~~}&64.58/55.45$\bullet$&66.09/51.88$\bullet$ &50.00/3.89$\bullet$  & 50.08$\pm$0.18/0.61$\pm$1.36$\bullet$ &75.43$\pm$0.44/58.24$\pm$1.22\\
&(ii)&   &67.71/56.38$\bullet$ &66.09/51.99$\bullet$ &50.05/50.57$\bullet$ & 53.37$\pm$4.85/12.50$\pm$17.68$\bullet$&$\mathbf{75.81}\pm$0.55/61.49$\pm$1.23\\
&(iii)  & $\bullet$&77.28/65.58~~&66.12/51.91$\bullet$ &50.05/50.57$\bullet$  & 57.31$\pm$4.30/25.04$\pm$14.44$\bullet$ &$\mathbf{78.12}\pm$1.70/$\mathbf{66.24}$$\pm$2.92\\
&(iv)  & $\bullet$&78.45/67.16~~& 66.27/51.94$\bullet$&50.05/50.57$\bullet$ &  59.68$\pm$1.05/30.49$\pm$2.41$\bullet$&$\mathbf{78.83}\pm$0.95/$\mathbf{68.08}$$\pm$1.33\\
\hline
\multirow{3}{*}{(o)}&(i) & \multirow{4}{*}{62.38$\pm$0.01/32.83$\pm$0.01~~}&55.00/25.62$\bullet$&51.74/12.69$\bullet$ &50.00/0.00$\bullet$ & 53.68$\pm$5.03/9.68$\pm$13.26$\bullet$&62.35$\pm$0.06/32.31$\pm$0.42\\
&(ii) &$\bullet$ &56.06/23.29$\bullet$ &51.74/12.71$\bullet$ &50.08/30.89$\bullet$ &54.46$\pm$6.14/8.52$\pm$11.73$\bullet$  &$\mathbf{63.29}\pm$0.03/31.78$\pm$0.14\\
&(iii)&$\bullet$ &56.00/25.52$\bullet$ &51.74/12.70$\bullet$ &50.00/35.79$\bullet$ & 62.68$\pm$7.09/24.69$\pm$13.80~~&$\mathbf{64.49}\pm$0.01/$\mathbf{37.24}$$\pm$0.11\\
&(iv) & $\bullet$&57.47/27.83$\bullet$ &51.74/12.70$\bullet$ &50.00/35.90$\bullet$ &61.32$\pm$0.19/29.62$\pm$1.84$\bullet$  &$\mathbf{66.57}\pm$0.25/$\mathbf{38.91}$$\pm$6.46\\
\hline
\multirow{3}{*}{(p)}&(i) & \multirow{4}{*}{76.67$\pm$0.02/60.90$\pm$0.04~~}&58.73/42.39$\bullet$&65.52/47.63$\bullet$&50.00/0.00$\bullet$ & 56.00$\pm$5.85/17.27$\pm$14.34$\bullet$&75.48$\pm$1.28/57.17$\pm$1.67\\
&(ii)& &59.02/50.05$\bullet$ & 65.58/47.80$\bullet$ &50.05/44.82$\bullet$ & 50.00$\pm$0.00/0.00$\pm$0.00$\bullet$ &$\mathbf{77.72}\pm$0.83/$\mathbf{61.65}\pm$0.54\\
&(iii)  & $\bullet$&66.34/53.32$\bullet$ &65.58/47.80$\bullet$ &65.84/44.89$\bullet$ & 51.97$\pm$4.42/7.06$\pm$13.42$\bullet$ &$\mathbf{78.70}\pm$0.68/$\mathbf{63.39}$$\pm$1.45\\
&(iv)& $\bullet$&68.81/53.76$\bullet$ &65.58/47.83$\bullet$ &75.11/59.13$\bullet$ & 52.52$\pm$5.63/6.47$\pm$14.48$\bullet$ &$\mathbf{79.64}\pm$0.79/$\mathbf{67.49}$$\pm$0.37\\
\hline
\multirow{3}{*}{(q)}&(i)& \multirow{4}{*}{87.95$\pm$17.42/69.92$\pm$33.47~~}&52.30/15.10$\bullet$&50.00/0.02$\bullet$ &50.00/27.89$\bullet$ &53.12$\pm$0.00/14.28$\pm$0.00$\bullet$  &87.20$\pm$0.05/63.91$\pm$0.12\\
&(ii)& &54.05/21.89$\bullet$ &50.00/0.02$\bullet$ &50.04/27.71$\bullet$ & 59.56$\pm$5.34/21.19$\pm$11.84$\bullet$ &88.08$\pm$0.06/65.84$\pm$0.15\\
&(iii)& $\bullet$&60.64/29.63$\bullet$ &50.00/0.02$\bullet$ &61.99/35.33$\bullet$ & 58.55$\pm$0.00/25.34$\pm$0.00$\bullet$ &$\mathbf{95.88}\pm$0.05/$\mathbf{85.28}$$\pm$0.16\\
&(iv) & $\bullet$&66.13/37.07$\bullet$ &50.00/0.02$\bullet$ &81.81/49.90$\bullet$ & 87.23$\pm$0.00/64.00$\pm$0.00$\bullet$&$\mathbf{95.95}\pm$0.42/$\mathbf{85.51}$$\pm$1.21\\
\hline
\multirow{3}{*}{(r)}&(i)& \multirow{4}{*}{50.41$\pm$0.16/6.66$\pm$1.46}$\bullet$ &--~~~~~~&--~~~~~~&--~~~~~~& 50.00$\pm$0.00/0.00$\pm$0.00$\bullet$&56.52$\pm$0.41/33.41$\pm$1.01\\
&(ii)   &$\bullet$ &--~~~~~~&--~~~~~~&--~~~~~~& 50.00$\pm$0.00/0.00$\pm$0.00$\bullet$ &56.86$\pm$0.51/34.71$\pm$1.10\\
&(iii)&$\bullet$ &--~~~~~~&--~~~~~~&--~~~~~~& 50.00$\pm$0.00/0.00$\pm$0.00$\bullet$ &$\mathbf{58.30}\pm$0.59/$\mathbf{38.87}$$\pm$0.82\\
&(iv)   &$\bullet$ &--~~~~~~&--~~~~~~&--~~~~~~& 50.00$\pm$0.00/0.00$\pm$0.00$\bullet$ &$\mathbf{58.51}\pm$0.48/$\mathbf{39.93}\pm$1.23\\
\hline
\multirow{3}{*}{(s)}&(i)& \multirow{4}{*}{60.20$\pm$2.37/47.07$\pm$6.51}$\bullet$  &--~~~~~~&--~~~~~~&--~~~~~~& 49.01$\pm$1.19/5.28$\pm$4.94$\bullet$&$\mathbf{78.35}\pm$1.77/$\mathbf{57.69}\pm$2.61\\
&(ii)   &$\bullet$ &--~~~~~~&--~~~~~~&--~~~~~~&49.97$\pm$0.01/0.21$\pm$0.48$\bullet$ &$\mathbf{78.59}\pm$1.47/$\mathbf{57.98}$$\pm$3.01\\
&(iii)  & $\bullet$ &--~~~~~~&--~~~~~~&--~~~~~~& 50.00$\pm$0.00/0.00$\pm$0.00$\bullet$&$\mathbf{79.74}\pm$1.95/$\mathbf{61.26}$$\pm$5.96\\
&(iv) & $\bullet$&--~~~~~~&--~~~~~~&--~~~~~~&50.14$\pm$0.32/1.78$\pm$3.98$\bullet$   &$\mathbf{79.73}\pm$1.29/$\mathbf{61.78}$$\pm$3.39\\
\hline
$\bullet$&&61/76&62/68&66/68&65/68&65/76\\
$\circ$&&1/76&0/76&0/76&0/76&1/76\\
\hline
\end{tabular}\label{Benchmark}
\\`--' denotes out of memory; $\bullet/\circ$ indicates FDC is significantly better/worse than compared model (paired t-tests at 95\% significance level).
\end{table*}

\begin{figure*}[htbp]
\centering
\subfigure[(b) Zoo]{\includegraphics[width=0.365\textheight]{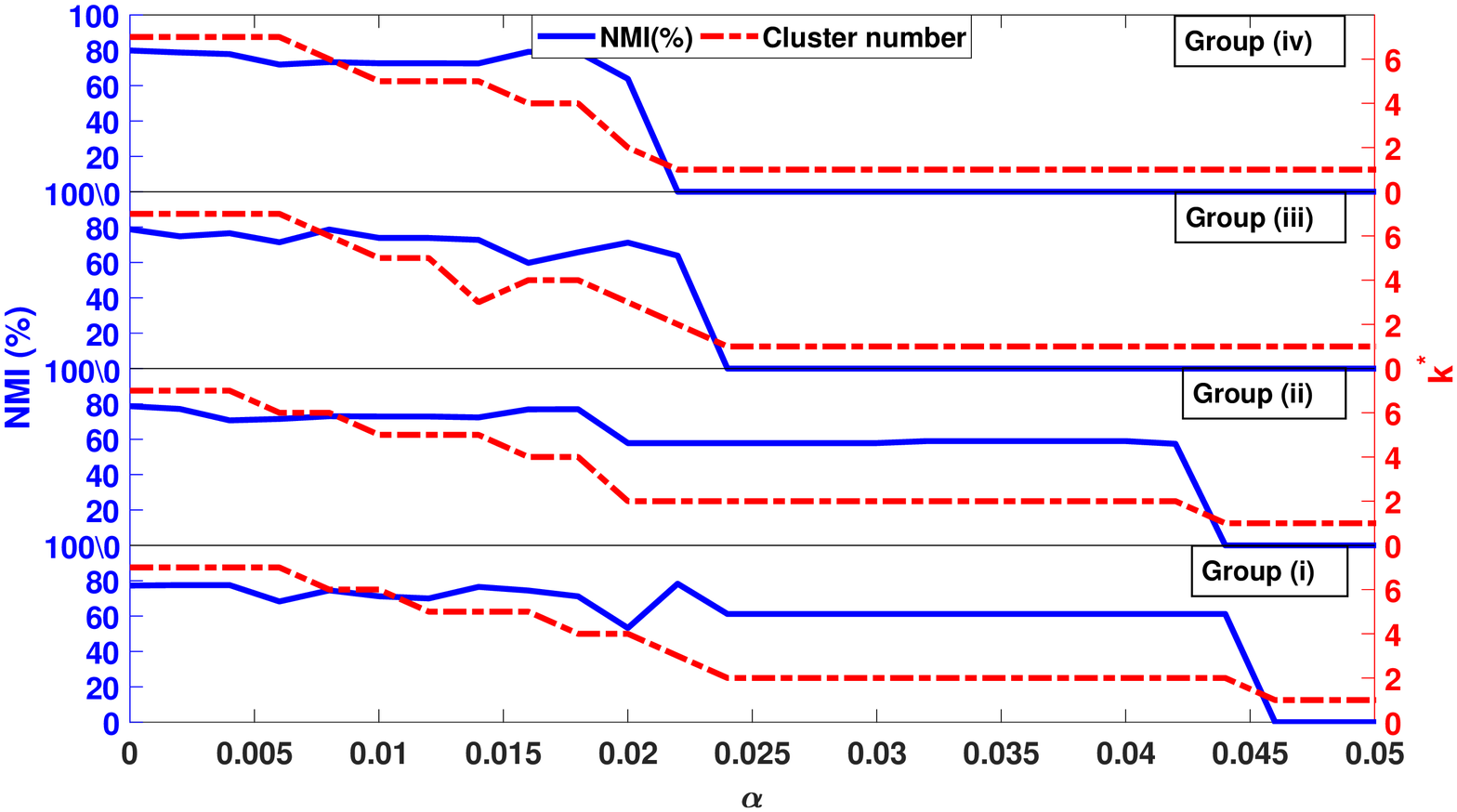}}
\subfigure[(h) Ecoli]{\includegraphics[width=0.365\textheight]{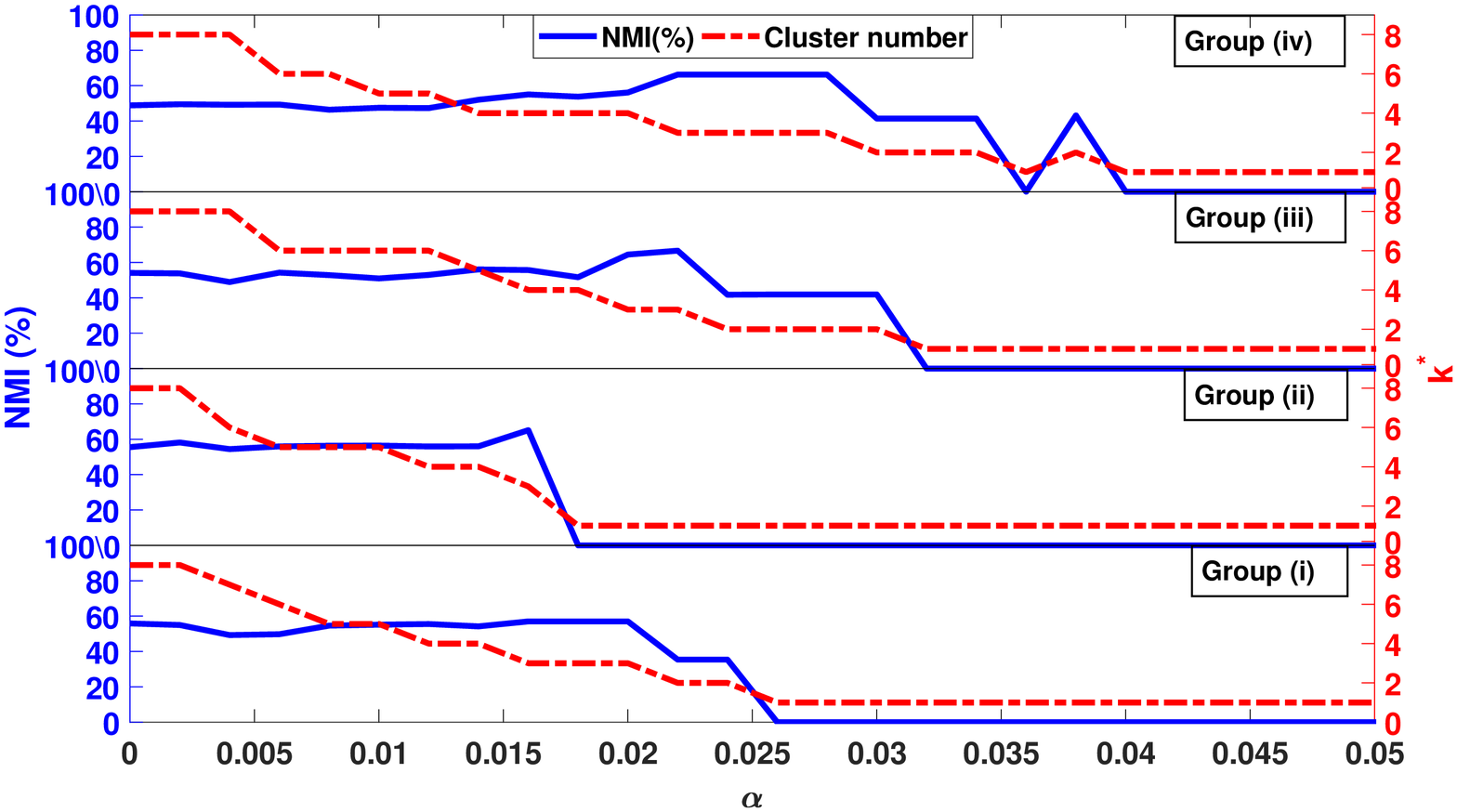}}
\subfigure[(i) Dermatology]{\includegraphics[width=0.365\textheight]{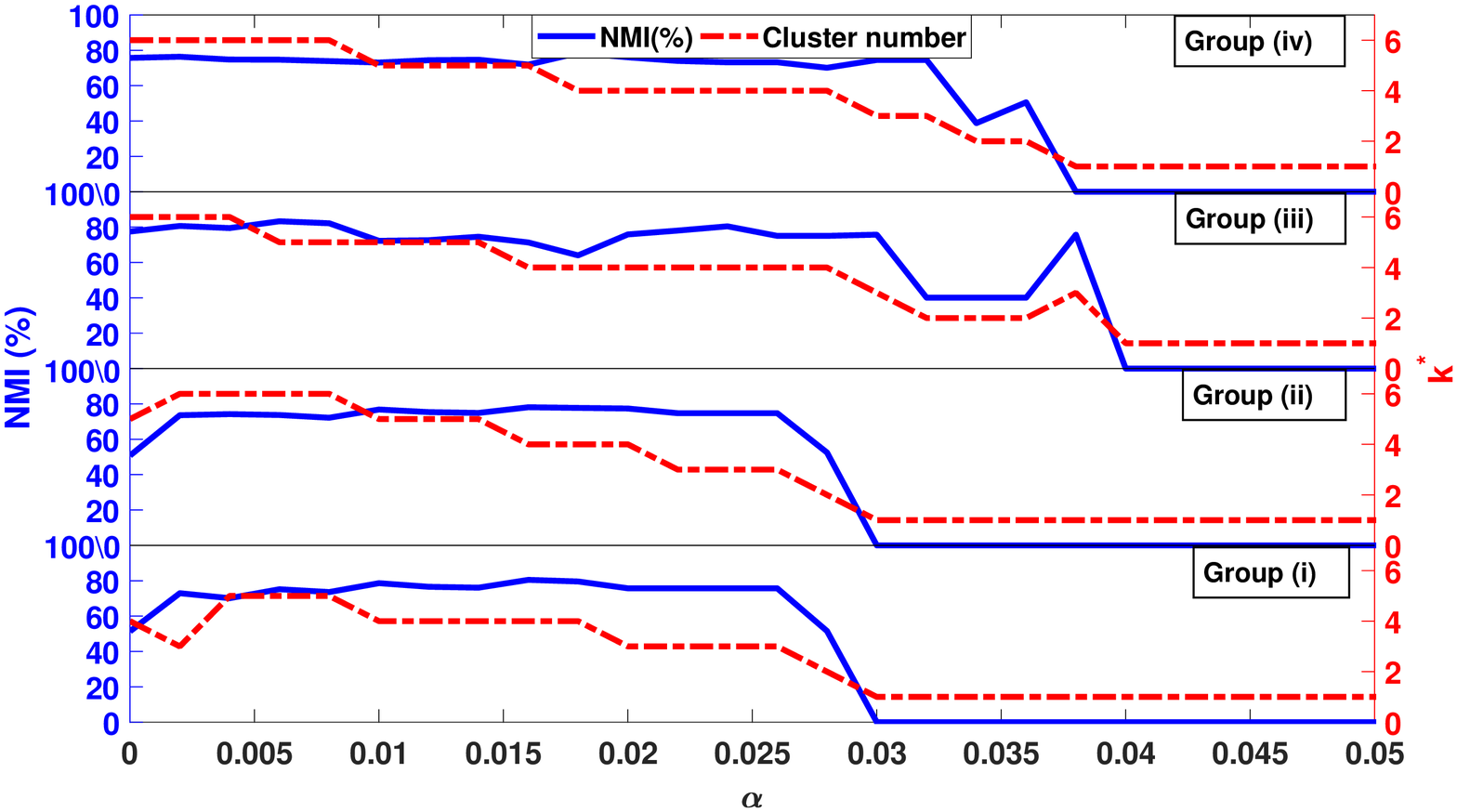}}
\subfigure[(n) Segment]{\includegraphics[width=0.365\textheight]{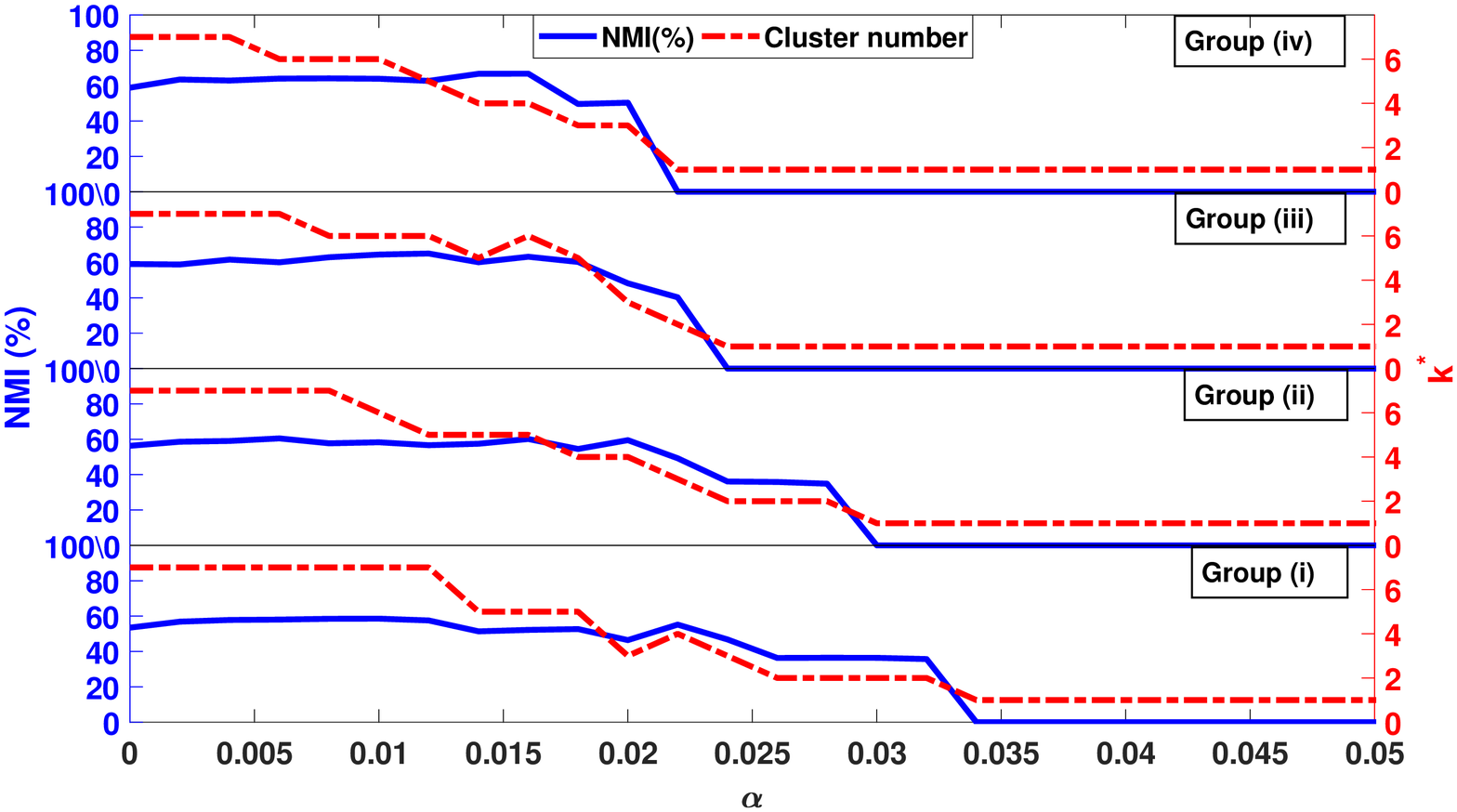}}
\subfigure[(p) Satimage]{\includegraphics[width=0.365\textheight]{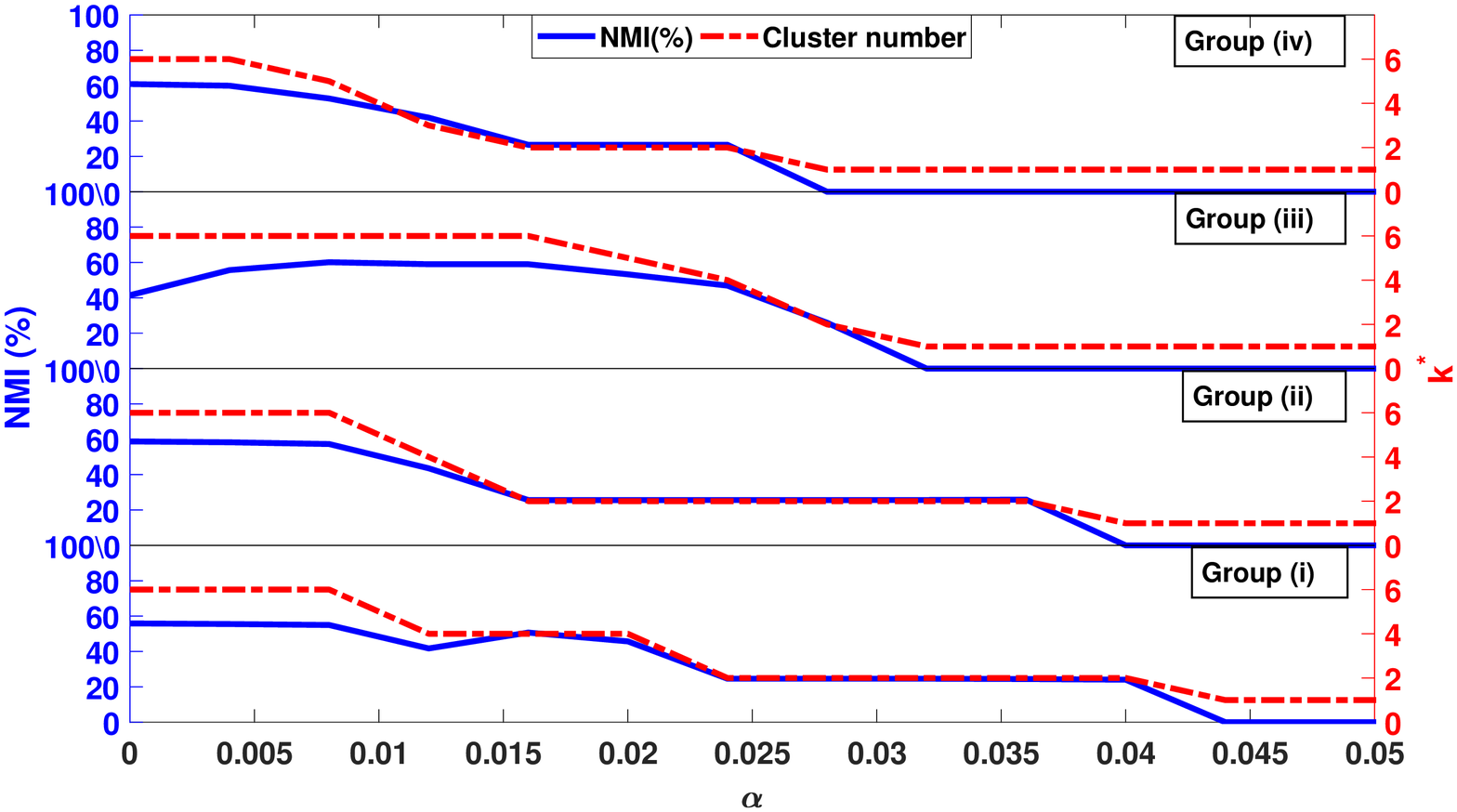}}
\subfigure[(r) Letter]{\includegraphics[width=0.365\textheight]{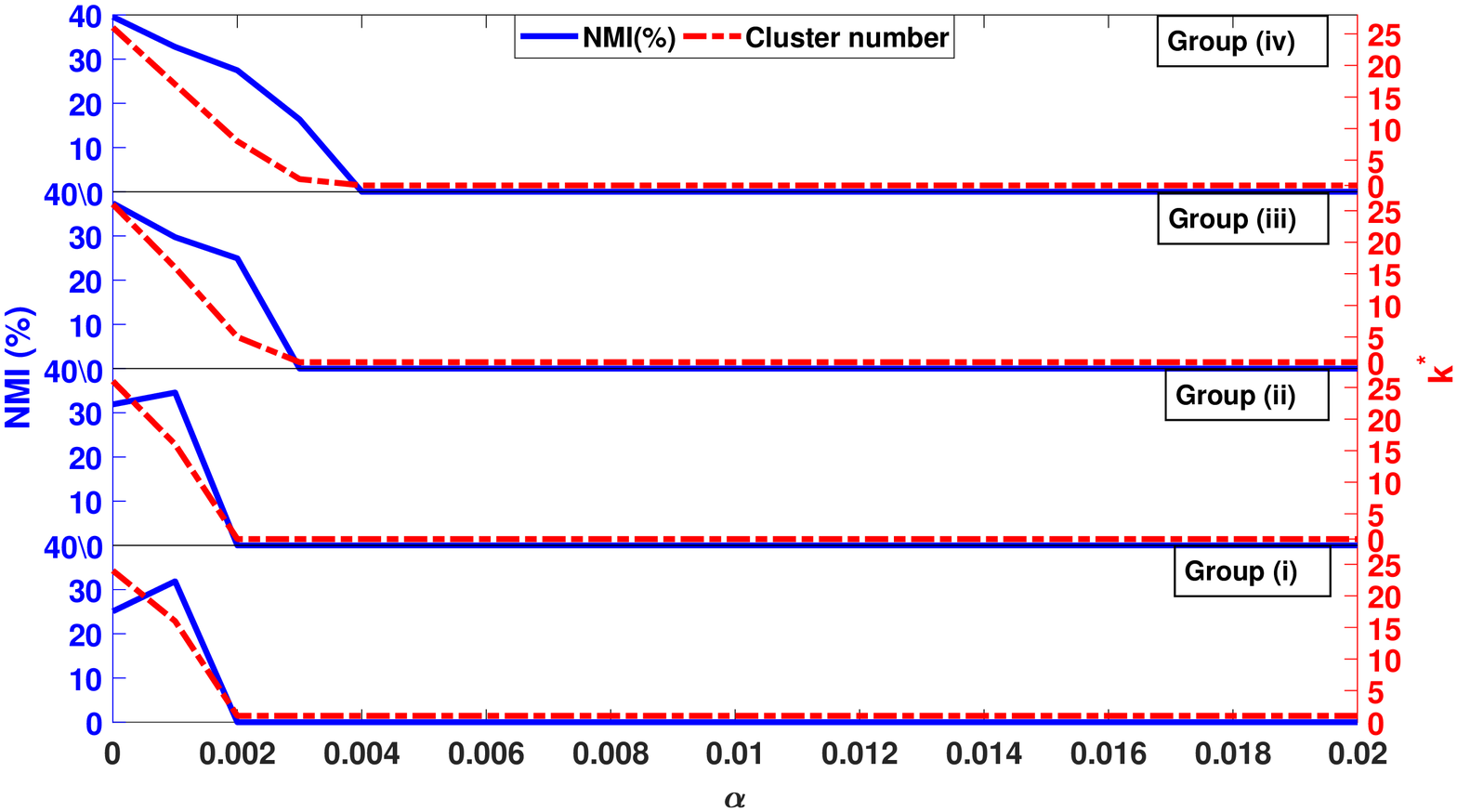}}
\caption{Parameter influence of FDC on the four groups of six benchmark datasets. The parameter $\alpha$ increases along with the horizontal axis, and the left and right vertical axes shows the corresponding NMI (blue solid line) and cluster number $k^*$ (red dash line), respectively.}\label{FigPara}
\end{figure*}

\begin{figure}[htbp]
\centering
\includegraphics[width=0.38\textheight]{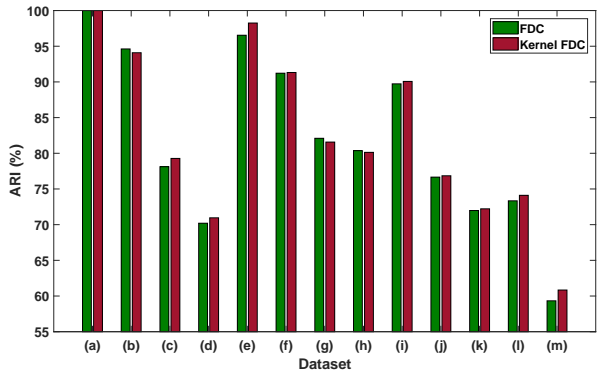}
\caption{Performance of kernel FDC on the benchmark datasets with group (iv) compared with linear FDC.}\label{FigNonlinear}
\end{figure}

\subsection{Benchmark Datasets}
In this subsection, we analyze the hard clustering performance of these models on $19$ benchmark datasets\footnote{\url{http://archive.ics.uci.edu/ml/datasets.php}}. The details of these datasets are shown in Table \ref{Datasets}. For each dataset, we offered four groups of fuzzy pairwise constraints: (i) This group contained $0.05m$ pairs, where the fuzzy pairwise constraints were opposite to the ground truth; (ii) This group contained $0.1m$ pairs, where half of them accorded with the ground truth and the rest were opposite; (iii) This group contained $0.05m$ pairs accorded with the ground truth; (iv) This group contained $0.1m$ pairs accorded with the ground truth. Thereinto, for data (s), the ratio of fuzzy pairwise constraints is a fivefold reduction. Since there are at most $m^2-m$ pairs for a dataset, the number of pairs in each group is much fewer than the maximum. For the models with traditional pairwise constraints (i.e., SSDC, DSC and PCCA), the pairs in these groups were moved into the must-link $\M$ and cannot-link $\C$. For the models with soft/fuzzy pairwise constraints (i.e., FHSS and our FDC), when the pair of samples was from the same class, the fuzzy value of a pair was set to $0.5$ plus a random number with uniform distribution $\mathbf{U}[0,0.5]$ if a sample in the pair is the 10-nearest neighbor of the other; Otherwise, it was set to a random number with $\mathbf{U}[0,1]$. When the pair of samples was from difference classes, the fuzzy value of a pair was set to $-0.5$ minus a random number with $\mathbf{U}[0,0.5]$ if each sample in the pair is not the 10-nearest neighbor of the other; Otherwise, it was set to a random number with $\mathbf{U}[-1,0]$. The random values and opposite pair settings in these groups were used to simulate the practical application. The fuzzy vectors obtained by the fuzzy models, including FCM, PCCA and our FDC, were transformed to labels by \eqref{UtoY}. The average ARI and NMI with the standard deviation for FCM, PCCA and our FDC, and the one-run ARI and NMI for SSDC, DSC and FHSS on these datasets were reported in Table \ref{Benchmark}. The highest ARI and NMI for each dataset were bold compared with the baseline. From Table \ref{Benchmark}, we observe that the SSDC, DSC and FHSS perform under the baseline even though some correct pairwise constraints are given in groups (iii) and (iv). There are a few datasets on which SSDC and FHSS exceed the baseline with correct pairwise constraints, e.g., data (a). However, PCCA and our FDC performs much better than the baselines on many datasets even though some opposite pairwise constraints are given in groups (i) and (ii), e.g., data (h) and (j). On the other datasets, the performance of PCCA and FDC relies on the pairwise constraints. Generally, they works better with correct pairwise constraints than with wrong pairwise constraints. However, the clustering ability of PCCA becomes poor on some large scale datasets, e.g., on data (r) and (s). Correspondingly, our FDC keeps its better performance than PCCA on most of these datasets. In conclusion, the SSDC, DSC and FHSS are not competitive with the other models, and they underuse the (soft) pairwise constraints, especially DSC. Besides, FCM, PCCA and our FDC are a series of fuzzy clustering models, and our FDC can utilize the fuzzy pairwise constraints more sufficiently than the pairwise constrains in PCCA. The wrong pairwise constraints mislead all of these clustering models, and PCCA and FDC are less affected because of their trade-off parameters. Moreover, the fuzzy characteristics of fuzzy pairwise constraint may further reduce the influence from the wrong pairwise constraint, which lead our FDC perform better than the PCCA on groups (i) and (ii) of many datasets.

Statistically, the paired $t$-test was adopted to compare the difference of our FDC and the other models on the benchmark datasets. For each group of the datasets in Table \ref{Benchmark}, the $\bullet/\circ$ indicates FDC is significantly better/worse than compared model at 95\% significance level, and the overall numbers of $\bullet/\circ$ are calculated in the last row. On most of the datasets, our FDC is significantly better than the other models with a large number of $\bullet$ and a low number of $\circ$. Therefore, it is statistical significant that our FDC is better than the other models on the benchmark datasets, which supports the previous observation.

\begin{table*}[htbp]
\caption{Clustering performance on the JAFFE database with five groups of fuzzy pairwise constraints} \centering
\begin{tabular}{llrrrrrrr}
\hline
 Group&Criterion&kmeans~~&FCM~~&SSDC~~&DSC~~& FHSS~~&PCCA~~&FDC\\\hline
&ARI(\%) &50.45$\pm$0.48~~ &50.86$\pm$0.38~~  &50.46$\bullet$ &50.45$\bullet$ & 49.60$\bullet$&49.97$\pm$0.18$\bullet$ &$\mathbf{51.72}\pm$0.50\\
(i)&NMI(\%) &7.32$\pm$1.70~~ & 5.17$\pm$1.17~~ & 5.97~~& 5.52~~& 6.73~~&3.25$\pm$1.07$\bullet$ &$\mathbf{8.27}\pm$2.64\\
&Acc.(\%)&50.96$\pm$10.21$\bullet$ &67.83$\pm$0.62~~ & 64.25$\bullet$& 26.62$\bullet$& 22.57$\bullet$& 52.54$\pm$3.11$\bullet$&$\mathbf{68.16}\pm$1.10\\\hline
&ARI(\%) &50.45$\pm$0.48$\bullet$ & 50.86$\pm$0.38$\bullet$ &51.07$\bullet$ & 50.63$\bullet$& 49.78$\bullet$&50.86$\pm$0.01$\bullet$ &$\mathbf{54.04}\pm$1.11\\
(ii)&NMI(\%) &7.32$\pm$1.70~~ &5.17$\pm$1.17$\bullet$  & 6.62$\bullet$& 6.74$\bullet$& 7.59~~& 5.65$\pm$0.63$\bullet$&$\mathbf{10.63}\pm$2.80\\
&Acc.(\%)& 50.96$\pm$10.21$\bullet$ & 67.83$\pm$0.62$\bullet$ & 58.50$\bullet$&28.05$\bullet$ & 22.92$\bullet$& 53.19$\pm$0.50$\bullet$&$\mathbf{70.36}\pm$0.77\\\hline
&ARI(\%) &50.45$\pm$0.48$\bullet$ & 50.86$\pm$0.38$\bullet$ &51.58$\bullet$ & 50.63$\bullet$& 49.92$\bullet$& 51.43$\pm$0.12$\bullet$&$\mathbf{54.05}\pm$0.85\\
(iii)&NMI(\%) & 7.32$\pm$1.70$\bullet$&5.17$\pm$1.17$\bullet$  & 6.89$\bullet$& 6.74$\bullet$& 8.23$\bullet$& 6.46$\pm$1.13$\bullet$&$\mathbf{11.76}\pm$1.30\\
&Acc.(\%)& 50.96$\pm$10.21$\bullet$ &67.83$\pm$0.62$\bullet$ &59.94$\bullet$ &28.05$\bullet$ &23.17$\bullet$&53.26$\pm$0.58$\bullet$ &$\mathbf{70.57}\pm$0.76\\\hline
&ARI(\%) &50.45$\pm$0.48$\bullet$ &50.86$\pm$0.38$\bullet$  &51.58$\bullet$ & 50.63$\bullet$&50.00$\bullet$ &51.96$\pm$0.61$\bullet$ &$\mathbf{55.74}\pm$1.14\\
(iv)&NMI(\%) &7.32$\pm$1.70$\bullet$ & 5.17$\pm$1.17$\bullet$ & 6.89$\bullet$& 6.74$\bullet$& 8.41$\bullet$& 7.67$\pm$2.22$\bullet$&$\mathbf{12.26}\pm$2.39\\
&Acc.(\%)& 50.96$\pm$10.21$\bullet$ & 67.83$\pm$0.62$\bullet$&59.94$\bullet$& 28.05$\bullet$& 23.44$\bullet$&54.68$\pm$3.71$\bullet$ &$\mathbf{71.96}\pm$1.20\\\hline
&ARI(\%) &50.45$\pm$0.48$\bullet$ & 50.86$\pm$0.38$\bullet$ &51.64$\bullet$ & 50.77$\bullet$&50.17$\bullet$ &52.60$\pm$0.39$\bullet$ &$\mathbf{56.61}\pm$1.07\\
(v)&NMI(\%) & 7.32$\pm$1.70$\bullet$&5.17$\pm$1.17$\bullet$  & 10.53$\bullet$& 7.81$\bullet$&10.51$\bullet$ &9.25$\pm$1.36$\bullet$ &$\mathbf{13.67}\pm$2.77\\
&Acc.(\%)&50.96$\pm$10.21$\bullet$  & 67.83$\pm$0.62$\bullet$&54.06$\bullet$& 28.82$\bullet$& 27.91$\bullet$& 56.04$\pm$3.57$\bullet$&$\mathbf{73.37}$$\pm$0.67\\\hline
\end{tabular}\label{JAFFEall}
\\$\bullet/\circ$ indicates FDC is significantly better/worse than compared model (paired t-tests at 95\% significance level).
\end{table*}

Subsequently, we analyze the influence of the parameters $\alpha$ and $\beta$ in the FDC. The purpose of $\beta$ is clear: larger $\beta$ indicates that we count on the fuzzy pairwise constraints more, and vice versa. Thus, the precision of fuzzy pairwise constraint directs the performance of FDC. The other parameter $\alpha$ can adjust the cluster number in theory. To verify the influence of $\alpha$, we augmented $\alpha$ from $0$ on the above four groups of six benchmark datasets with larger cluster numbers. The clustering results and the cluster numbers were reported in Fig. \ref{FigPara}, where the cluster number was decided by \eqref{UtoY}. It is obvious that the cluster number decreases with the increasing $\alpha$ generally, and there is a threshold smaller than 1 such that the cluster number would be fixed to 1 when $\alpha$ is larger than this threshold. Notice that the cluster number may increase with a larger $\alpha$, e.g., in group (iii), Fig. \ref{FigPara} (i), for the cluster prototypes and cluster numbers may be inconsistent. Though we set $k$ to be the truth from Table \eqref{Datasets}, smaller clusters would be obtained with $\alpha=0$, e.g., on data (i). However, more clusters were formulated with nonzero $\alpha$, and then better performance was obtained. In practice, we should adjust $\alpha$ carefully for different datasets, and its upper bound would be much smaller than 1, especially for large scale datasets.

Finally, we tested the FDC for various metric spaces. Generally, an appropriate metric can promote the performance of FDC, and the metric changes with the data space, which leads to the metric learning problems \cite{MetricLP,MetricLP2}. For fairness, we implemented the kernel FDC on these benchmark datasets with group (iv) compared with the linear FDC, where
the Gaussian kernel \cite{Kernel2} $K(\bfx_1,\bfx_2)=\exp\{-\mu||\bfx_1-\bfx_2||^2\}$ was used and its parameter $\mu$ was selected from $\{2^i |i = -10,-9,\ldots,5\}$. The comparisons were depicted in Fig. \ref{FigNonlinear}. Apparently, the Gaussian kernel can further improve the performance of FDC on most of the datasets. Once an appropriate metric is decided for a data space, we can apply it into FDC without any difficulty.

\begin{figure*}[htbp]
\centering
\subfigure[Group (i)]{\includegraphics[width=0.24\textheight]{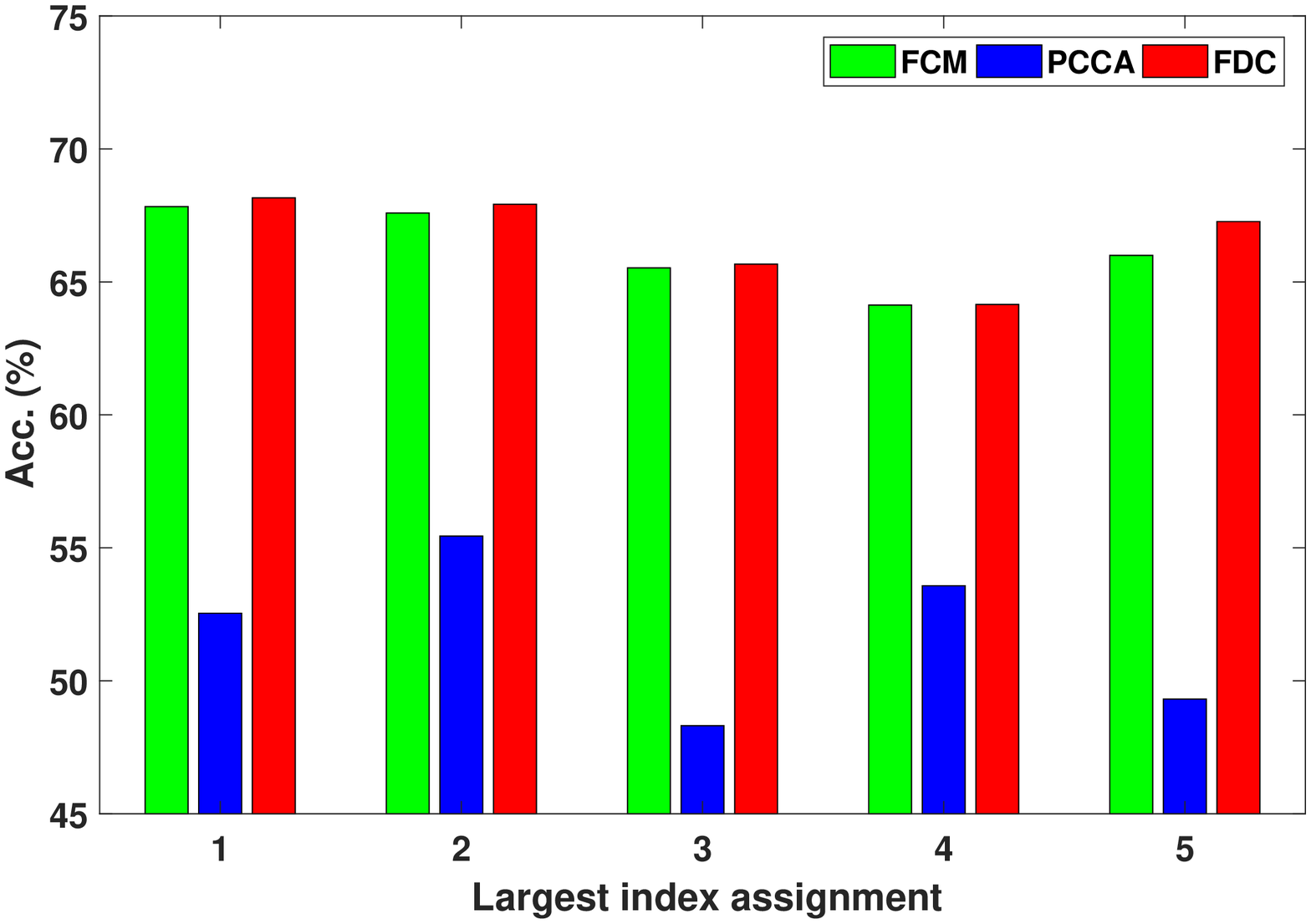}}
\subfigure[Group (ii)]{\includegraphics[width=0.24\textheight]{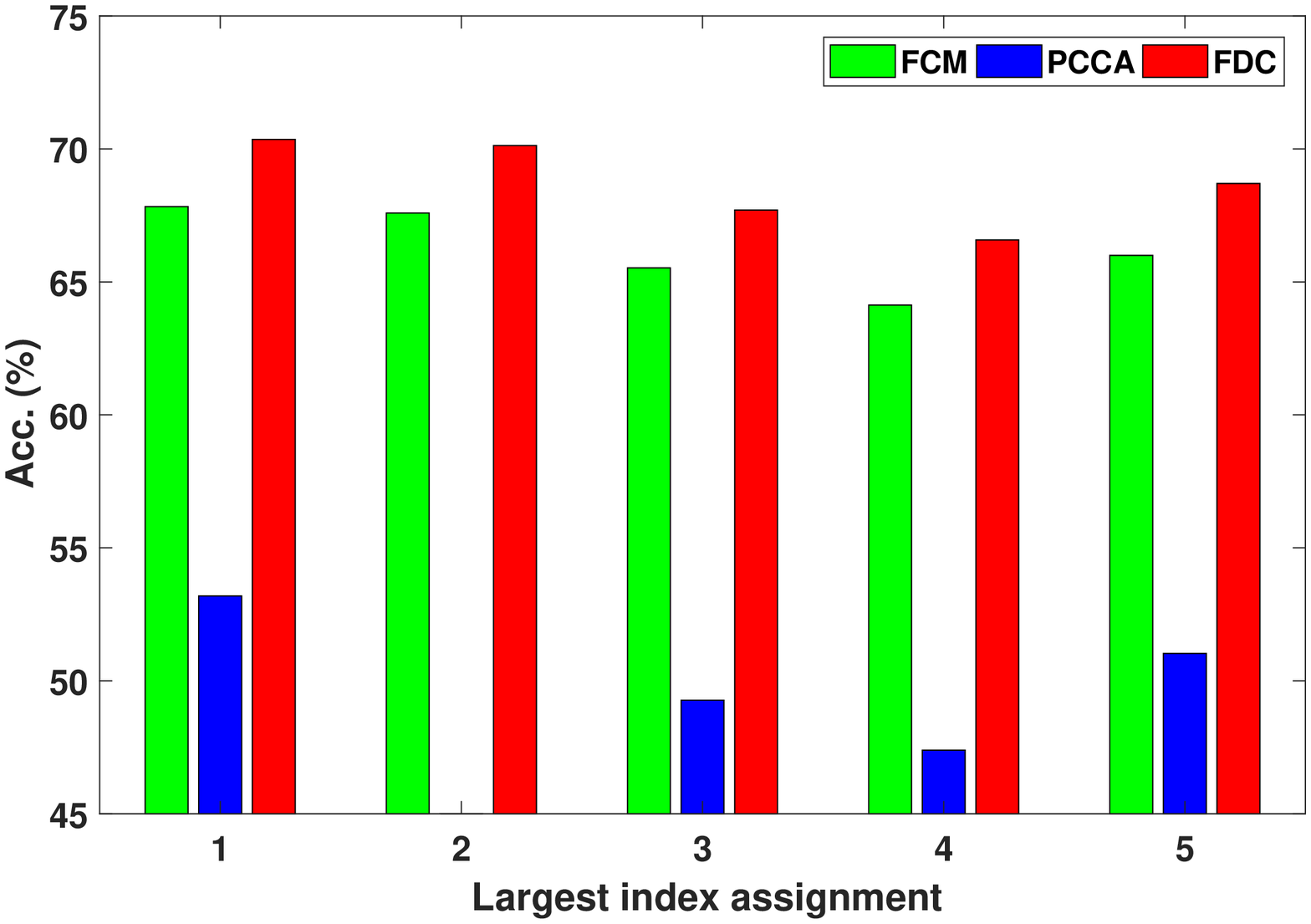}}
\subfigure[Group (iii)]{\includegraphics[width=0.24\textheight]{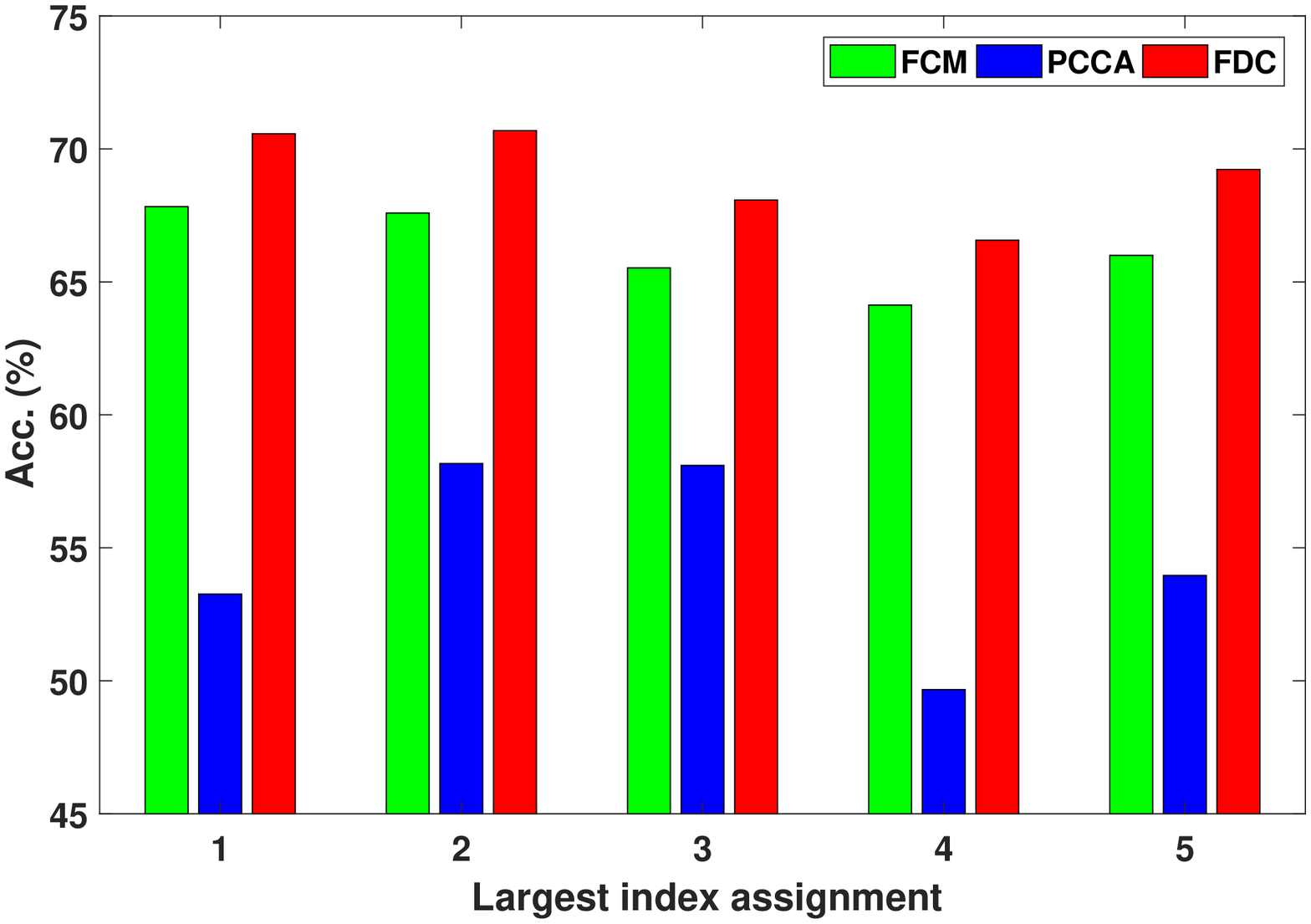}}
\subfigure[Group (iv)]{\includegraphics[width=0.24\textheight]{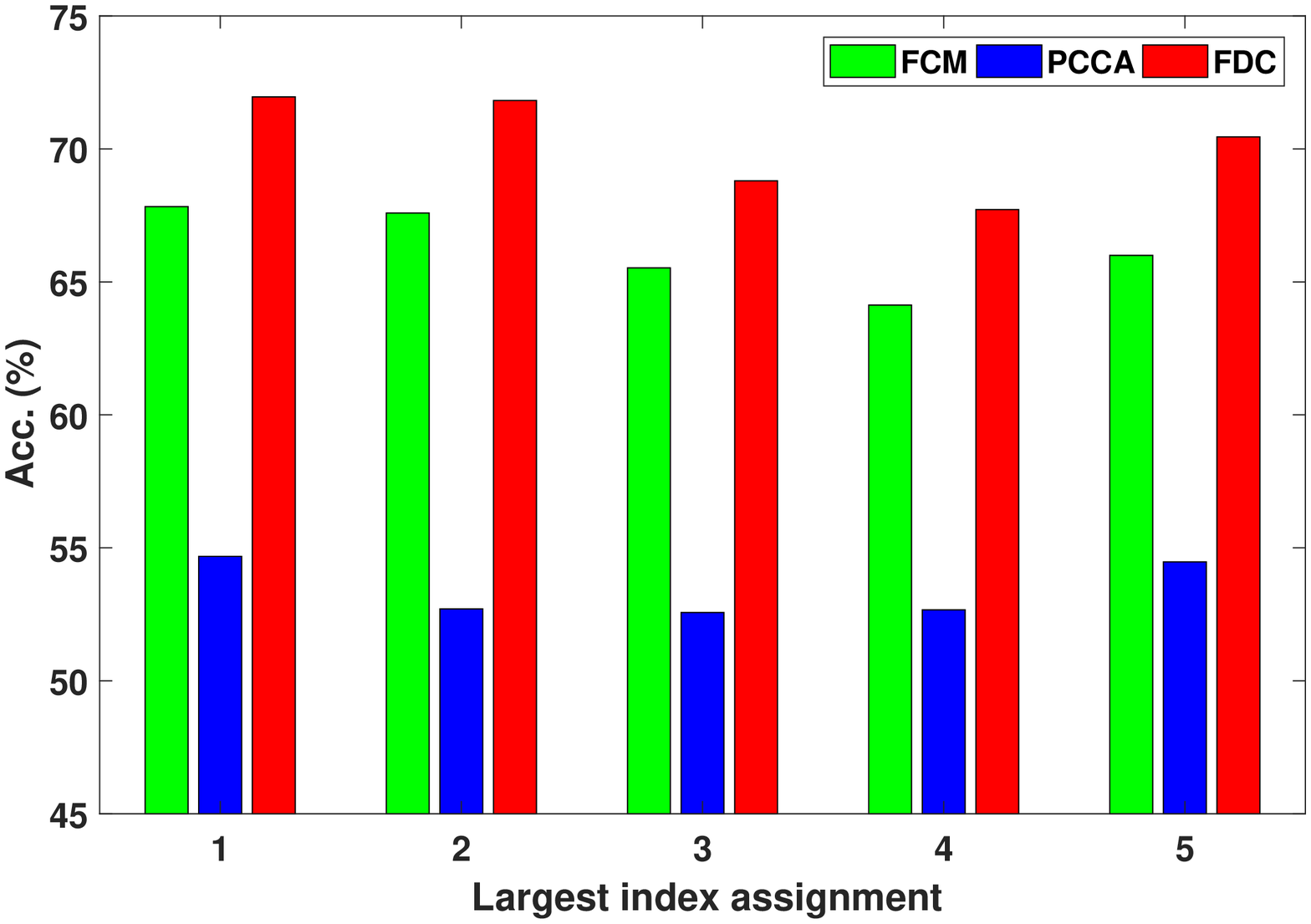}}
\subfigure[Group (v)]{\includegraphics[width=0.24\textheight]{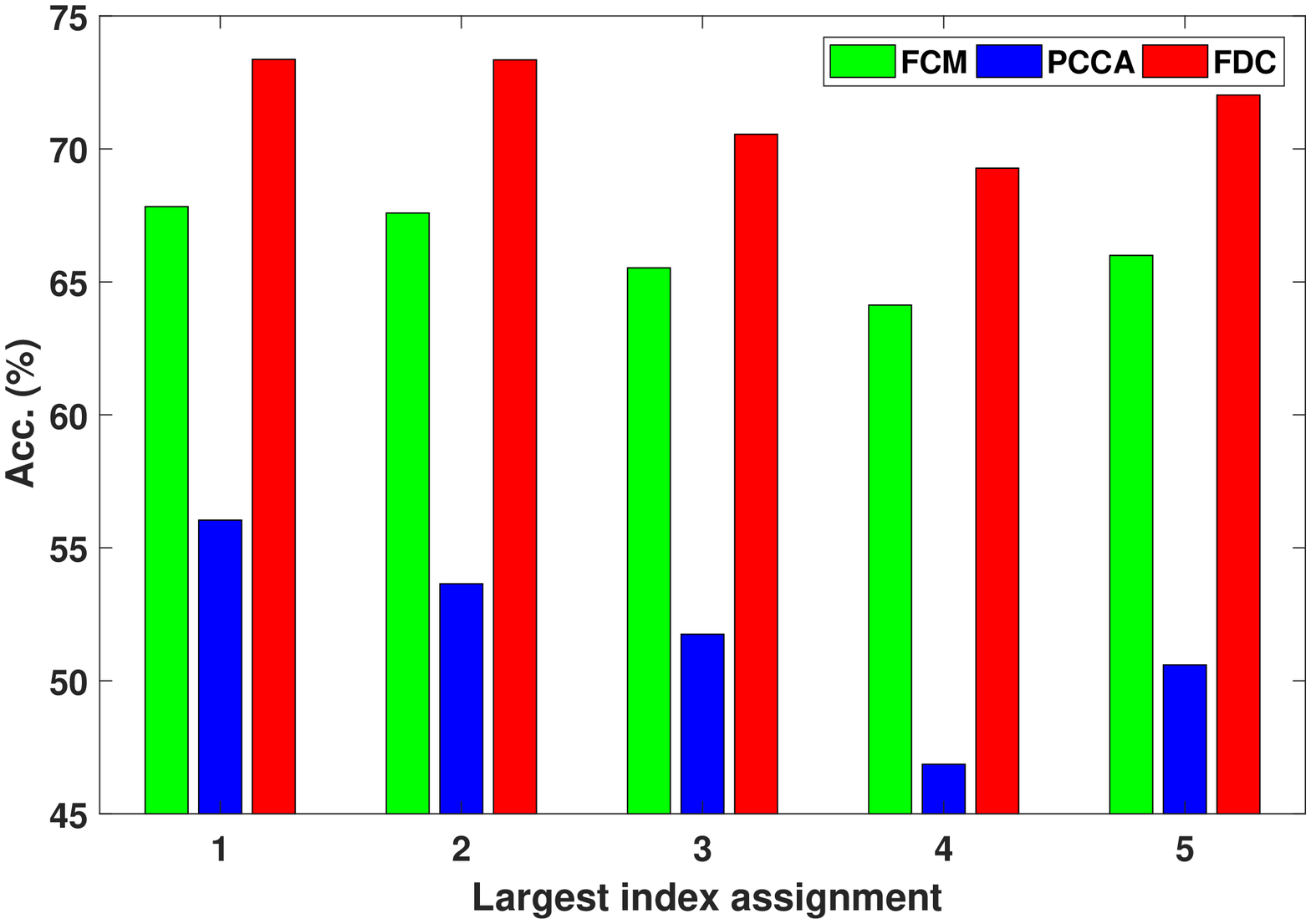}}
\caption{LIA-based Acc. of the fuzzy clustering models on JAFFE with the five groups of fuzzy pairwise constraints.}\label{FigLIA}
\end{figure*}

\begin{table*}[htbp]
\caption{Examples of the fuzzy vectors by the fuzzy clustering models on JAFFE} \centering
\begin{tabular}{lccccc}
\hline
ID&Image& Ground truth&FCM&PCCA&FDC\\\hline
45&\includegraphics[width=0.076\textheight]{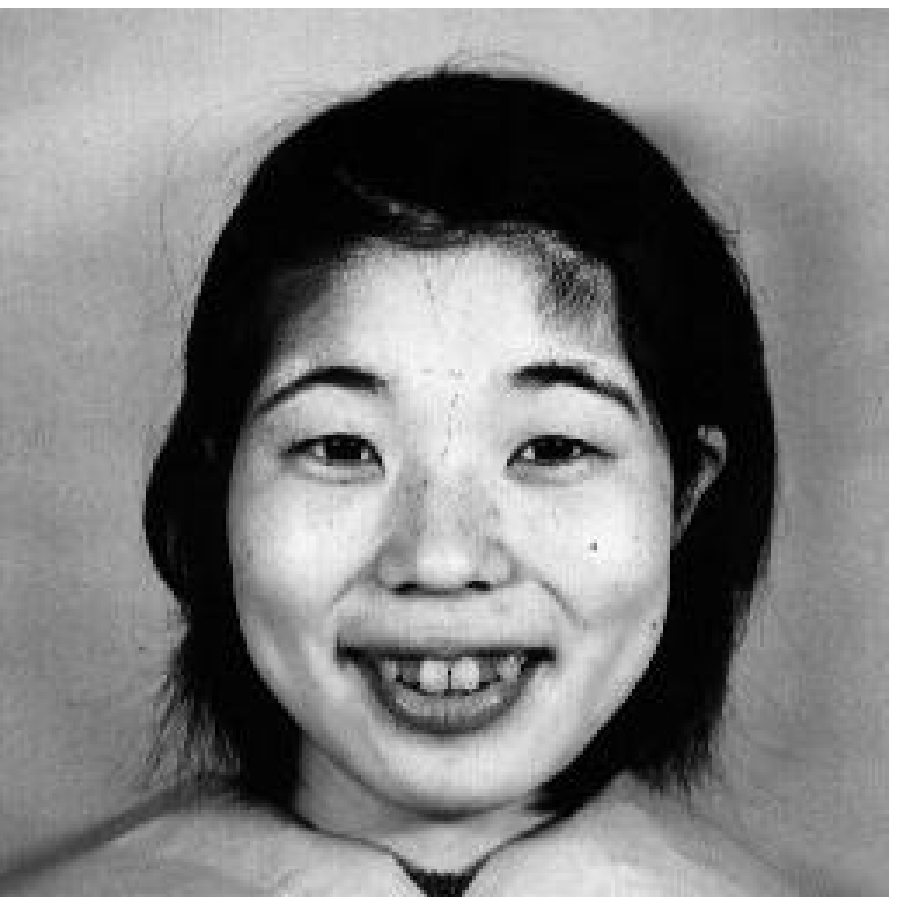}&
\includegraphics[width=0.12\textheight,height=0.08\textheight]{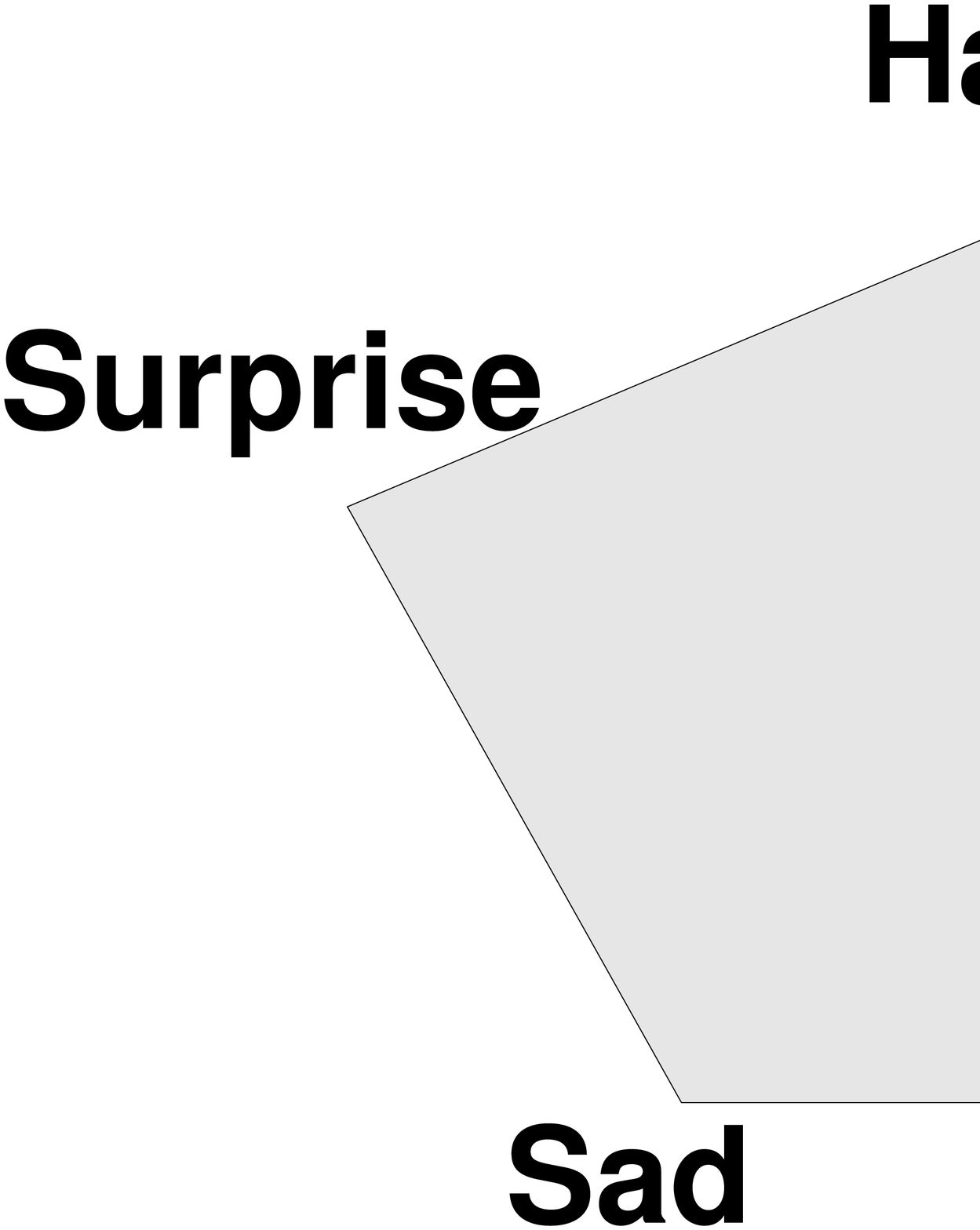}&
\includegraphics[width=0.12\textheight,height=0.08\textheight]{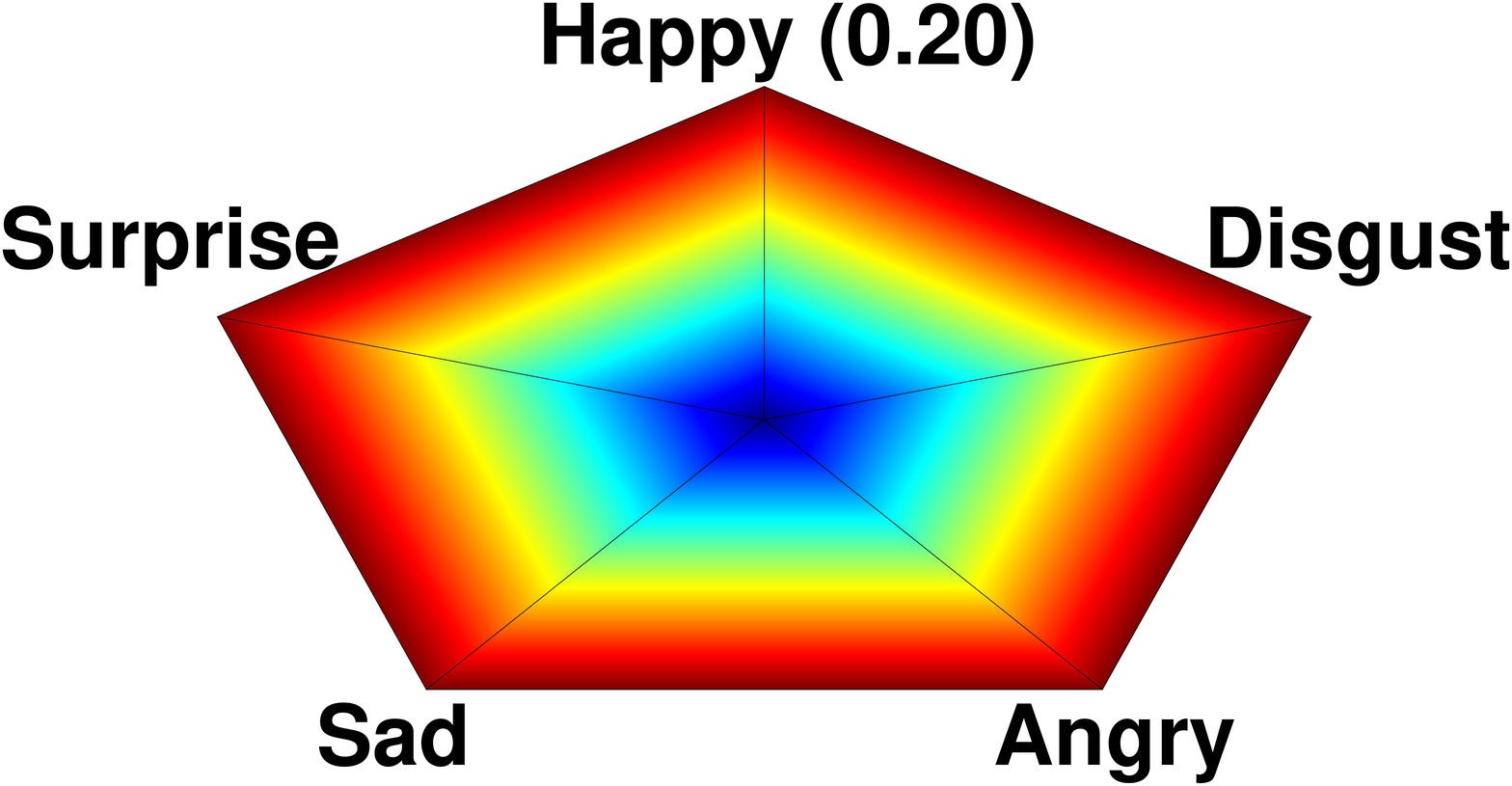}&
\includegraphics[width=0.12\textheight,height=0.08\textheight]{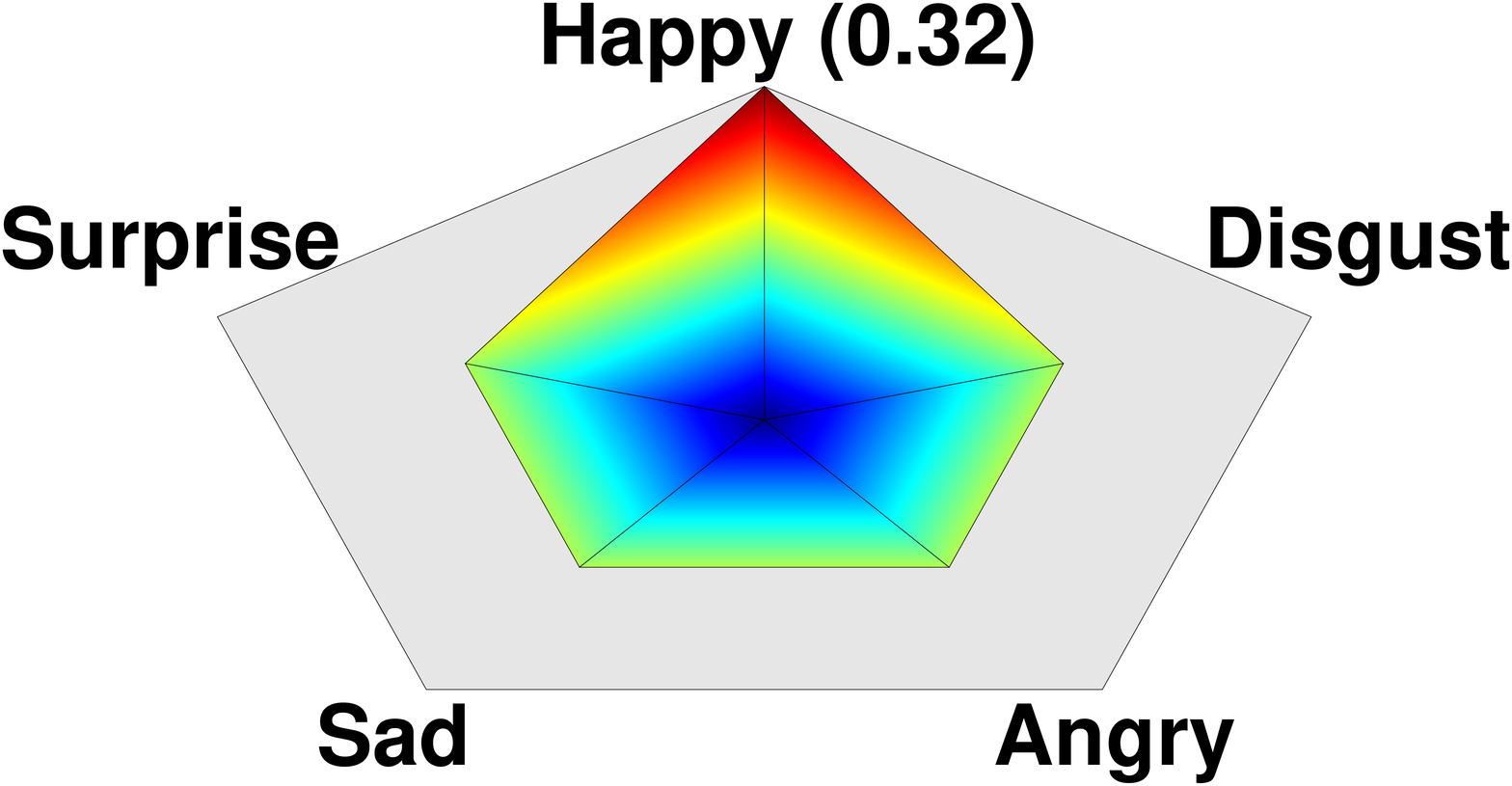}&
\includegraphics[width=0.12\textheight,height=0.08\textheight]{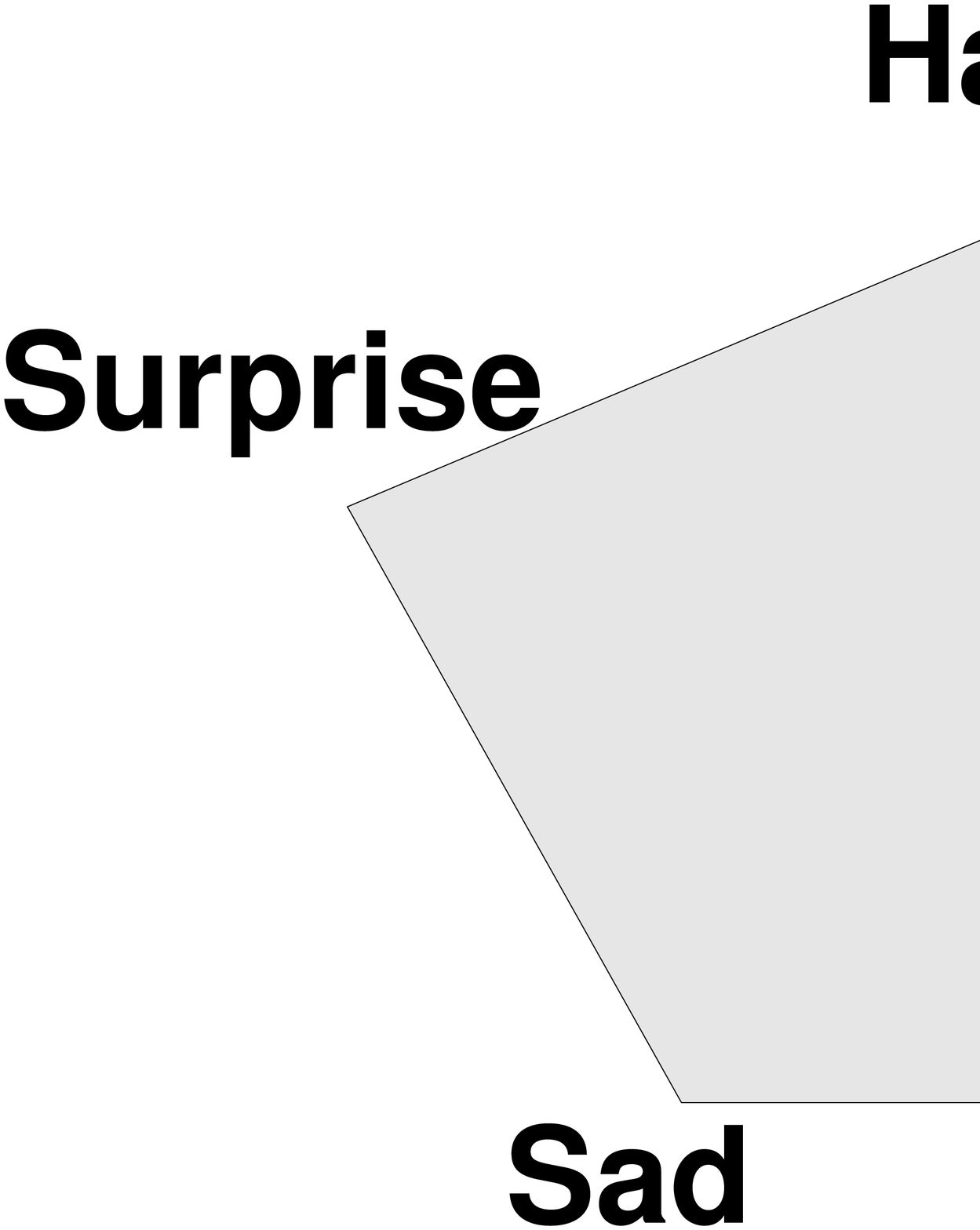} \\\hline
56&\includegraphics[width=0.076\textheight]{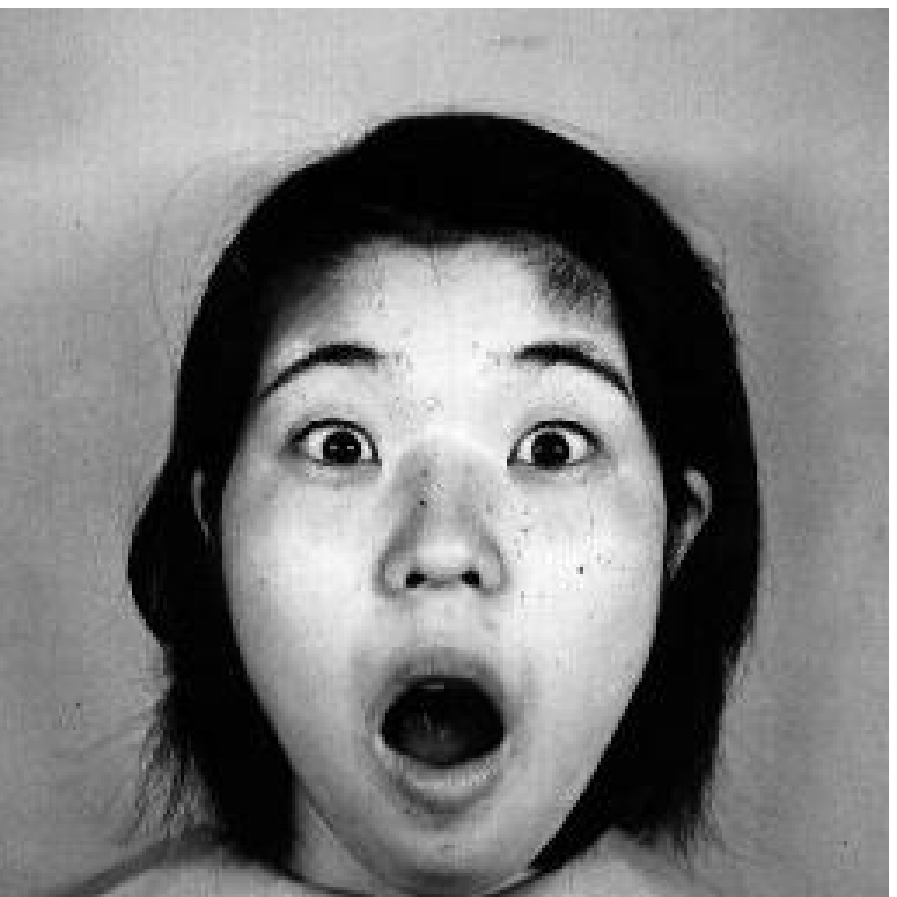}&
\includegraphics[width=0.12\textheight,height=0.08\textheight]{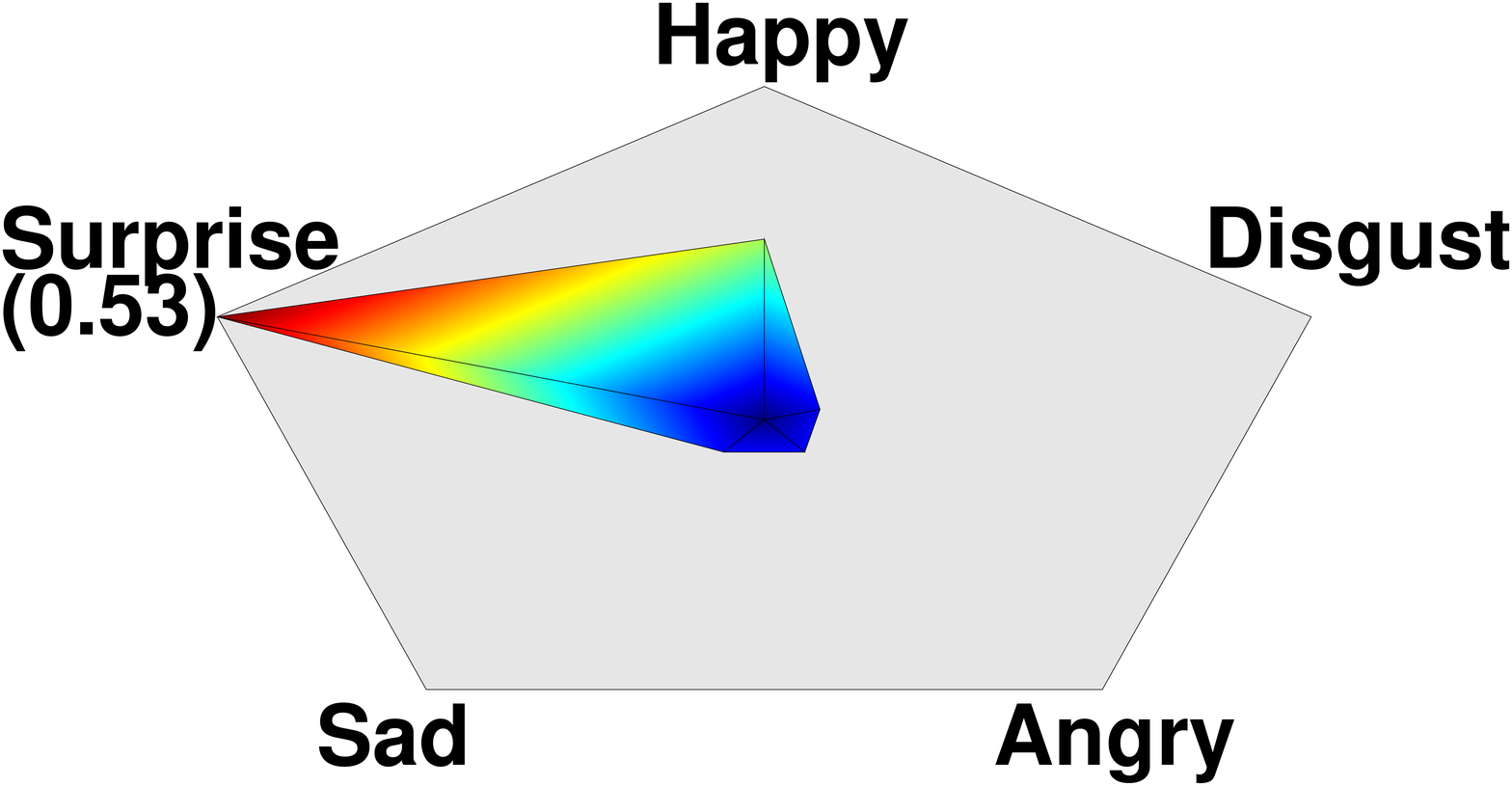}&
\includegraphics[width=0.12\textheight,height=0.08\textheight]{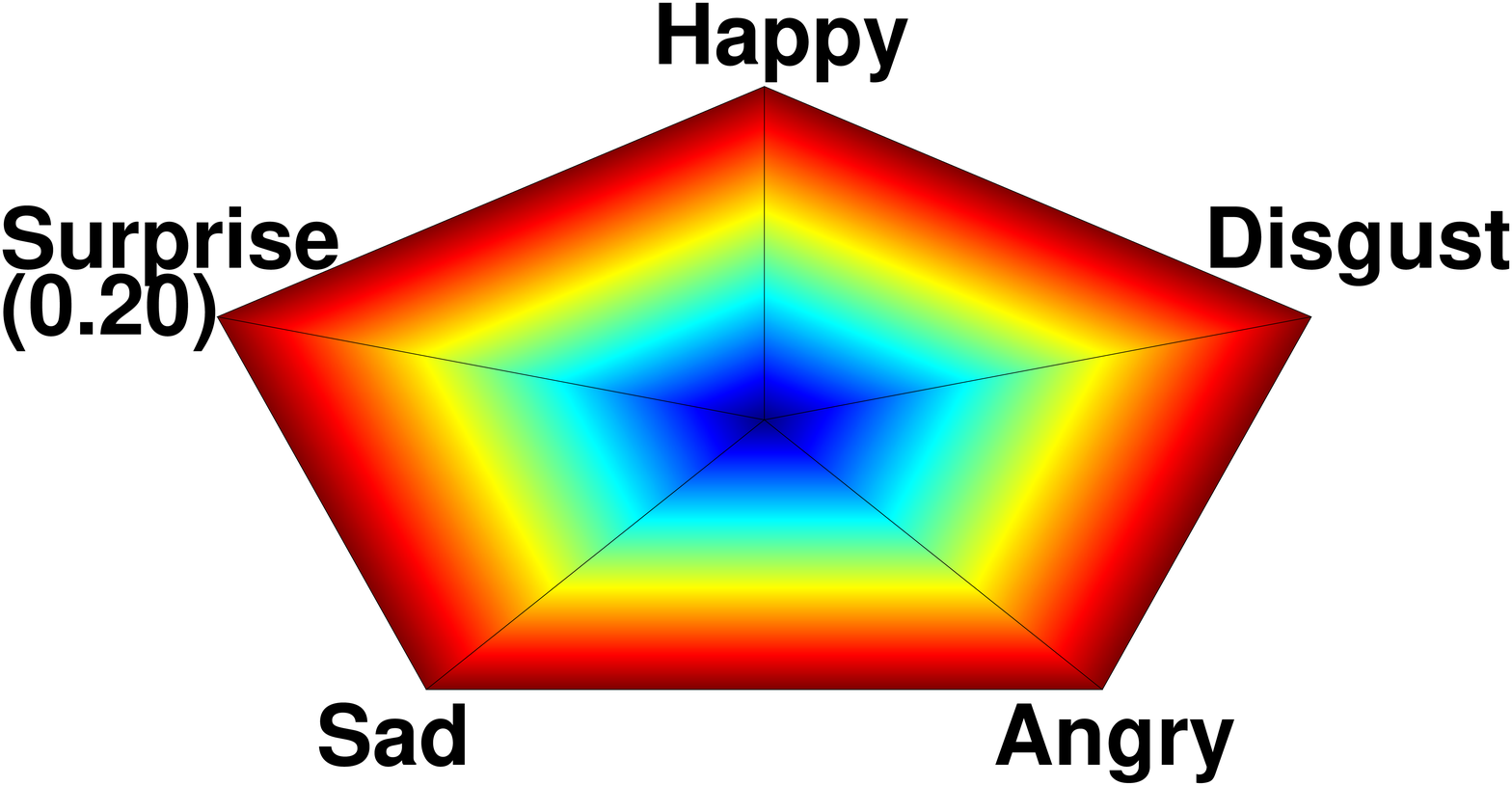}& \includegraphics[width=0.12\textheight,height=0.08\textheight]{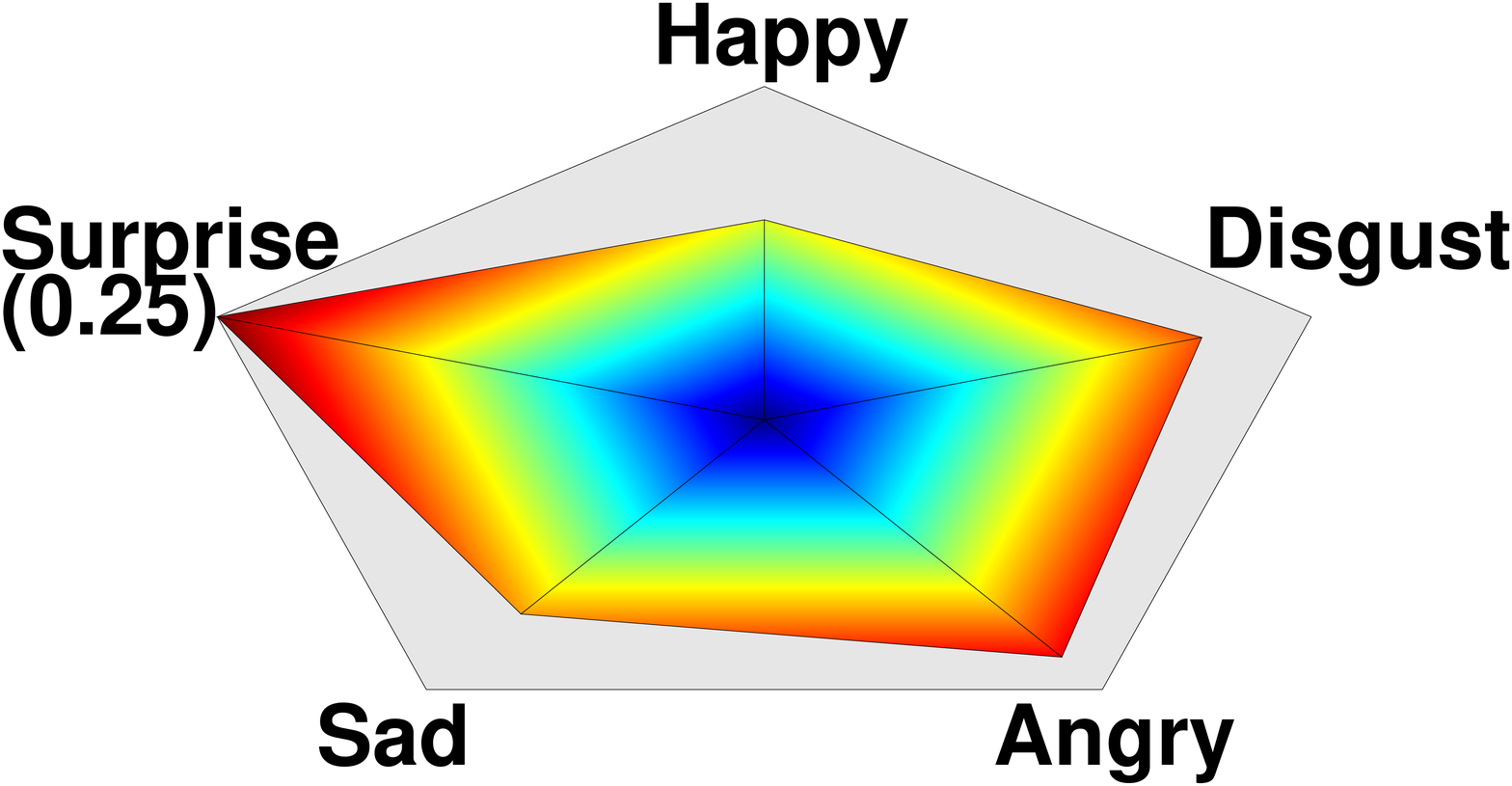}&
\includegraphics[width=0.12\textheight,height=0.08\textheight]{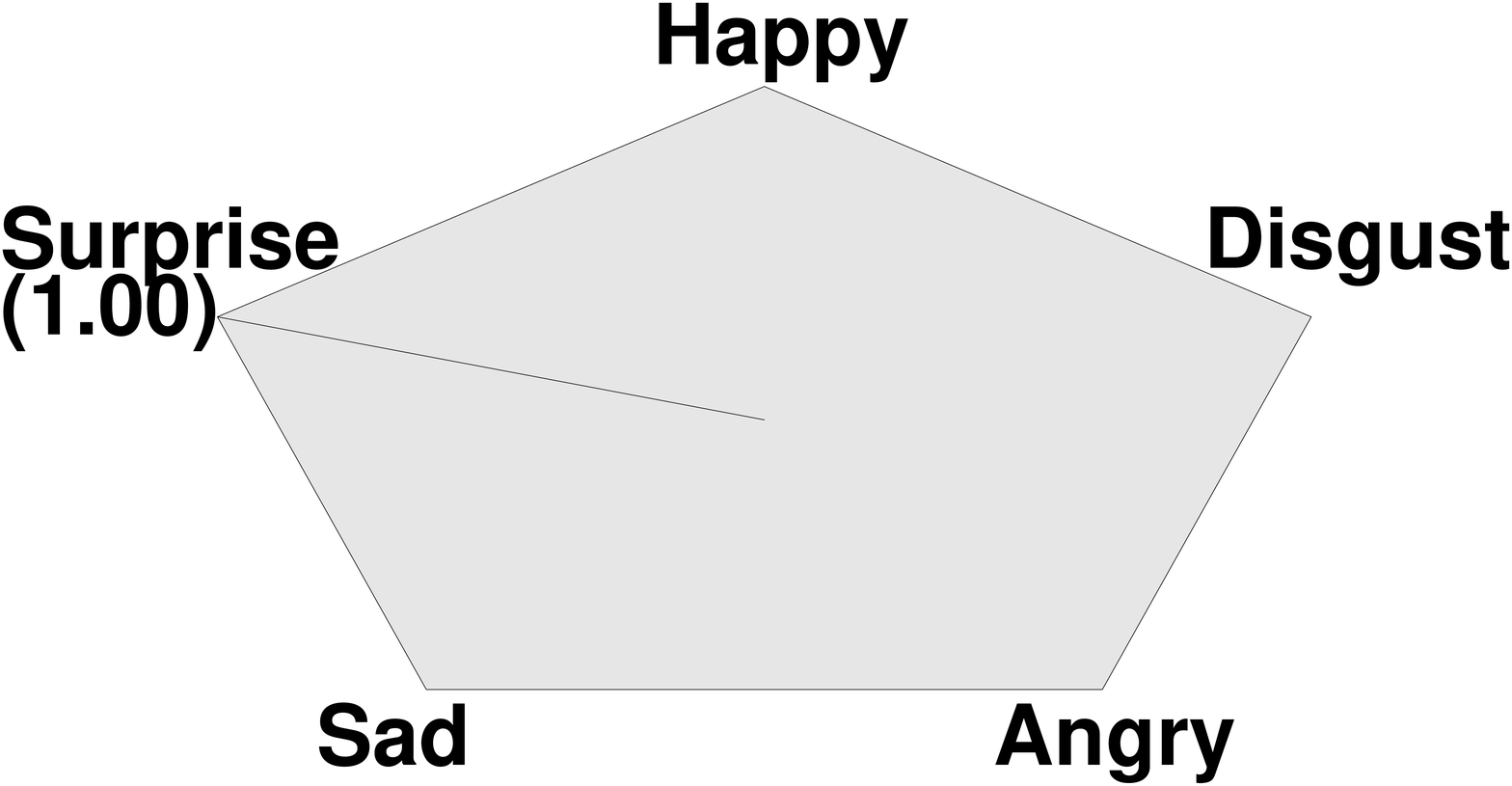} \\\hline
75&\includegraphics[width=0.076\textheight]{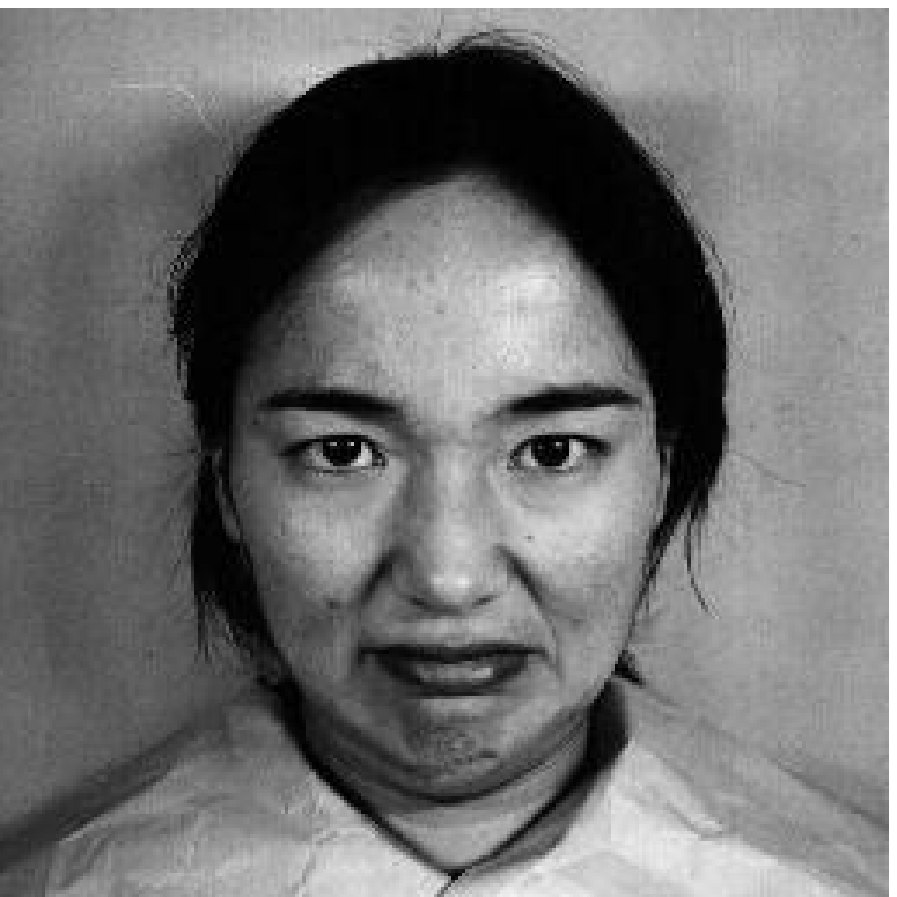}&
\includegraphics[width=0.12\textheight,height=0.08\textheight]{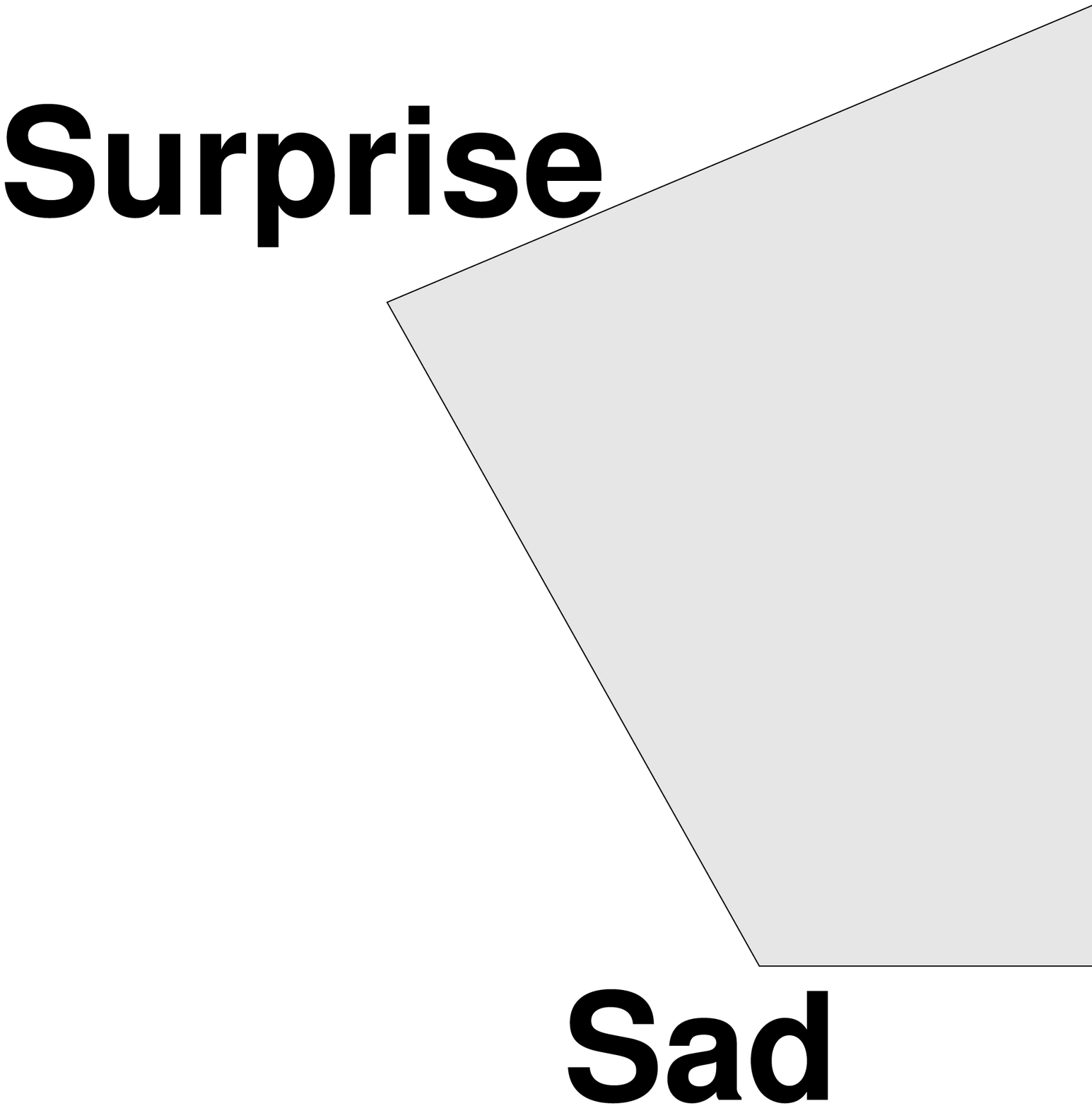}&
\includegraphics[width=0.12\textheight,height=0.08\textheight]{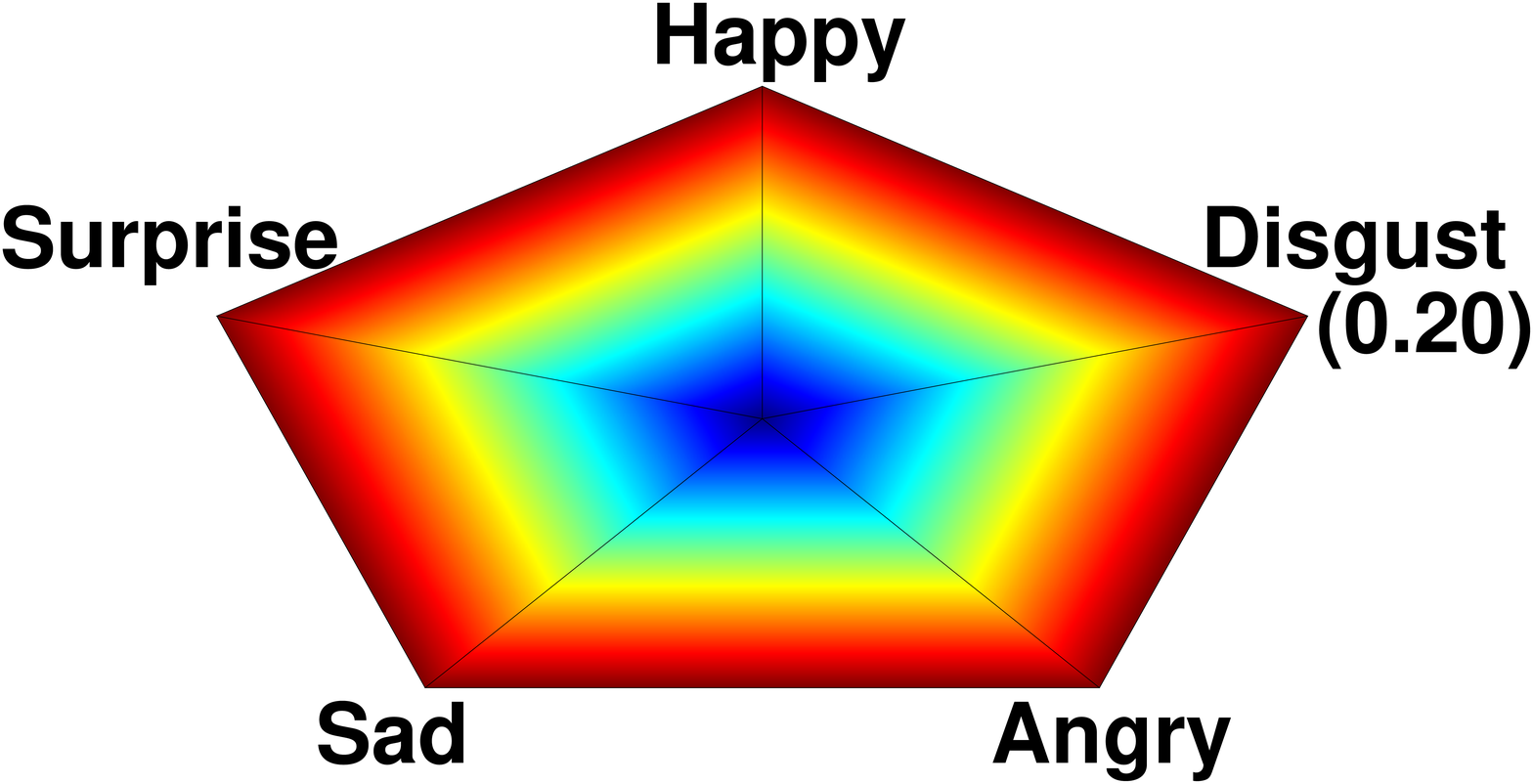}&
\includegraphics[width=0.12\textheight,height=0.08\textheight]{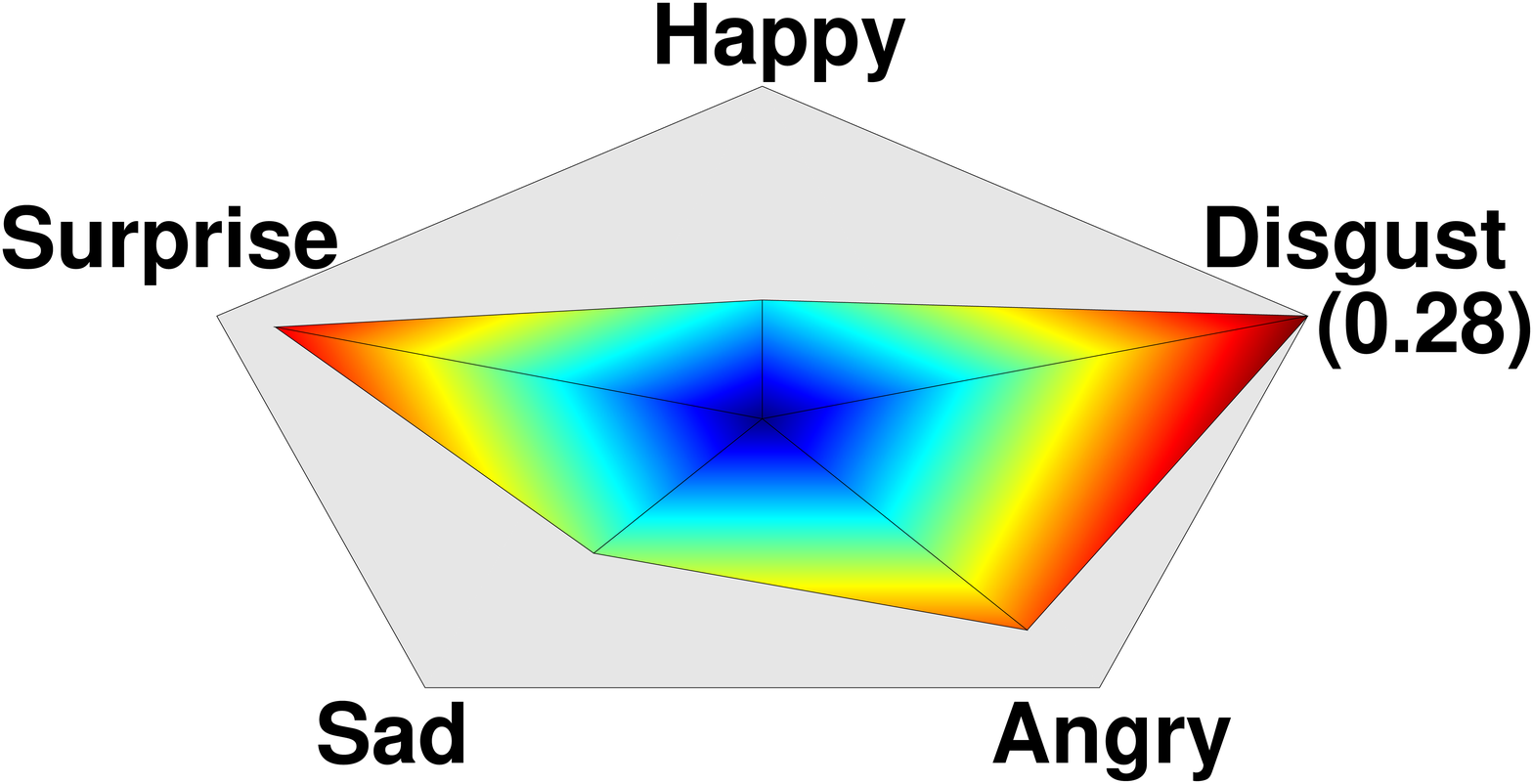}&
\includegraphics[width=0.12\textheight,height=0.08\textheight]{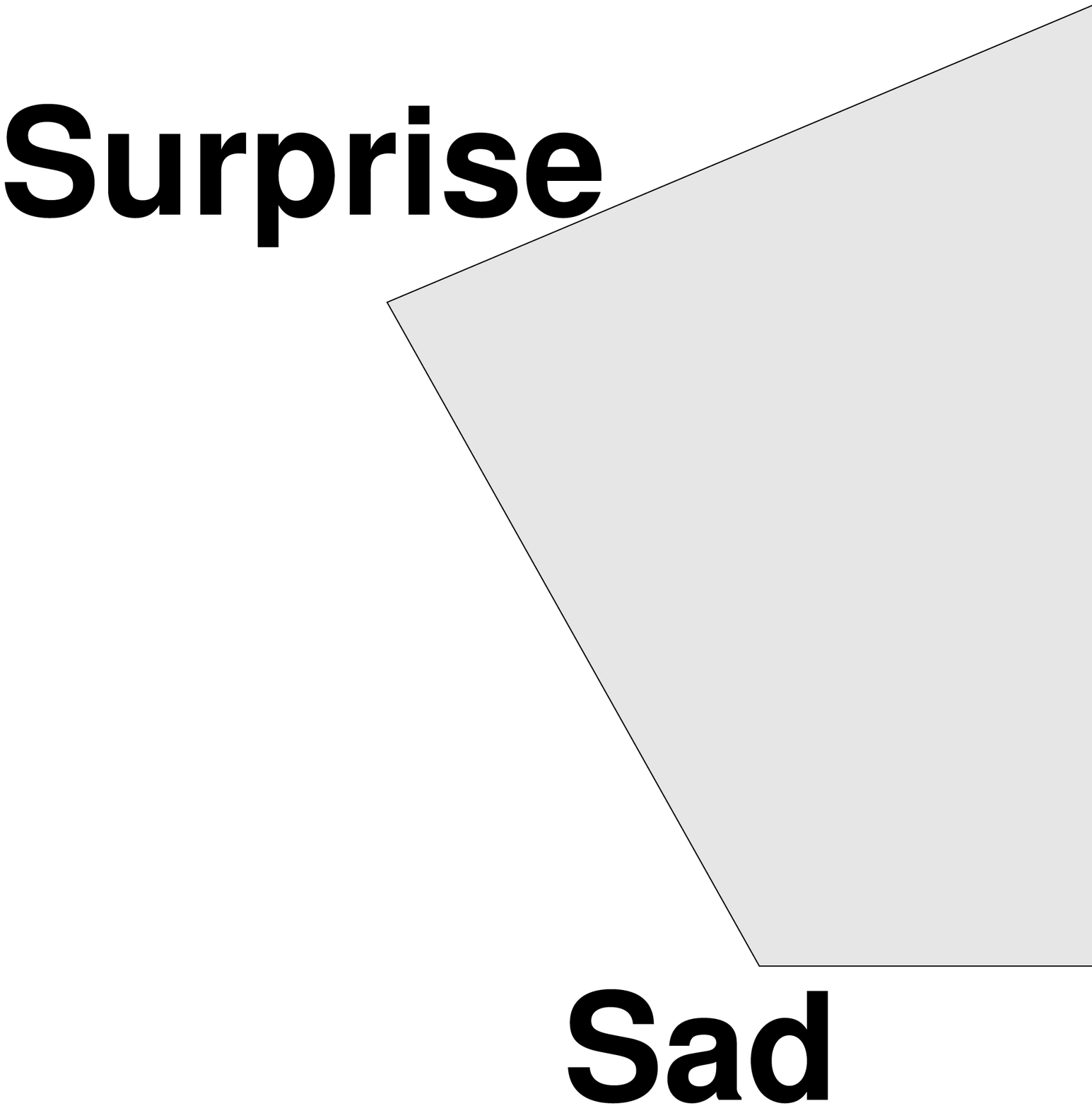} \\\hline
180&\includegraphics[width=0.076\textheight]{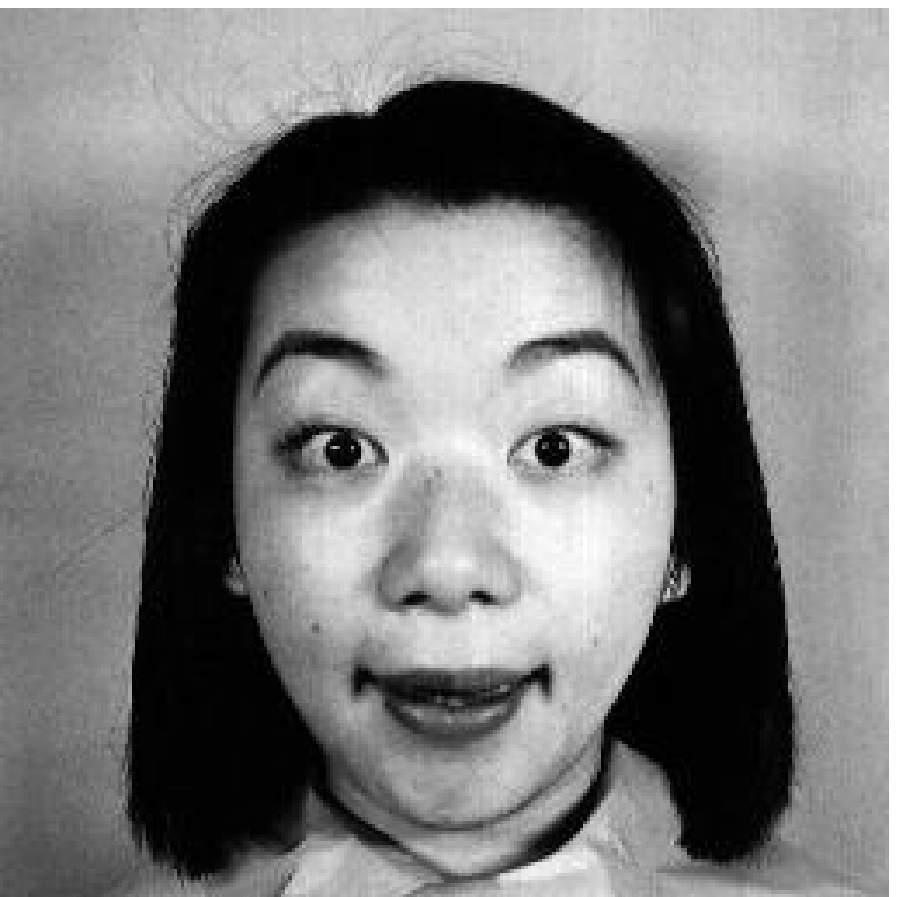}&
\includegraphics[width=0.12\textheight,height=0.08\textheight]{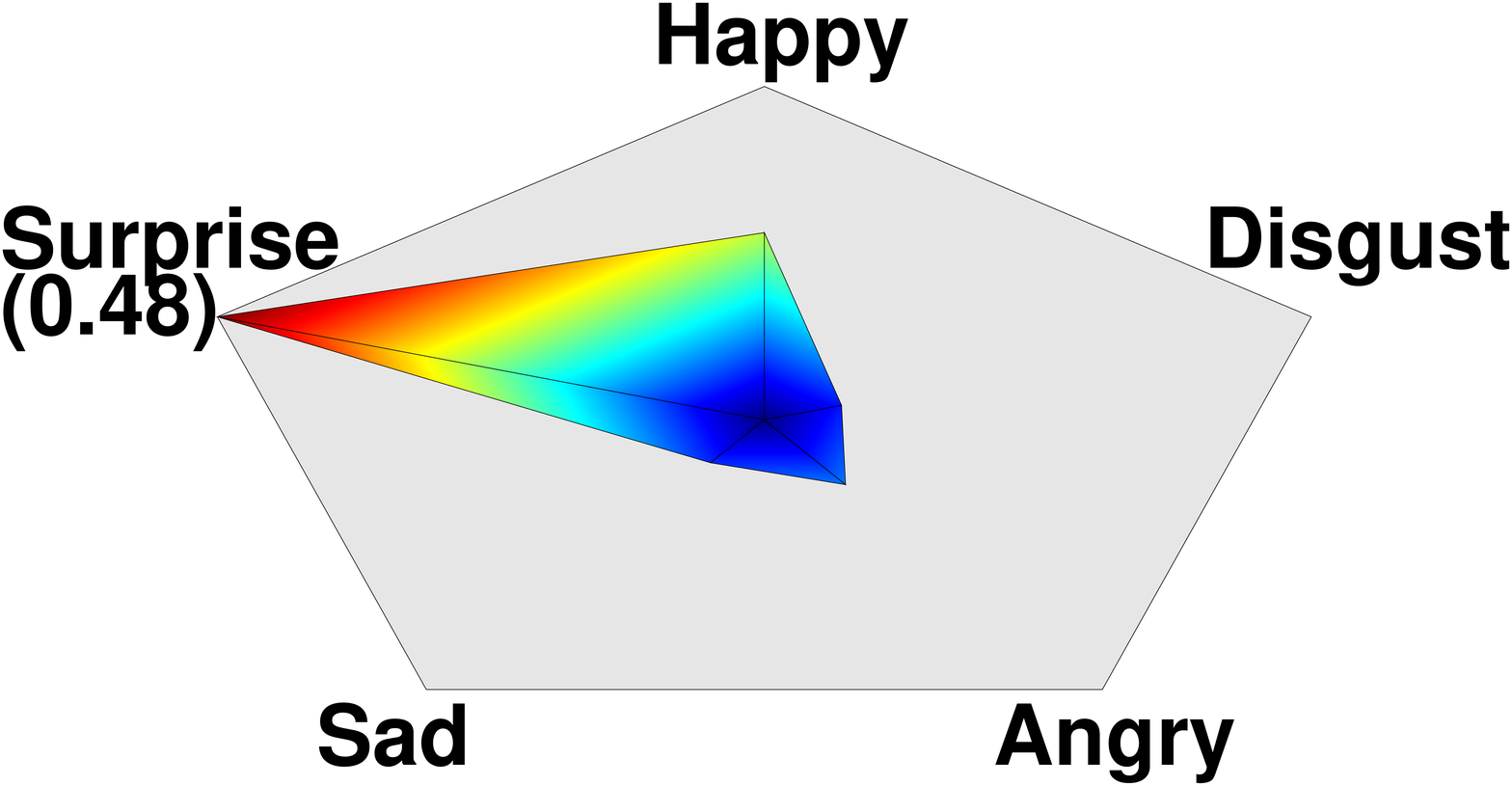}&
\includegraphics[width=0.12\textheight,height=0.08\textheight]{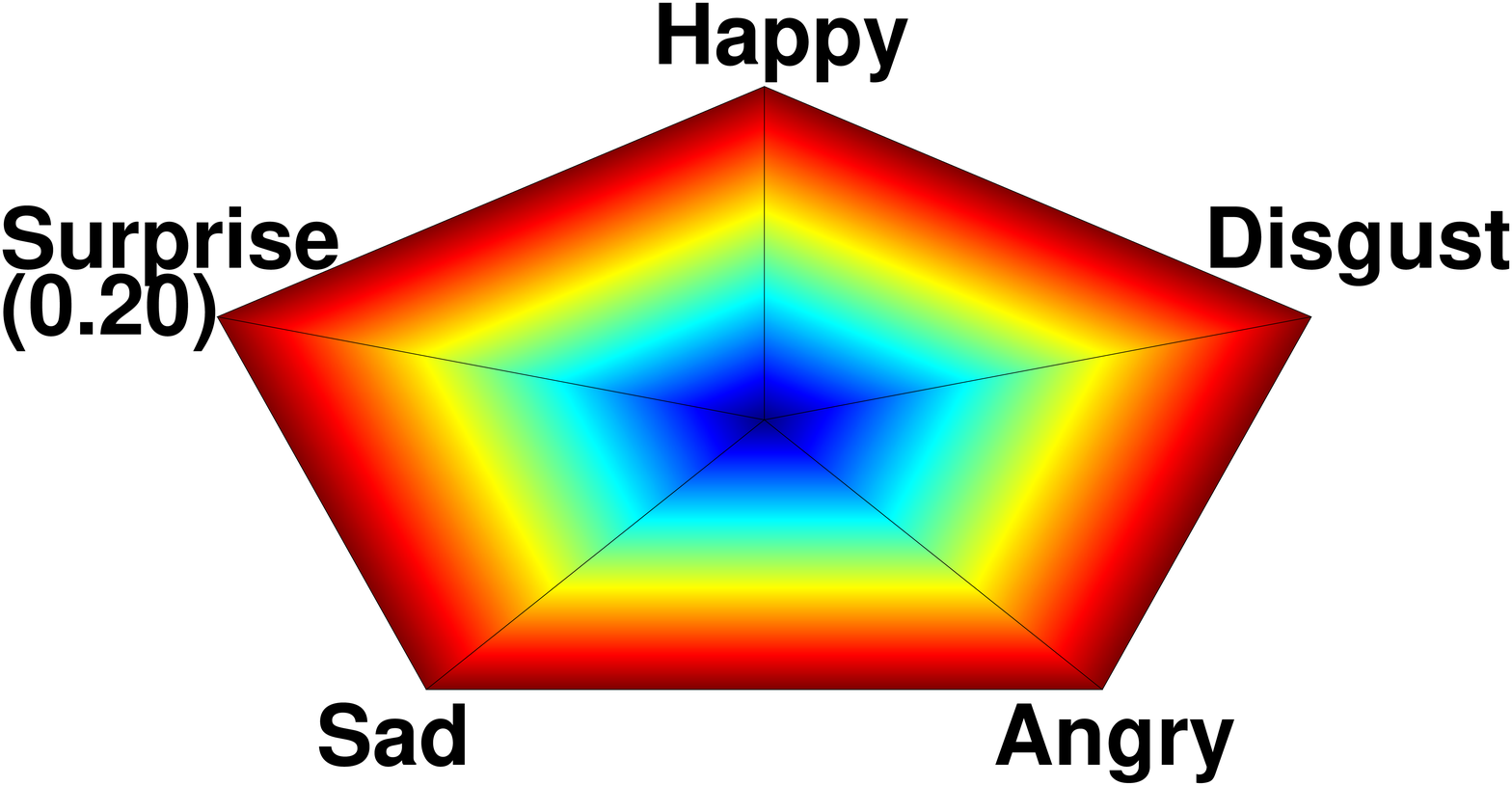}&
\includegraphics[width=0.12\textheight,height=0.08\textheight]{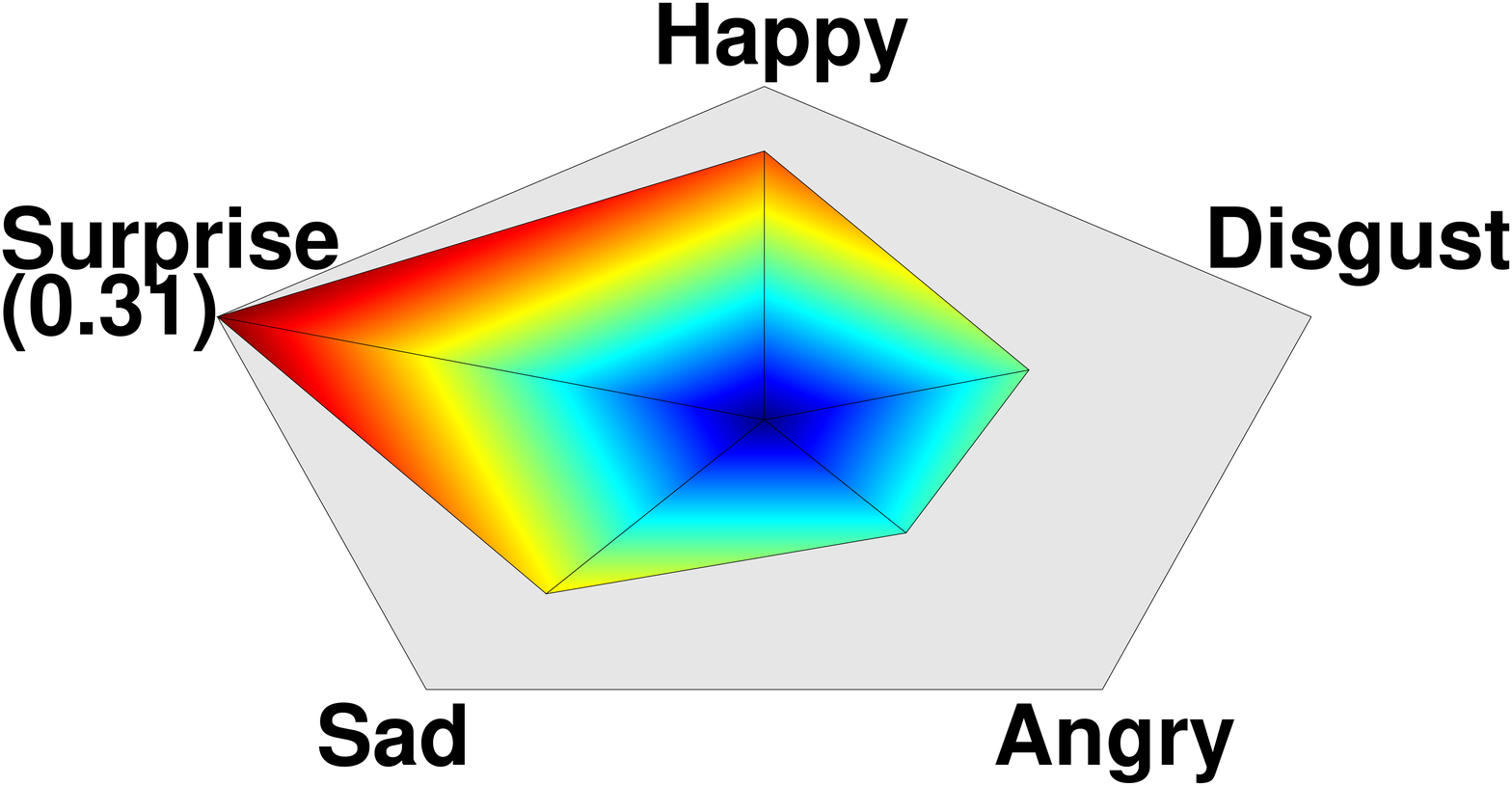}&
\includegraphics[width=0.12\textheight,height=0.08\textheight]{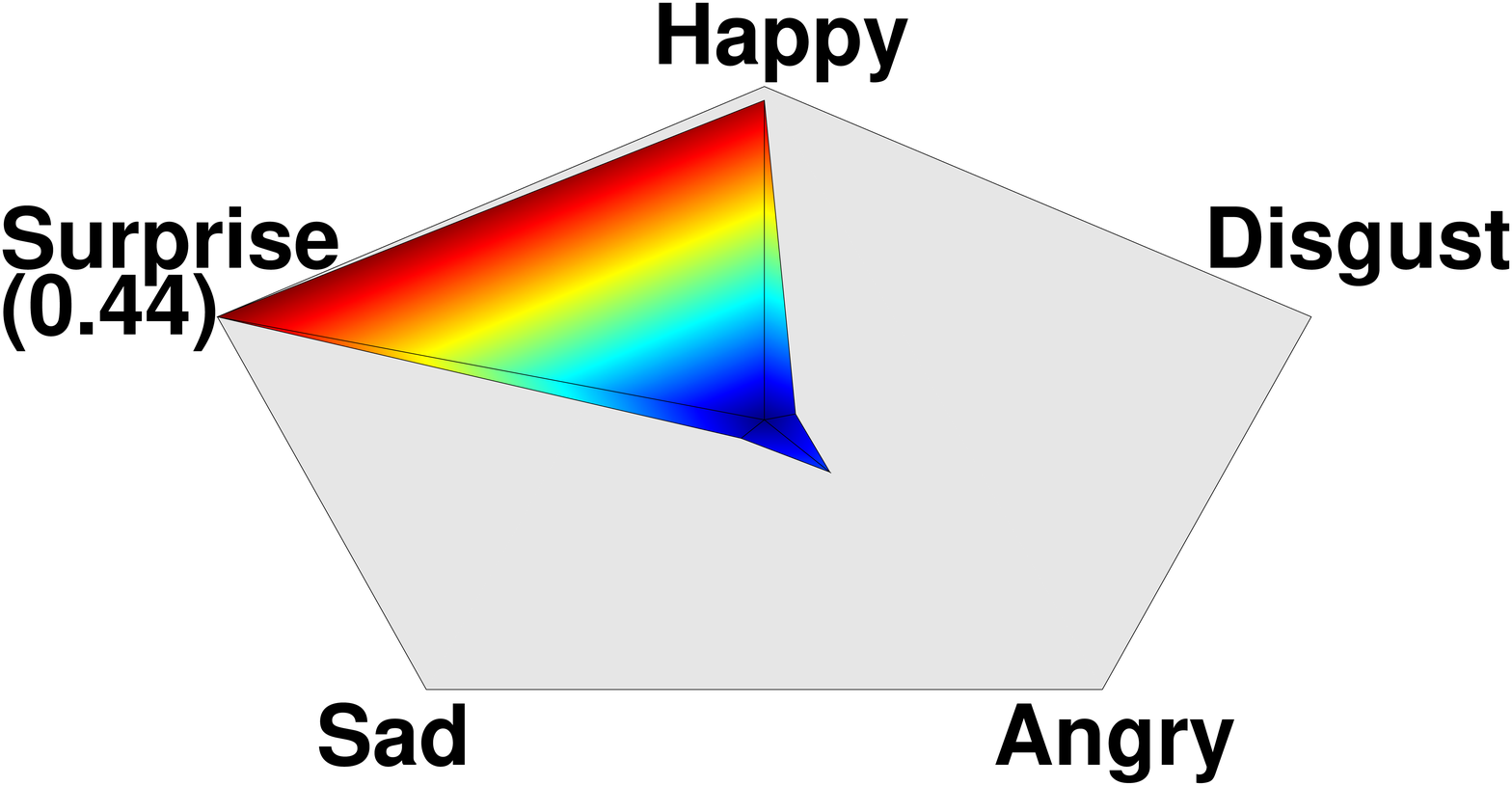} \\\hline
53&\includegraphics[width=0.076\textheight]{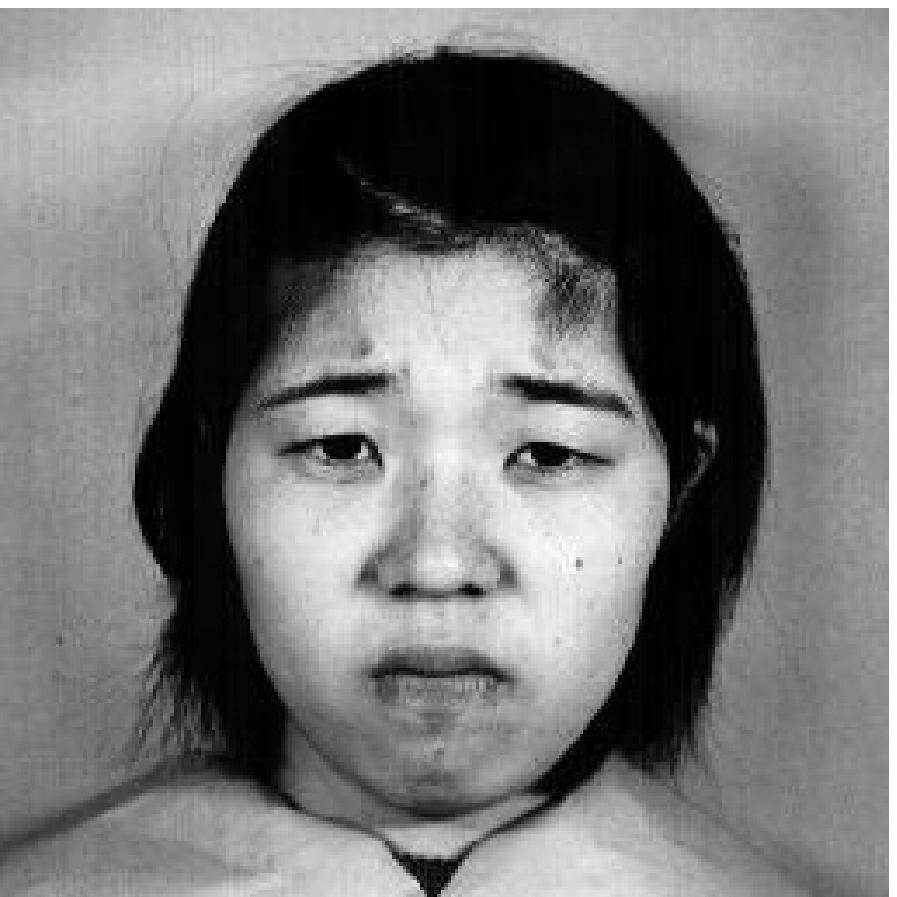}&
\includegraphics[width=0.12\textheight,height=0.08\textheight]{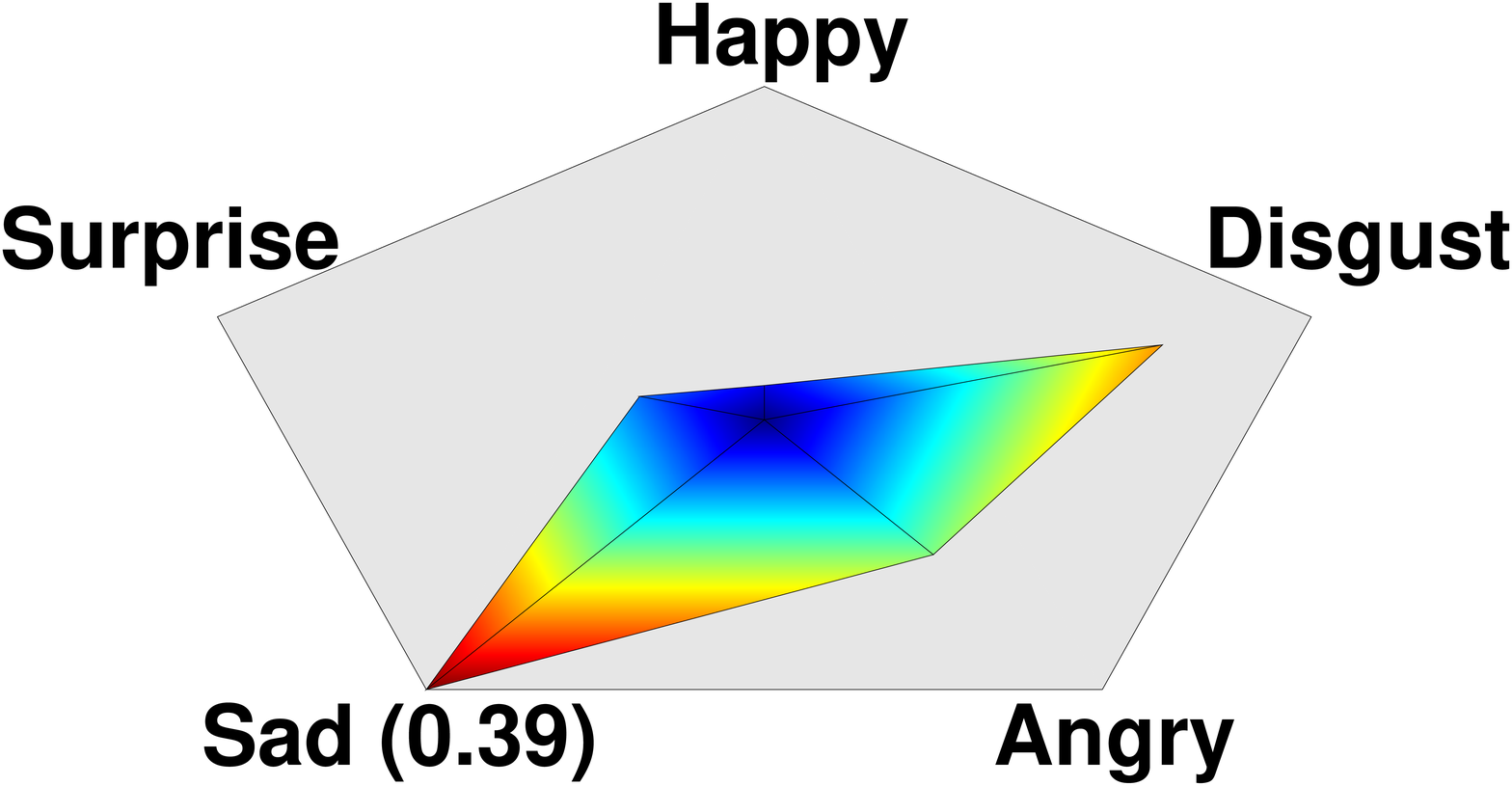}&
\includegraphics[width=0.12\textheight,height=0.08\textheight]{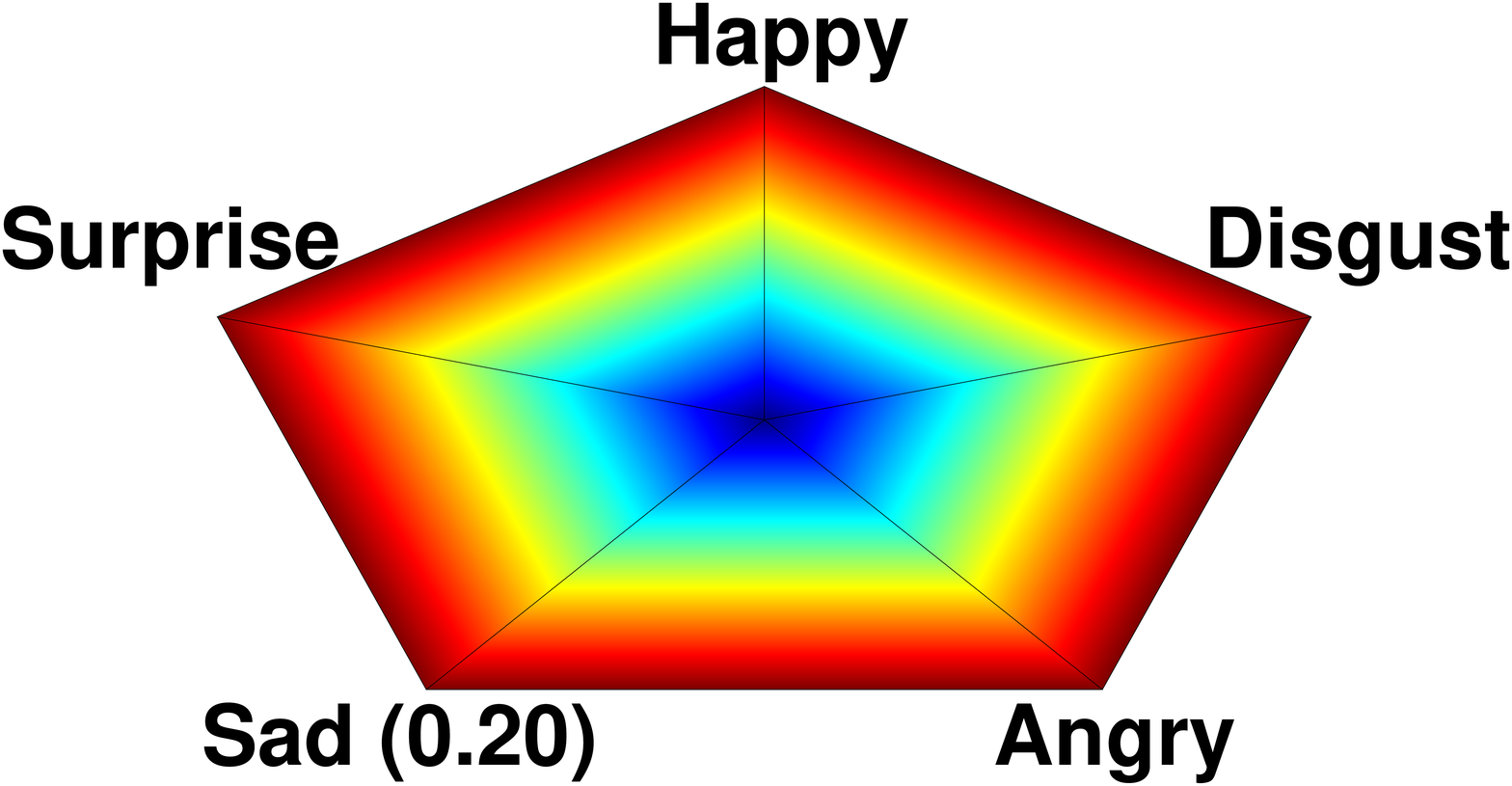}&
\includegraphics[width=0.12\textheight,height=0.08\textheight]{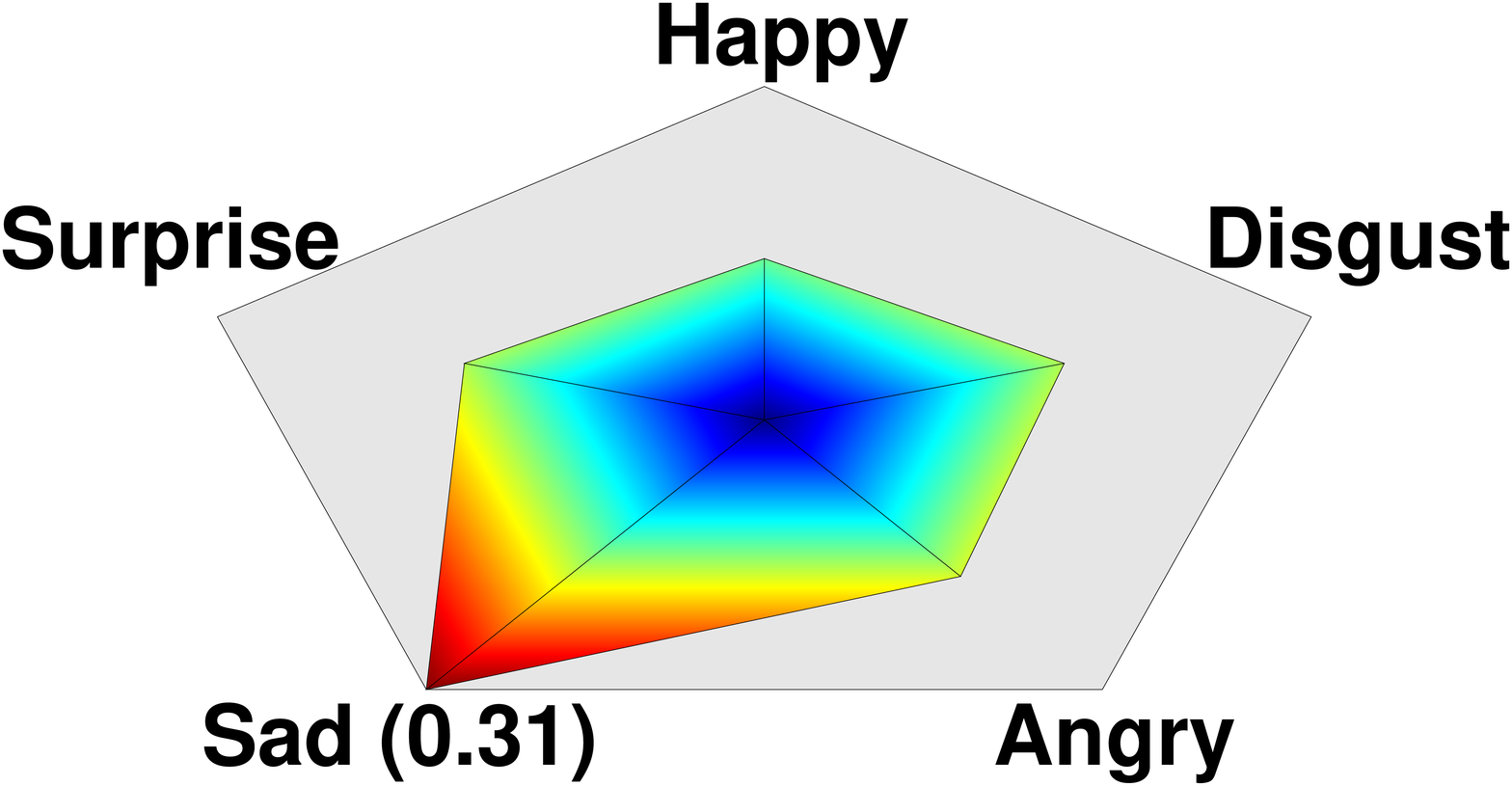}&
\includegraphics[width=0.12\textheight,height=0.08\textheight]{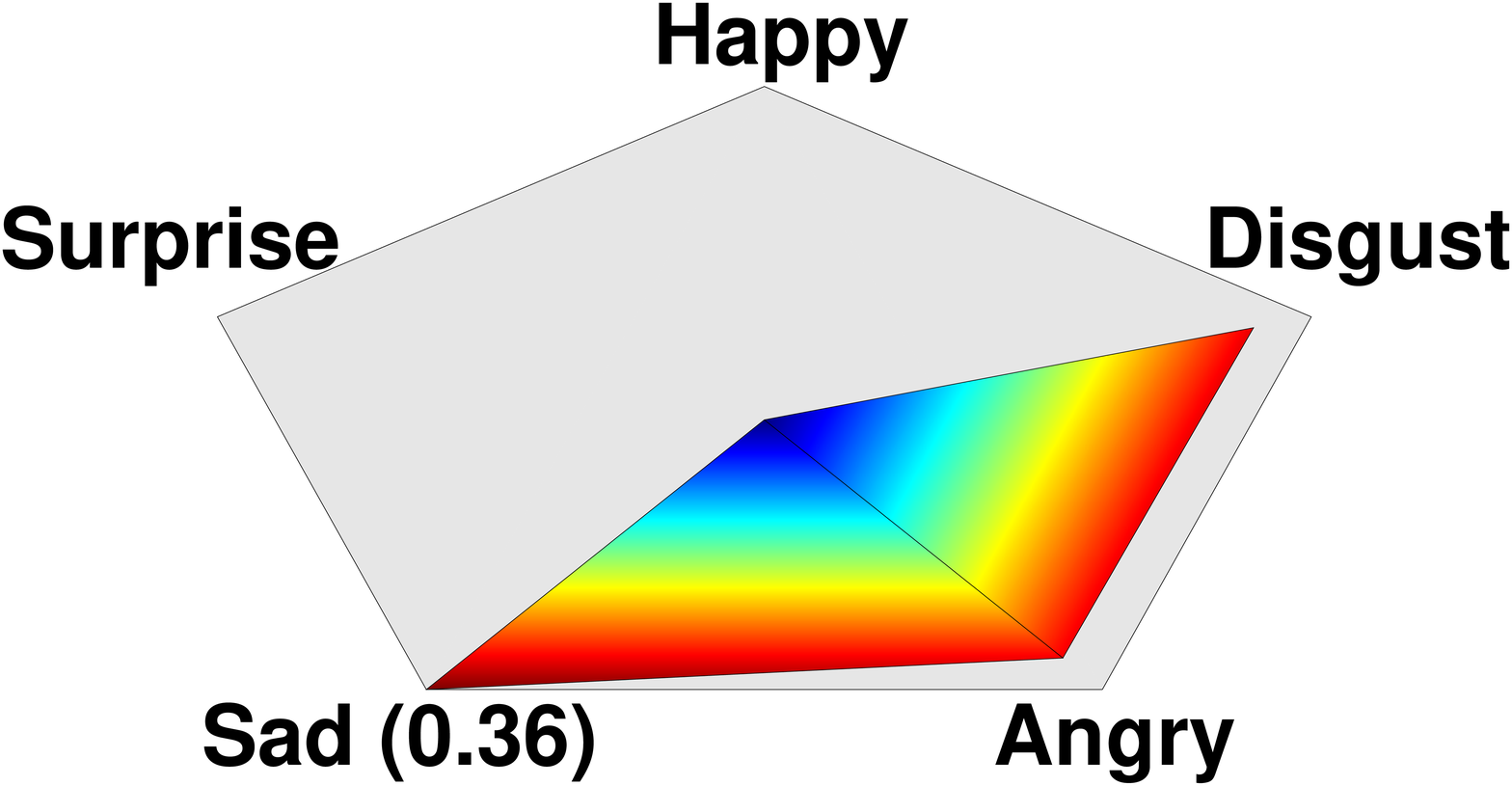} \\\hline
1&\includegraphics[width=0.076\textheight]{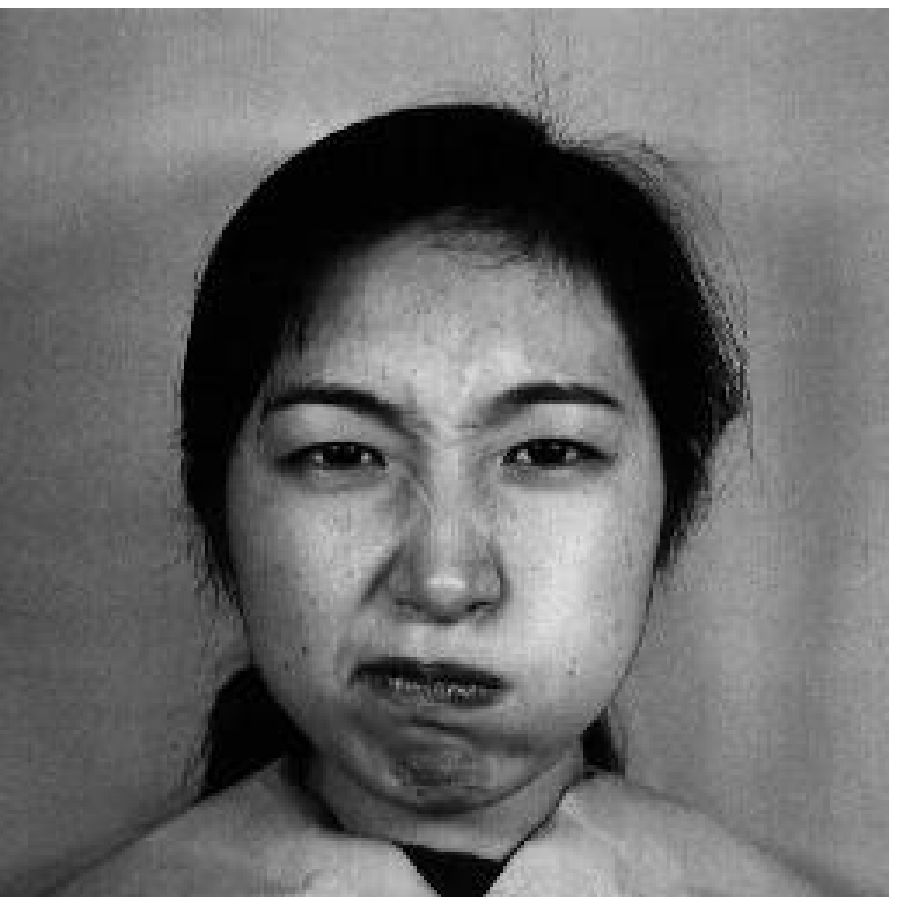}&
\includegraphics[width=0.12\textheight,height=0.08\textheight]{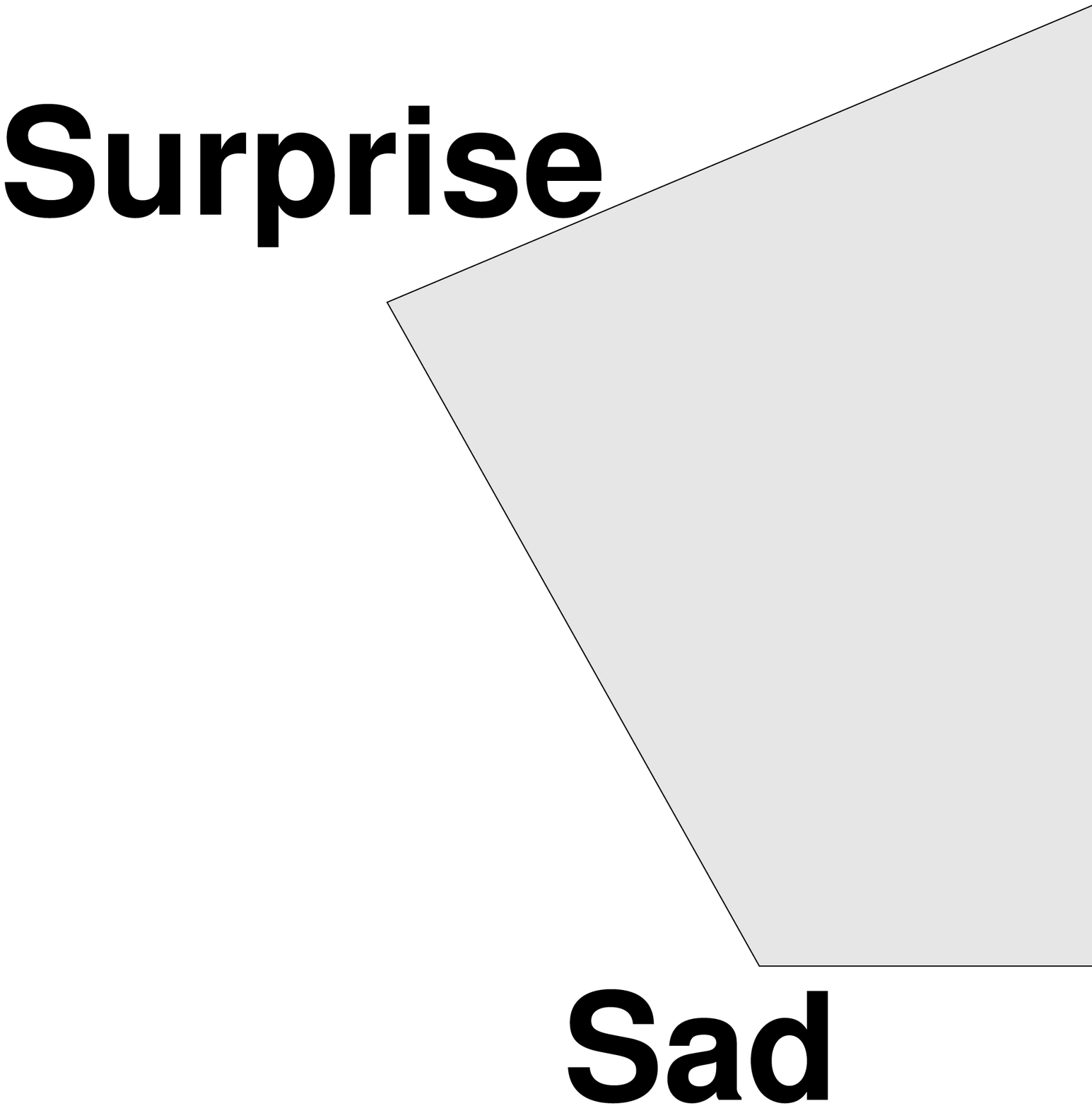}&
\includegraphics[width=0.12\textheight,height=0.08\textheight]{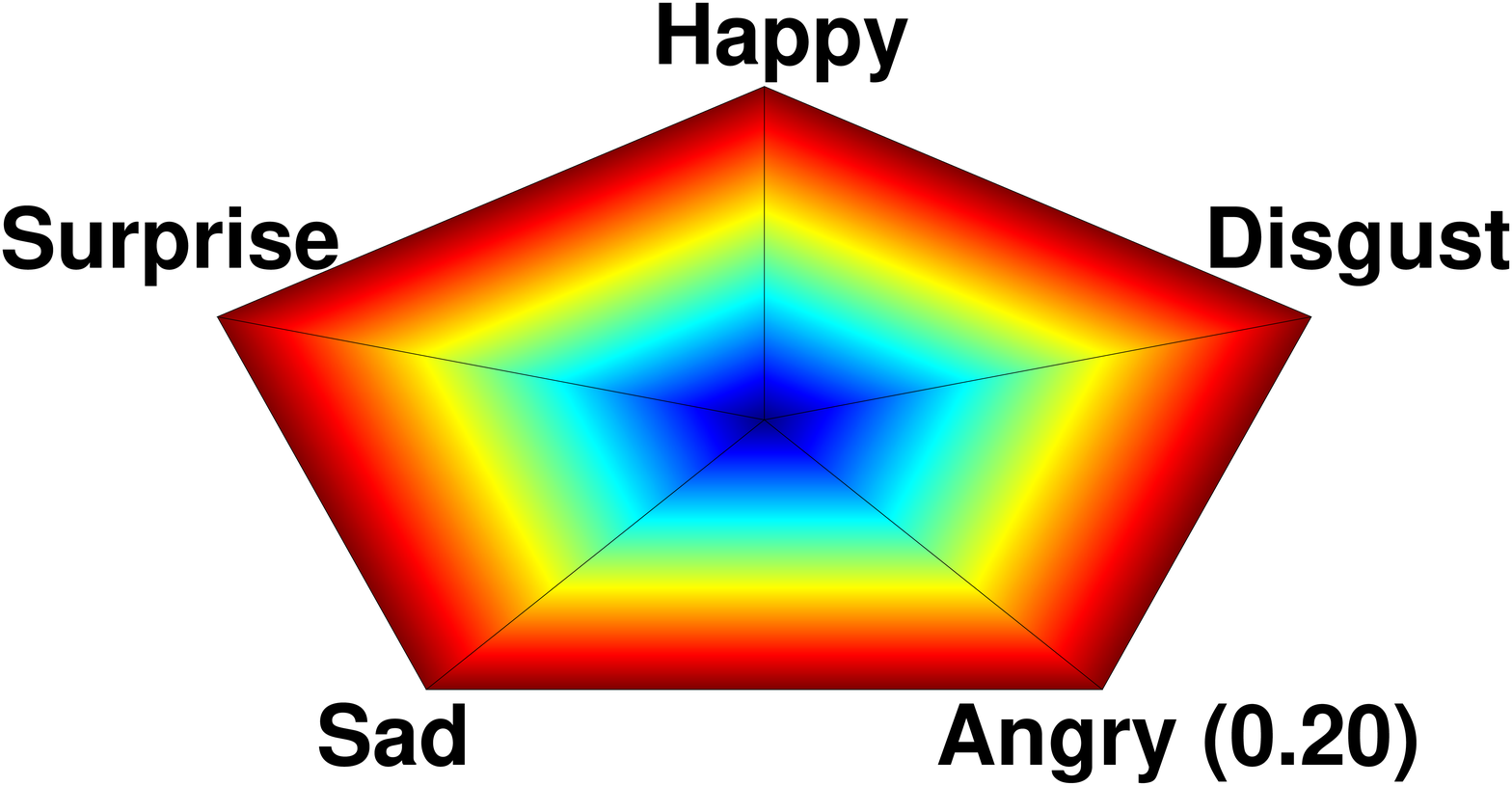}&
\includegraphics[width=0.12\textheight,height=0.08\textheight]{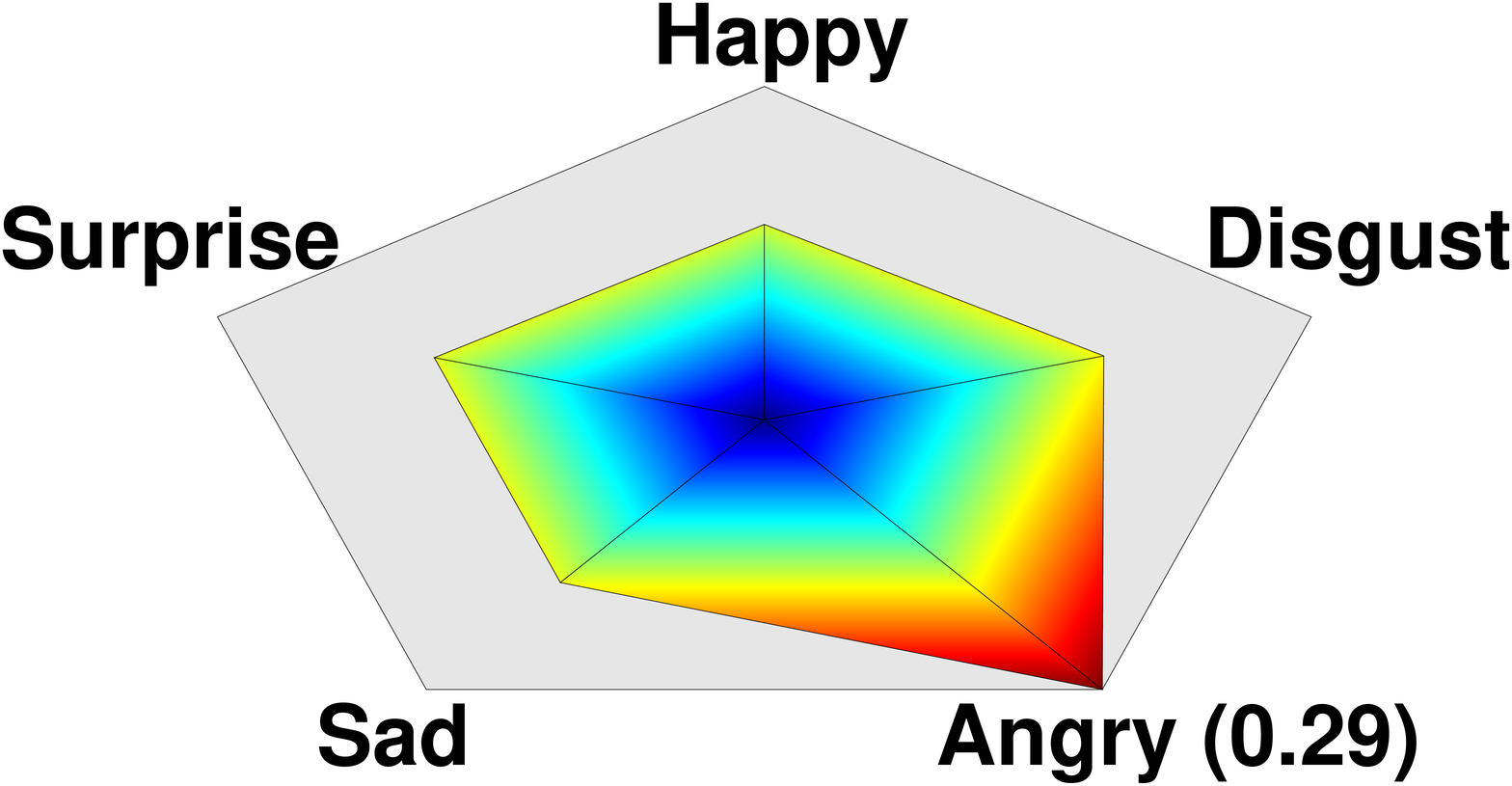}&
\includegraphics[width=0.12\textheight,height=0.08\textheight]{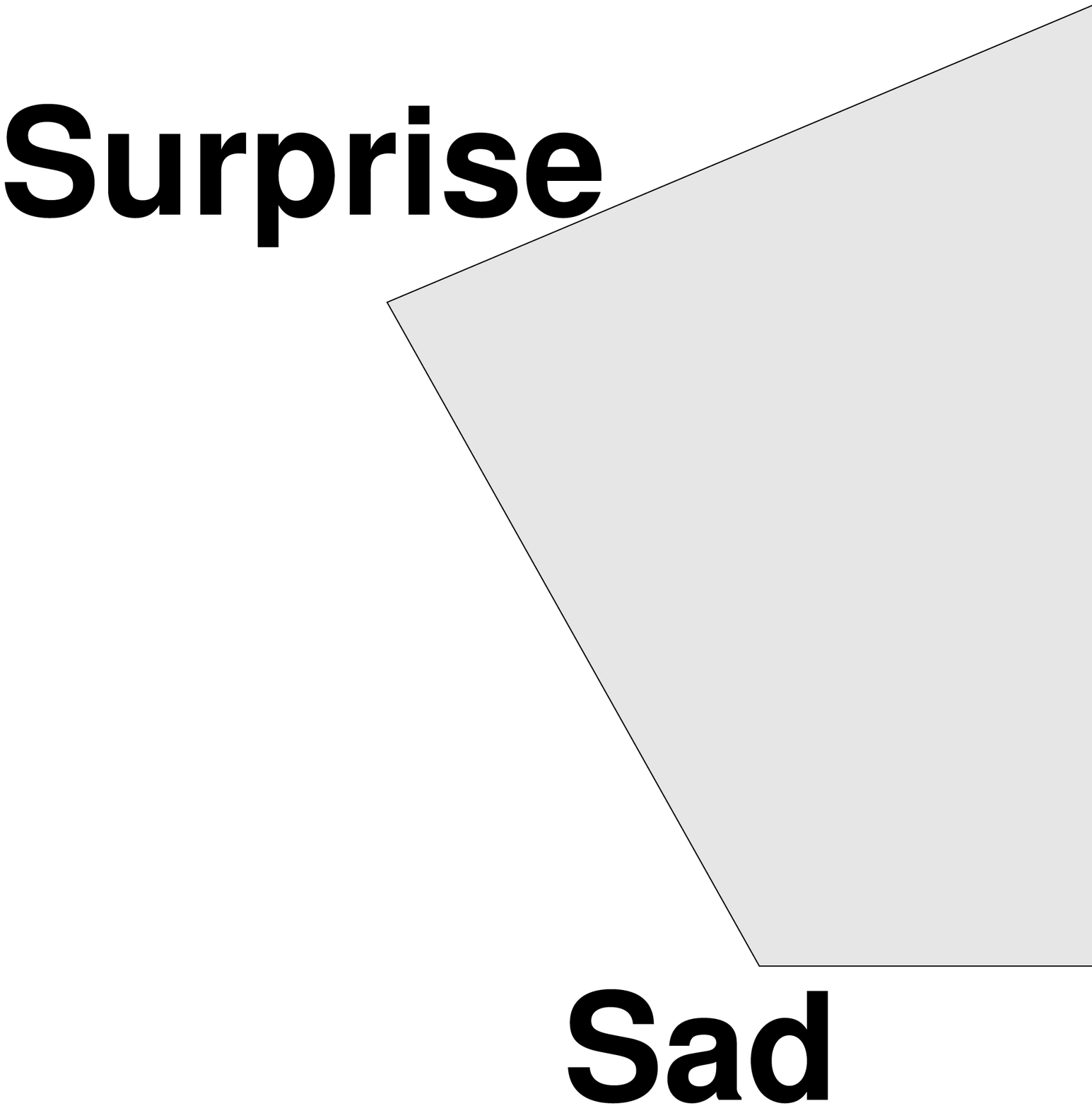} \\\hline
150&\includegraphics[width=0.076\textheight]{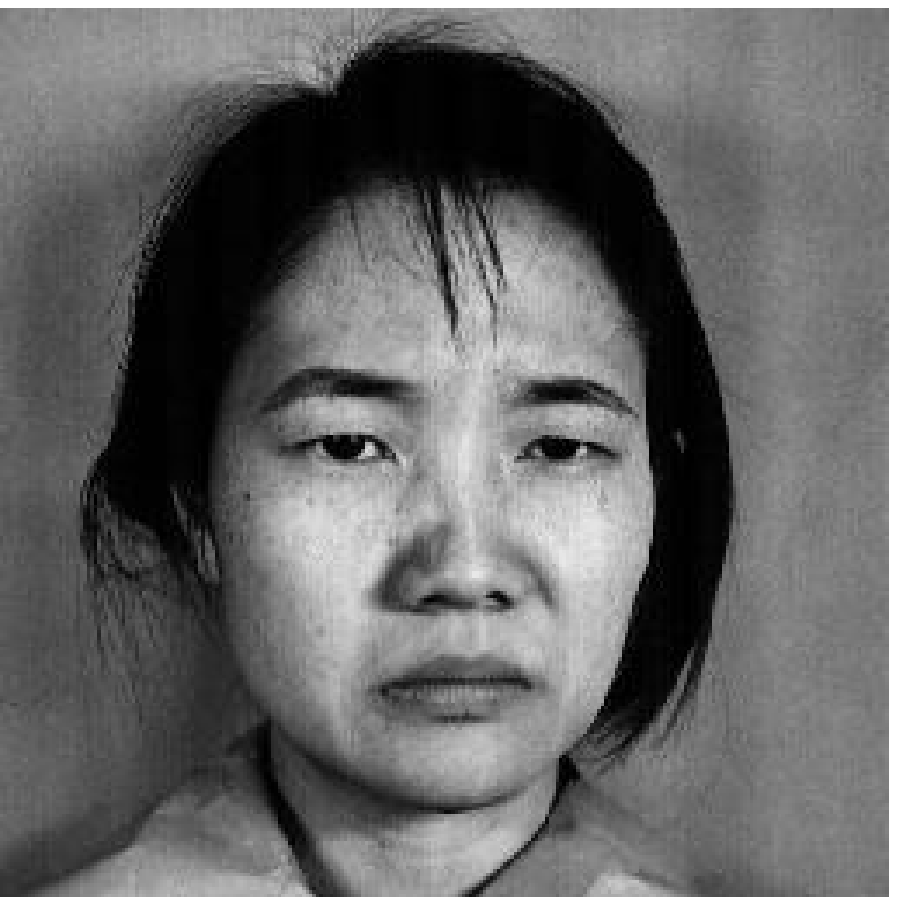}&
\includegraphics[width=0.12\textheight,height=0.08\textheight]{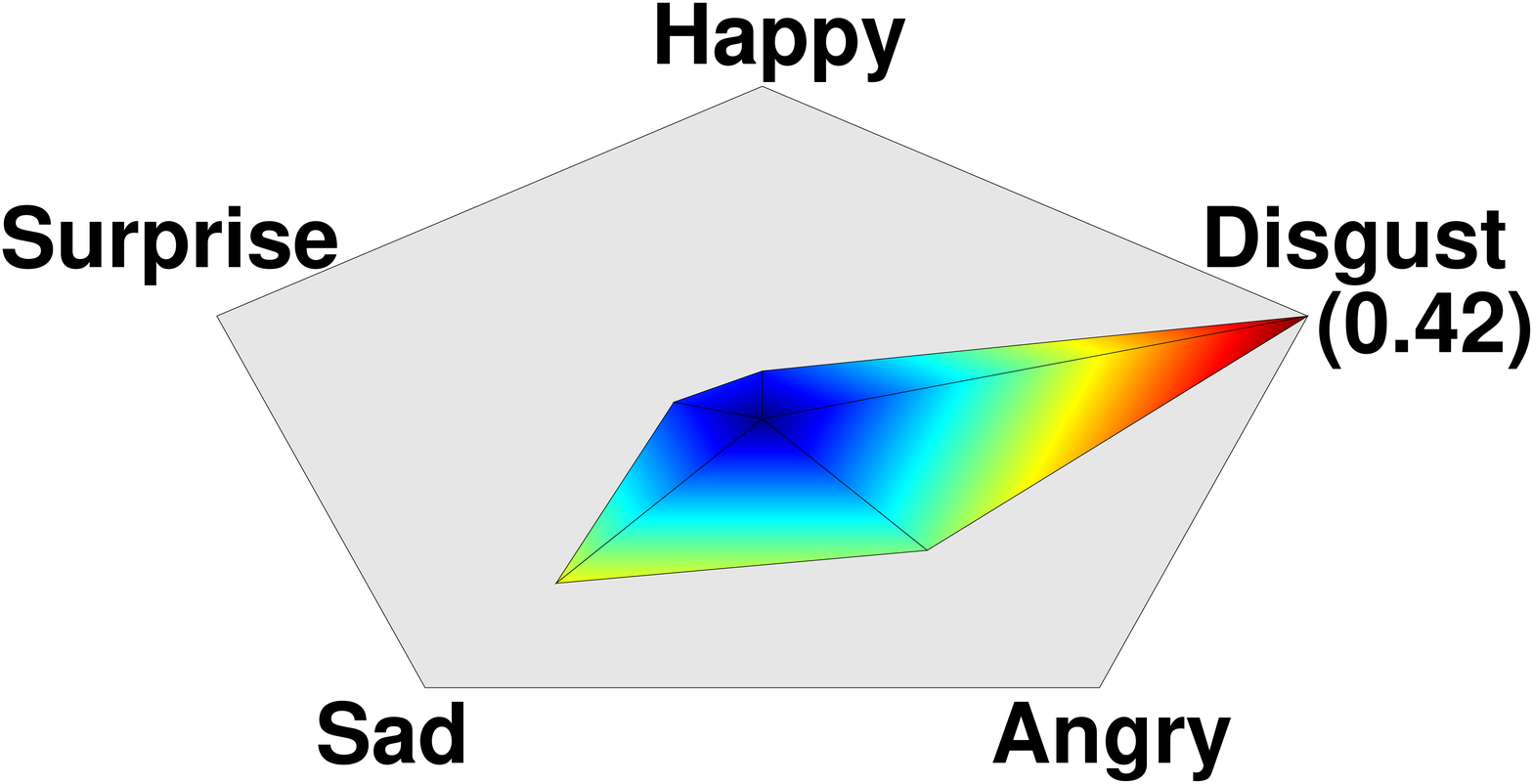}&
\includegraphics[width=0.12\textheight,height=0.08\textheight]{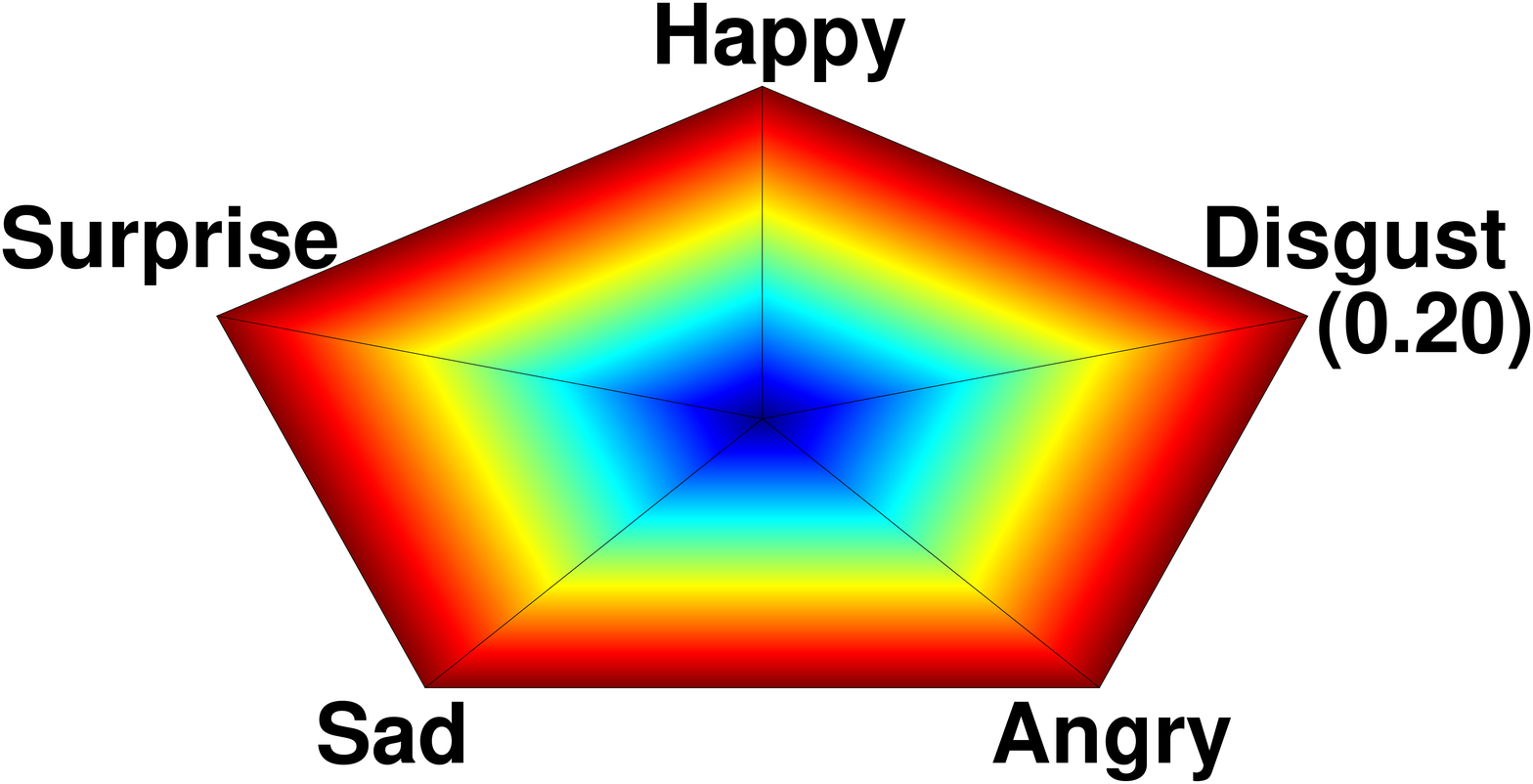}&
\includegraphics[width=0.12\textheight,height=0.08\textheight]{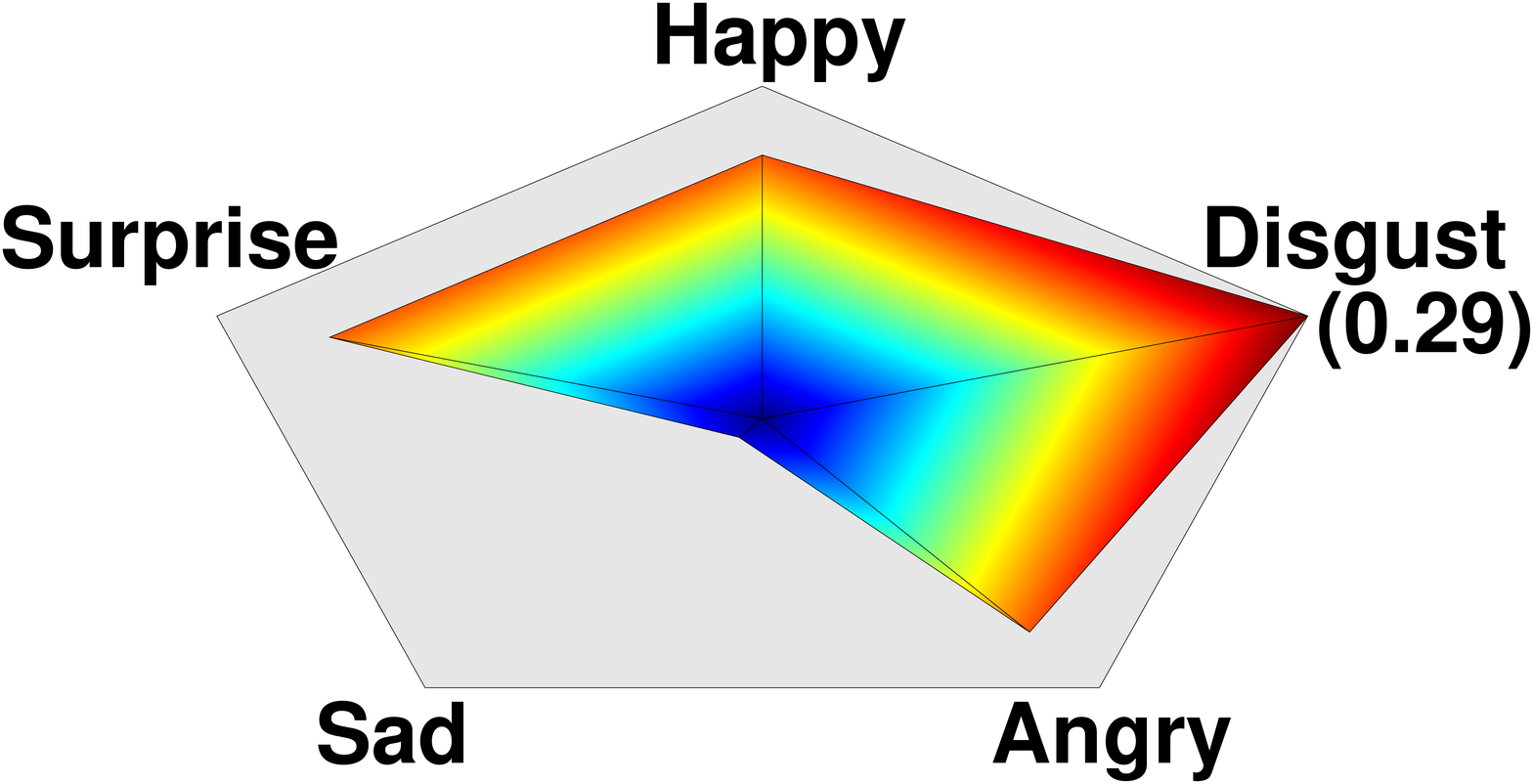}&
\includegraphics[width=0.12\textheight,height=0.08\textheight]{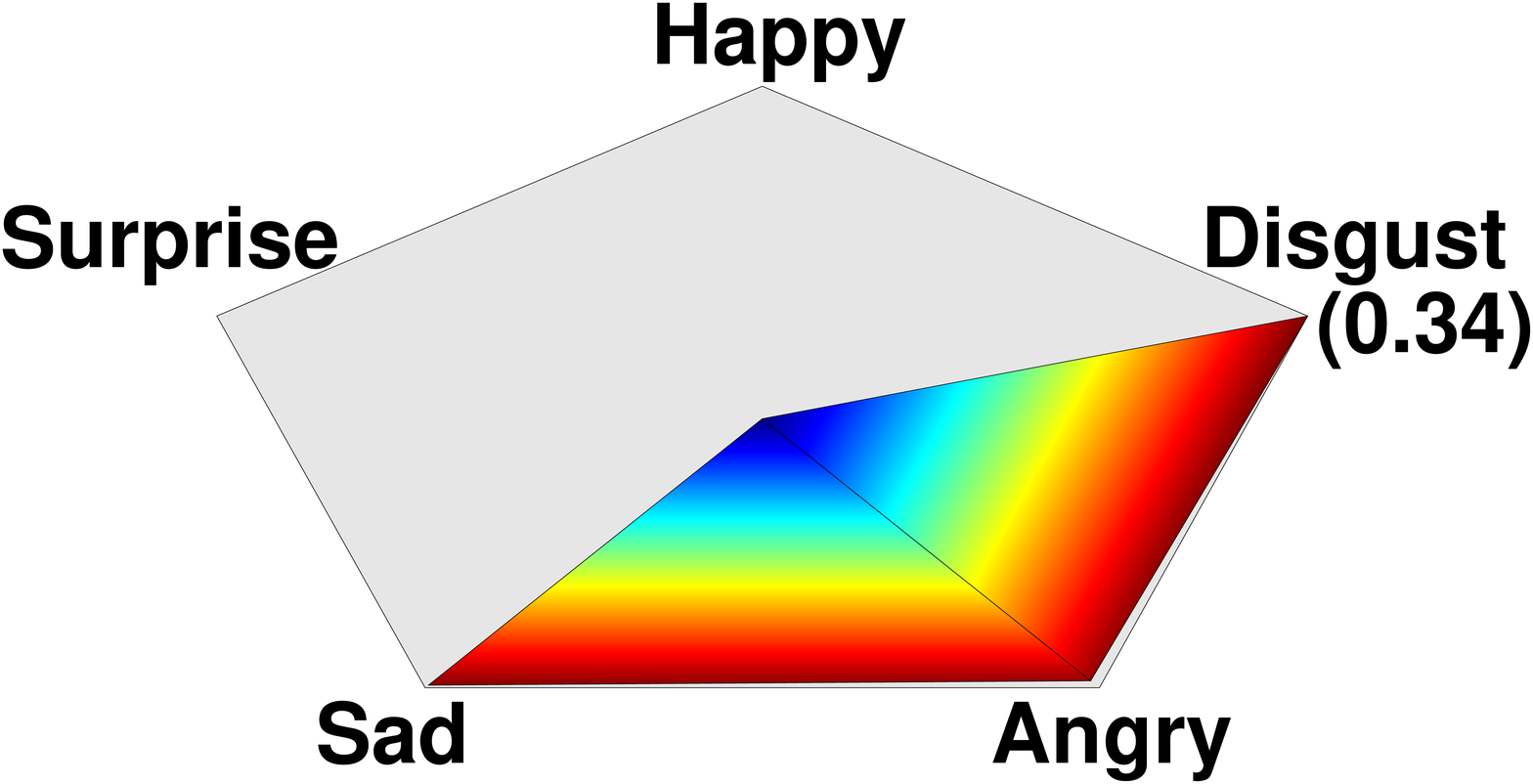} \\\hline
\end{tabular}\label{JAFFEexp}
\end{table*}

\subsection{Facial Expression Clustering}
This subsection experiments the facial expression clustering on the Japanese Female Facial Expression (JAFFE) Database \cite{JAFFE}. Data JAFFE includes two datasets of images from ten Japanese female expressers, whose averaged semantic ratings on six facial expressions are offered by $60$ Japanese viewers. The six basic facial expressions are ``Happy'', ``Surprise'', ``Sad'', ``Angry'', ``Disgust'' and ``Fear''. In this database, the first dataset consists of 213 images and the second dataset consists of 181 images by excluding the ``fear'' attributes and images. We hired the second dataset and offered five groups of fuzzy pairwise constraints, where these groups in sequence contains $0.2m$, $0.4m$, $0.6m$, $0.8m$ and $m$ fuzzy pairwise constraints. These fuzzy pairwise constraints were offered following the equation \eqref{Cost3}. Then, we implemented these clustering models on the dataset with these groups of fuzzy pairwise constraints and reported the results in Table \ref{JAFFEall}, where the cluster number is set to $5$. Another criterion Accuracy (Acc.) \cite{TWSVC} and the classical $k$means \cite{Kmeans2} were added in the experiment to reveal the overall clustering performance. Table \ref{JAFFEall} shows that the clustering performance is improved with the increasing number of fuzzy pairwise constraints for all these semi-supervised models, which implies that the pairwise constraints are very useful to facial expression clustering problems. Thereinto, the largest improvement arises in our FDC indicates its outperformance with fuzzy pairwise constraints. Moreover, the highest criteria of FDC reveals that it discovers more facial expressions than other models.

For facial expression clustering problem, the challenge is to discover more expressions in an image if it contains more than one expressions as shown in Fig. \ref{FigFicalExp}. Due to fuzzy clustering obtains fuzzy vectors regarded as ratings on the clusters, we focus on FCM, PCCA and FDC to evaluate the ability of discovering more expressions in the following. We propose two vector-level criteria based on ranking: (i) Minimal Average Hamming Distance (MAHD); (ii) Largest Index Assignment (LIA). After ranking two fuzzy vectors, their Hamming distance \cite{Hamming} can be calculated easily if their cluster indices are aligned. The ground truth cluster indices of JAFFE have been given, but the cluster indices of fuzzy matrix are not known by fuzzy clustering. Thus, we define the MAHD as the minimum of the average Hamming distance between the ground truth matrix and a fuzzy matrix for all possible cluster indices. Table \ref{Hamming} reported the MAHDs of the three fuzzy clustering models on the JAFFE with five groups of fuzzy pairwise constraints, where the smallest values were bold. Compared with FCM and PCCA, the MAHDs of our FDC are the smallest ones on the five groups, which implies that the fuzzy matrix by FDC is more similar to the ground truth. Though more fuzzy pairwise constraints greatly improve the hard clustering results in Table \ref{JAFFEall}, the improvements in Table \ref{Hamming} are inapparent. Therefore, we infer from Table \ref{Hamming} that our FDC may discover more expressions than FCM and PCCA, but the number of discovered expressions would be limited.

\begin{table}[htbp]
\caption{Minimal average Hamming distance (MAHD) on JAFFE} \centering
\begin{tabular}{llll}
\hline
 Group&FCM&PCCA&FDC\\\hline
(i)&0.7614$\pm$0.0087 &0.7603$\pm$0.0209 &$\mathbf{0.7233}$$\pm$0.0126 \\\hline
(ii)& 0.7614$\pm$0.0087&0.7606$\pm$0.0257 &$\mathbf{0.7162}$$\pm$0.0092 \\\hline
(iii)&0.7614$\pm$0.0087 &0.7554$\pm$0.0131 &$\mathbf{0.7176}$$\pm$0.0106 \\\hline
(iv)&0.7614$\pm$0.0087 &0.7599$\pm$0.0144 &$\mathbf{0.7179}$$\pm$0.0090 \\\hline
(v)&0.7614$\pm$0.0087 &0.7504$\pm$0.0204 &$\mathbf{0.7117}$$\pm$0.0095 \\\hline
\end{tabular}\label{Hamming}
\end{table}

The criterion MAHD has several shortcomings, e.g., it cannot detail each cluster, the calculation only suits for the same size of fuzzy matrices, and the permutation number is going to be huge for a slightly larger $k$. Hence, we hire another LIA to evaluate the contributions of fuzzy memberships based on the hard clustering criterion. Given a hard clustering criterion such as Acc., LIA obtains $k^*$ results by labeling the samples with the 1st, 2nd, $\ldots$, or $k^*$-th largest indices for all the fuzzy vectors in the ground truth and prediction, respectively. For instance, the criteria based on \eqref{UtoY} are actually based on the 1st LIA. The coherence reflexes the ability of discovering more expressions. Fig. \ref{FigLIA} reported the LIA-based Acc. of FCM, PCCA and our FDC on the JAFFE database with the five groups of fuzzy pairwise constraints. From Fig. \ref{FigLIA}, we observe that the 1st LIA-based Acc. of PCCA is lower than FCM and FDC, which is consistent with the results in Table \ref{JAFFEall}. Moreover, the lower overall LIA-based Acc. of PCCA implies that the recognition ability of PCCA on the facial expressions is lower than FCM and our FDC. Compared with FCM, the overall LIA-based Acc. of our FDC is more higher, which is consistent with MAHD. By comparing the coherence among the LIA-based Acc. in Fig. \ref{FigLIA}, we observe that: (i) The coherence of PCCA is disordered; (iii) The 1st and 2nd LIA-based Acc.'s of FCM are coherent, with the 3rd slightly lower LIA-based Acc.; (iii) The phenomenon of FCM appears in FDC. Consequently, it is inferred that for the ability of discovering more expressions in an image, FCM and our FDC can discover the 2nd expression with the ability the same as discovering the 1st one, and they may discover the 3rd expression in an image but the ability is not as strong as previous. For PCCA, its ability is obviously weaker than FCM and FDC.

To further reveal the differences among these fuzzy clustering models, Table \ref{JAFFEexp} illustrated some examples of the results by the three fuzzy clustering models on JAFFE with the fuzzy pairwise constraints of group (v). Since the facial expression labels are unknown in these models, we selected several representatives and mixtures of the clusters in Table \ref{JAFFEexp}, where the number along with the facial expression is the largest fuzzy membership in the fuzzy vector. Apparently, the differences of fuzzy memberships in FCM are tiny, and these in PCCA are disordered, supporting the conclusion from Fig. \ref{FigLIA}. From the images in Table \ref{JAFFEexp}, we deem that the representatives of our FDC on ``Happy'', ``Surprise'' and ``Disgust'' are much accurate. For the images with more expressions, the distinctiveness of FDC is superior to FCM and PCCA obviously. Therefore, our FDC outperforms FCM and PCCA on JAFFE from various perspectives.

\section{Conclusion}
The fuzzy pairwise constraint has been proposed in fuzzy clustering, and a fuzzy discriminant clustering (FDC) model has also been proposed to utilize the fuzzy pairwise constraints. The discriminant structure of cluster prototypes and piecewise cost function of fuzzy pairwise constraint allow our FDC to present the fuzzy characteristics precisely. The nonconvex optimization problem in FDC has been decomposed into several CQPPs and IQPPs by the MEM algorithm, where the global solutions to these CQPPs have been given explicitly or solved by some CQPP solvers, and the stationary points to these IQPPs have been obtained by a proposed DBCD algorithm efficiently. Under certain conditions, e.g., binary clustering problem with disjoint fuzzy pairwise constraints, it has been proved that the global solutions to these IQPPs can be obtained by the DBCD algorithm. Moreover, FDC has been extended into various metric spaces to suit for different applications. Experimental results on the benchmark datasets and facial expression clustering problem have indicated that our FDC outperforms many other state-of-the-art clustering models. For practical convenience, the corresponding FDC codes have been uploaded upon \url{https://github.com/gamer1882/FDC}. Future work includes applying fuzzy pairwise constraint for other fuzzy models and designing specific discriminant structures \cite{DPC,MFPC} to suit for various applications.

\section*{Acknowledgment}
This work is supported in part by National Natural Science Foundation of
China (Nos. 61966024, 61866010 and 11871183), in part by Natural Science Foundation of Inner Mongolia Autonomous Region (Nos. 2019BS01009, 2019MS06008), and in part by the Fundamental Research Funds for the Central Universities, JLU.

\ifCLASSOPTIONcaptionsoff
  \newpage
\fi



%

%
\bibliographystyle{IEEEtran}
\bibliography{FBib}
\clearpage

\section*{Appendices}

\subsection{Proof of Theorem \ref{ThmFrame}}
\begin{IEEEproof}
From the procedure of MEM, it is obvious that the cluster number may reduce to 1 at most. Suppose the final cluster number is $k^*$. Then, once the cluster number reduce to $k^*$, the objective of \eqref{1} in neither the expectation step nor maximization step increases in iteration. Since the objective in \eqref{1} has a lower bound based on its constraints, the series of the objectives of problem \eqref{1} obtained by MEM converges.
\end{IEEEproof}

\subsection{Proof of Theorem \ref{ThmConvert}}
\begin{IEEEproof}
From the constraint $\bfA\bfu=\mathbf{1}^{r}$ in problem \eqref{OriginQPP}, the equation $\bfu=\hat{\bfu}+\bfB\bfv$ always holds, where the columns of $\bfB$ build the fundamental system of solutions to the system of homogeneous linear equations $\bfA\bfu=\mathbf{0}^r$. Substituting the above equation into problem \eqref{OriginQPP}, it is convert to problem \eqref{MainIQPP}. Therefore, the solutions to problems \eqref{OriginQPP} and \eqref{MainIQPP} satisfy equation \eqref{uvSolution}.
\end{IEEEproof}

\subsection{Proof of Lemma \ref{Thm2}}
\begin{IEEEproof}
Suppose $\bfB^\top\bfD\bfB\bfv=\lambda\bfv$, where $\lambda<0$ is an eigenvalue and $\bfv$ is the corresponding eigenvector. We have $\bfv^\top\bfB^\top\bfD\bfB\bfv=\lambda$, i.e, $(\bfB\bfv)^\top\bfD(\bfB\bfv)<0$. Thus, $\bfD$ is indefinite or negative semi-definite alternatively. From Lemma \ref{lem1}, $\bfD$ is indefinite.
\end{IEEEproof}

\subsection{Proof of Theorem \ref{ThmConvergence}}
\begin{IEEEproof}
From the constraints of problem \eqref{OriginQPP}, $\mathbf{0}^{rk}\leq \bfu\leq \mathbf{1}^{rk}$ holds. Thus, the objective $\frac{1}{2}\bfu^\top \bfD\bfu$ has a lower bound. The CQPPs \eqref{QPP1} are always solved in the steps of DBCD, which guarantees that the objective does not increase in each step. Combining the above facts, the series of the objective values $\{\frac{1}{2}\bfu^{(t)}\bfD\bfu^{(t)}|t=1,2,\ldots\}$ converges.
Since the block subproblem \eqref{QPP1} is strictly convex, its unique global solution can always be obtained. Thus, any accumulation points of the block coordinate descent algorithm (i.e., DBCD) are the stationary points \cite{BCDtheory1,BCDtheory2}.

\end{IEEEproof}

\subsection{Proof of Corollary \ref{Cor1}}
\begin{IEEEproof}
Note that for $k=r=2$, problem \eqref{MainIQPP} becomes to
\begin{eqnarray}\label{MainIQPP2}
\begin{array}{l}
\underset{\bfv}{\min}~~\frac{1}{2}\bfv^\top \bfB^\top \bfD\bfB\bfv+\hat{\bfu}^\top \bfD\bfB\bfv\\
s.t.~~~~\mathbf{0}^{2}\leq \bfv\leq\mathbf{1}^{2}.
\end{array}
\end{eqnarray}
From the proof of Theorem \ref{ThmConvert}, we just need to prove that there is a vertex $\bfv^{(0)}$ to problem \eqref{MainIQPP2} such that the equivalence of DBCD with $\bfv^{(0)}$ converges to the global solution to problem \eqref{MainIQPP2}.

Suppose $\bfv^*=(v_1^*,v_2^*)^\top$ is the global solution to problem \eqref{MainIQPP2}, where $v_1^*,v_2^*\in \mathbb{R}$. Then, we will show that $v_1^*$ is 0 or 1, or $v_2^*$ is 0 or 1 alternatively. Suppose $0<v_1^*<1$ and $0<v_2^*<1$. Since $\bfB^\top \bfD\bfB$ is indefinite, there is always a feasible descent direction for all feasible interior points, which is contradict with the fact that $v^*$ is the global solution. Thus, $0<v_1^*<1$ and $0<v_2^*<1$ cannot holds simultaneously. Without loss of generality, we suppose $v_1^*$ is 0 or 1.

Note that the equivalence of DBCD becomes to the classical coordinate descent algorithm. If $v_2^*=0$ or $v_2^*=1$, the conclusion of corollary holds obviously. Without loss of generality, suppose $v^*=(0,v_2^*)^\top$, where $0<v_2^*<1$. Then, we will show that the classical coordinate descent algorithm with initial vertex $v^{(0)}$ which equals $(0,0)^\top$ or $(0,1)^\top$ converges to $v^*$. For $v^*$, there is always an infeasible descent direction $(d_1,d_2)^\top$, which is the eigenvector of $\bfB^\top \bfD\bfB$ corresponding to the negative eigenvalue, towards the negative infinity. We have $d_1<0$ obviously. Moreover, $d_2\neq0$ holds; Otherwise, the direction $(d_1,0)^\top$ (i.e., the coordinate direction) is towards the negative infinity, which is contradict with the fact that for any fixed $v_2$ problem \eqref{MainIQPP2} w.r.t. $v_1$ is a CQPP. Therefore, there is an infeasible point $v'$ equals $(v_1',0)^\top$ or $(v_1',1)^\top$ with $v_1'<0$, which decreases the objective of \eqref{MainIQPP2} from $v^*$. Then, the objective at $v'$ is less than it at corresponding $(0,0)^\top$ or $(0,1)^\top$, which implies that 0 is global solution to problem \eqref{MainIQPP2} w.r.t. $v_1$ in the coordinate descent algorithm. In the next step, the coordinate descent algorithm obtain the global solution $(0,v_2^*)$ from CQPP \eqref{MainIQPP2} w.r.t. $v_2$.
\end{IEEEproof}






\end{document}